\documentclass[a4paper, 10pt, conference]{ieeeconf}      
\PassOptionsToPackage{dvipsnames}{xcolor}
\usepackage{FG2027}
\usepackage[T1]{fontenc}
\usepackage[dvipsnames]{xcolor}
\usepackage{graphicx,amsmath,amssymb,booktabs,bm,xspace}
\usepackage[normalem]{ulem}
\usepackage{wrapfig,siunitx}
\usepackage{caption,subcaption}
\usepackage{cite}
\usepackage[breaklinks,hidelinks,hypertexnames=false]{hyperref}
\usepackage{hypcap}
\usepackage{orcidlink}
\usepackage[accsupp]{axessibility}
\usepackage[capitalize]{cleveref}
\crefname{section}{Sec.}{Secs.}
\Crefname{section}{Section}{Sections}
\crefname{table}{Tab.}{Tabs.}
\Crefname{table}{Table}{Tables}
\DeclareUnicodeCharacter{2010}{-}
\DeclareUnicodeCharacter{00D7}{\ensuremath{\times}}

\newif\ifarxiv
\arxivtrue 

\newcommand{\msup}[2]{#1\textsuperscript{\scriptsize$\pm$#2}}          
\newcommand{\bmsup}[2]{\textbf{#1}\textsuperscript{\textbf{\scriptsize$\pm$#2}}} 
\newcommand{\meanstd}[2]{%
  \num[round-mode=figures,round-precision=3]{#1}%
  \textsuperscript{\scriptsize$\pm$\,\num[round-mode=figures,round-precision=2]{#2}}%
}
\newcommand{\bmeanstd}[2]{%
  {\sisetup{detect-all}%
   \textbf{\num[round-mode=places,round-precision=1]{#1}}%
   \textsuperscript{\scriptsize$\pm$\,\textbf{\num[round-mode=places,round-precision=2]{#2}}}%
  }%
}

\makeatletter
\DeclareRobustCommand\onedot{\futurelet\@let@token\@onedot}
\def\@onedot{\ifx\@let@token.\else.\null\fi\xspace}

\def\eg{\emph{e.g}\onedot}

\makeatother

\newcommand{\maketitlesupplementary}{%
  \clearpage
  \twocolumn[{\centering\Large\textbf{FIELDS: Face reconstruction with accurate Inference of Expression using Learning with Direct Supervision}\\[0.5em]
  Supplementary Material\\[1em]}]%
}
\newcommand{\umeanstd}[2]{%
  \underline{\num[round-mode=figures,round-precision=3]{#1}}%
  \textsuperscript{\scriptsize$\pm$\,%
  \num[round-mode=figures,round-precision=2]{#2}}%
}

\FGfinalcopy 

\IEEEoverridecommandlockouts                              
\def\FGPaperID{3} 

\title{\LARGE \bf
FIELDS: Face reconstruction with accurate Inference of Expression using
Learning with Direct Supervision
}

\ifFGfinal
\author{\parbox{16cm}{\centering
    {\large Chen Ling$^1$\orcidlink{0009-0008-1914-1694} \quad
    Henglin Shi$^{1,2}$\orcidlink{0000-0001-5884-8475} \quad
    Hedvig Kjellström$^1$\orcidlink{0000-0002-5750-9655}}\\
    {\normalsize
    $^1$ KTH Royal Institute of Technology, Sweden \quad $^2$ Linköping University, Sweden\\
    {\ttfamily\small \{chenlin,hedvig\}@kth.se \quad henglin.shi@liu.se}}
}}
\else
\author{Anonymous FG2027 submission\\ Paper ID \FGPaperID \\}
\fi

\begin{document}
\ifarxiv
\pagestyle{plain}
\else
  \ifFGfinal
  \thispagestyle{empty}
  \pagestyle{empty}
  \else
  \pagestyle{plain}
  \fi
\fi


\twocolumn[{%
  \renewcommand\twocolumn[1][]{#1}%
  \maketitle
  \captionsetup{type=figure}
  \begin{center}
  \capstart
    \centerline{\includegraphics[width=0.98\textwidth]{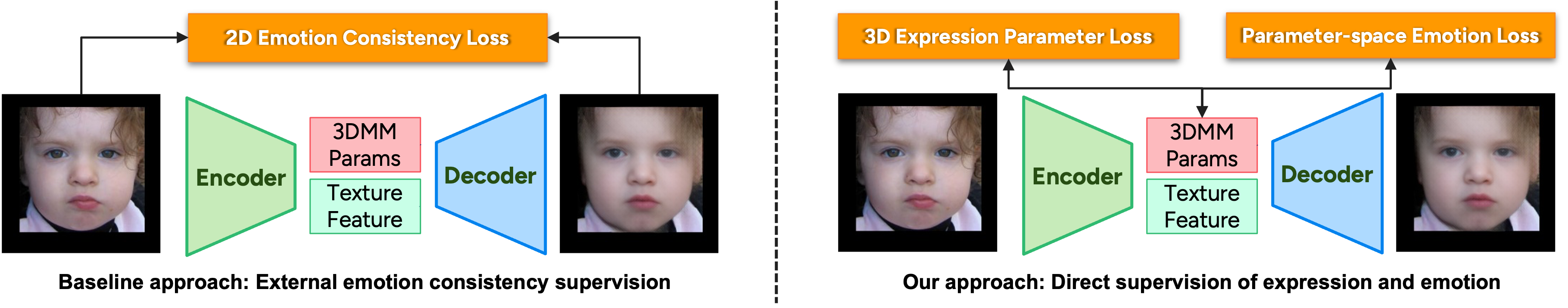}}
  \captionof{figure}{
  Comparison between existing emotion supervision paradigms and the proposed strategy.
EMOCA~\cite{emoca} and SMIRK~\cite{smirk} are trained with an image-level
emotion consistency loss computed between input and reconstructed images,
while FIELDS is directly supervised by scan-derived expression parameters
and a parameter-space emotion loss.}
  \label{fig:compare}
  \end{center}
}]

\ifarxiv
\thispagestyle{plain}
\fi

\begin{abstract}
Monocular 3D face reconstruction estimates the parameters of a
3D morphable face model (3DMM) from a single image, providing an explicit
geometric representation for facial expression analysis.
Owing to the scarcity of paired 2D images and 3D facial meshes, most methods are trained in a self-supervised manner by rendering the predicted face and minimizing image-level reconstruction losses, such as photometric and landmark errors, against the input image.
However, these losses enforce visual alignment but do not ensure that the
predicted parameters preserve affective information.
Recent methods add image-level emotion consistency loss, but it can be satisfied by exaggerating facial deformations.
To address these issues, we propose FIELDS (Face reconstruction with accurate Inference of
Expression using Learning with Direct Supervision), which learns affect-informative expression representations
in the FLAME parameter space.
Specifically, FIELDS augments 2D self-supervised reconstruction with scan-derived
expression anchoring and a parameter-space emotion loss.
During training,
scan-fitted targets directly constrain expression-related FLAME parameters to discourage exaggerated deformations,
while a jointly trained emotion head uses discrete
emotion and continuous valence--arousal annotations to make the predicted parameters informative for affect analysis.
Experiments show that FIELDS improves affective utility from the predicted
parameters in both within- and cross-dataset settings.
Notably, it also achieves lower reconstruction errors
overall and across most facial regions than the comparison baselines
under both rigid and similarity alignment.
\end{abstract}

\section{Introduction}
\label{sec:intro}

In interpersonal communication, nonverbal signals including facial expressions and body gestures convey substantial emotional content beyond spoken words \cite{nonverbal, universals}. 
In particular, facial expressions convey rich and fine-grained affective cues, making facial expression recognition (FER) a fundamental pillar of affective computing~\cite{affect}. 
FER enables systems to map visible facial movements to internal emotional states~\cite{er}, which is indispensable in emotion-aware applications such as social robots~\cite{socialrobot}, driver fatigue detection~\cite{driversfatigue}, and in-clinic pain assessment~\cite{pain}.
However, despite the progress driven by deep learning~\cite{2020FERsurvey,2023FERsurvey,2024FERsurvey}, purely appearance-based cues remain unreliable in the wild due to variations in pose, illumination, occlusion, and identity-specific morphology, making robust, geometry-aware expression representations critical for dependable FER.
In this context, 3D morphable models (3DMMs) offer an interpretable parameterization that factorizes facial geometry into pose, identity shape, and expression. Beyond face reconstruction, their compact expression parameters have proven effective for categorical emotion recognition and continuous affect estimation~\cite{koujan2020real,3DMMcontinuous,3dexpau,emoca}.


A common route to obtain such geometry-aware expression representations is monocular 3DMM reconstruction.
In learning-based approaches,
an encoder regresses 3DMM parameters (FLAME~\cite{FLAME} in this work) from a single image, and a differentiable renderer or an image-to-image synthesizer decodes them back to an image~\cite{mofa,3dfacereconsurvey}. These encoder--decoder pipelines are mainly trained in a self-supervised manner, where the consistency between the input and the rendered/synthesized images is used as the primary training signal~\cite{deng19accurate, regress3D, deca}.
Nevertheless, photometric consistency objectives alone do not necessarily make the learned expression-related 3DMM parameters capture the underlying affect information.

To address this, recent methods introduce affect- or expression-aware supervision that operates through the synthesized images, such as a perceptual emotion consistency loss~\cite{emoca} and expression re-targeting with cycle consistency~\cite{smirk}. However, an emotion consistency loss on the synthesized image~\cite{emoca} is arguably a quite weak supervision signal. Moreover, adding emotion consistency with a pre-trained external emotion recognition module has been observed to 
encourage reconstructed 3D faces 
with exaggerated expressions~\cite{smirk}. An image-level consistency loss can also be sensitive to a rendered-vs-real appearance gap~\cite{smirk}. Together, these issues expose a trade-off between the affect utility and its ability to render plausible images, motivating methods that circumvent this trade-off by explicitly supervising the 3DMM parameters.

\begin{figure}[t]
  \centering
  \includegraphics[width=0.93\columnwidth]{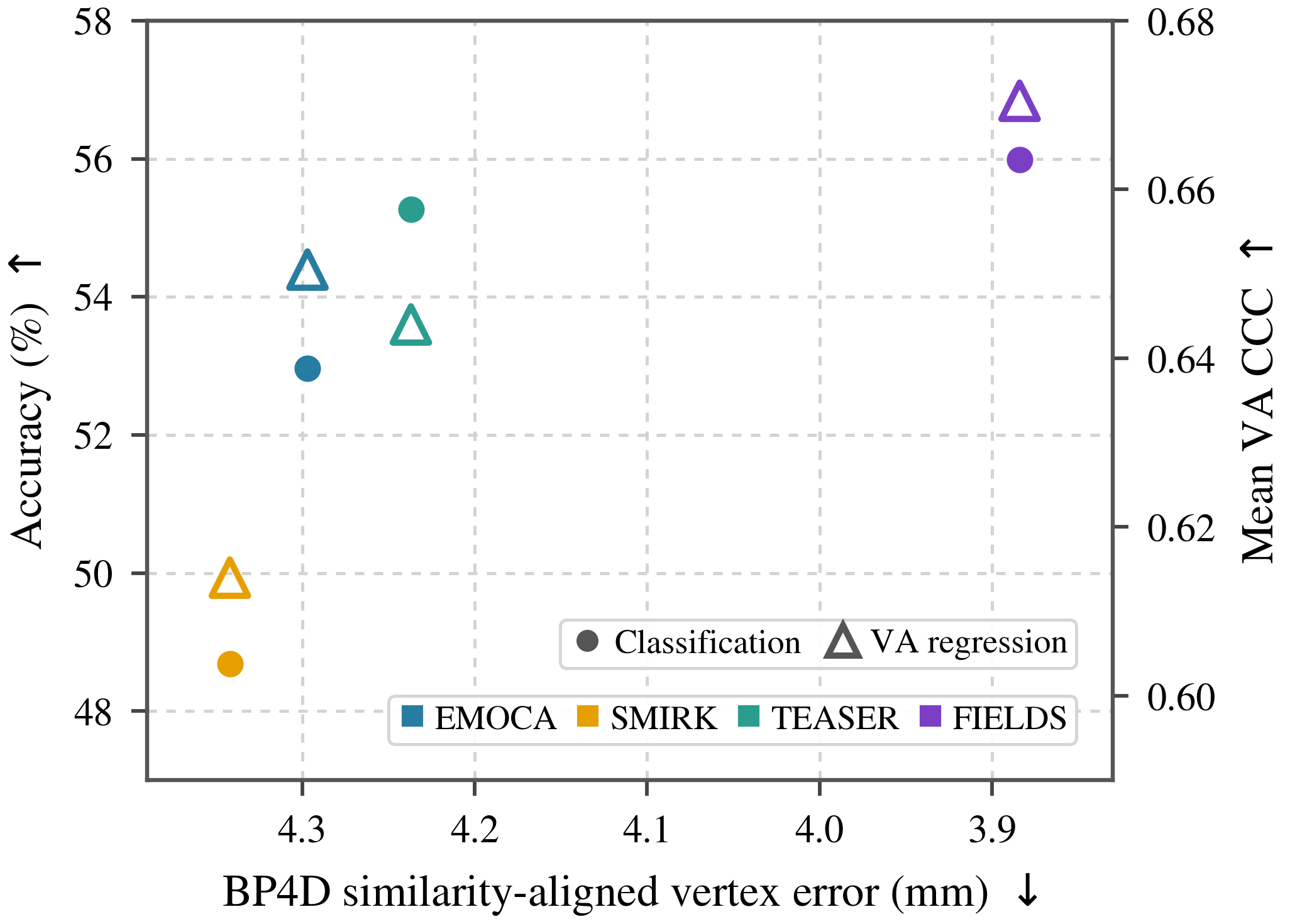}
  \caption{Affective utility vs.\ geometric fidelity. AffectNet accuracy (circles, left axis) and mean VA CCC
(triangles, right axis). Upper right is better.}
  \label{fig:tradeoff}
\end{figure}

We therefore propose FIELDS (\textbf{F}ace reconstruction with accurate \textbf{I}nference of \textbf{E}xpression using \textbf{L}earning with \textbf{D}irect \textbf{S}upervision) as illustrated in~\cref{fig:compare}, a task-driven 3DMM representation learning framework designed to address the utility-fidelity tension for FER. Built on the encoder-decoder self-supervision paradigm, FIELDS introduces hybrid supervision that encourages FLAME expression parameters to be more informative for affect while remaining geometrically plausible. 
Specifically, we obtain fit-derived anchors from BP4D scans and use them to supervise FLAME expression-related parameters, which helps to stabilize expression geometry and discourage exaggerated facial deformations.
In addition, we introduce auxiliary emotion supervision directly in the 3DMM parameter space (including continuous valence-arousal signals) to learn affect-informative expression representations without relying on perceptual comparisons between rendered and real images. As a preview, \cref{fig:tradeoff} shows that FIELDS achieves
stronger affective utility on AffectNet~\cite{affectnet} together
with lower vertex errors on BP4D~\cite{bp4d}
than recent leading 3D face reconstruction methods. 


In summary, our contributions are threefold:
\begin{itemize}
    \item We propose FIELDS, a hybrid supervision framework that combines 2D self-supervised reconstruction with scan-derived expression anchoring and a jointly trained affect recognition head that applies supervision directly in the FLAME parameter space.
    \item We construct a scan-derived FLAME parameter resource for BP4D to support reproducible expression learning and geometric evaluation. We will make the fitted parameters available to licensed BP4D users, subject to approval from the dataset owners.
    \item We demonstrate improved discrete and continuous affect prediction on AffectNet, and further show that the predicted 3DMM parameters support stronger within- and cross-dataset emotion classification on RAF-DB~\cite{li2017reliablerafdb,li2019reliablerafdb}. FIELDS also achieves lower BP4D reconstruction errors overall and across most facial regions under both rigid and similarity alignment than recent expression-focused monocular 3D face reconstruction methods.
\end{itemize}

\section{Related Work}
\label{sec:relatedworks}

\subsection{Facial Representation in Emotion Analysis}

Emotion analysis from images is typically defined according to one of two annotation methods~\cite{2020FERsurvey,2023FERsurvey,2024FERsurvey}: \emph{discrete} categories, most often the six universal emotions of \emph{Anger, Fear, Happiness, Surprise, Disgust, Sadness} as defined by Paul Ekman~\cite{universals}, or the \emph{continuous} valence-arousal space \cite{feldman1998independence,valencearousal}.

\begin{figure}[t]
  \centering
  \includegraphics[width=0.85\columnwidth]{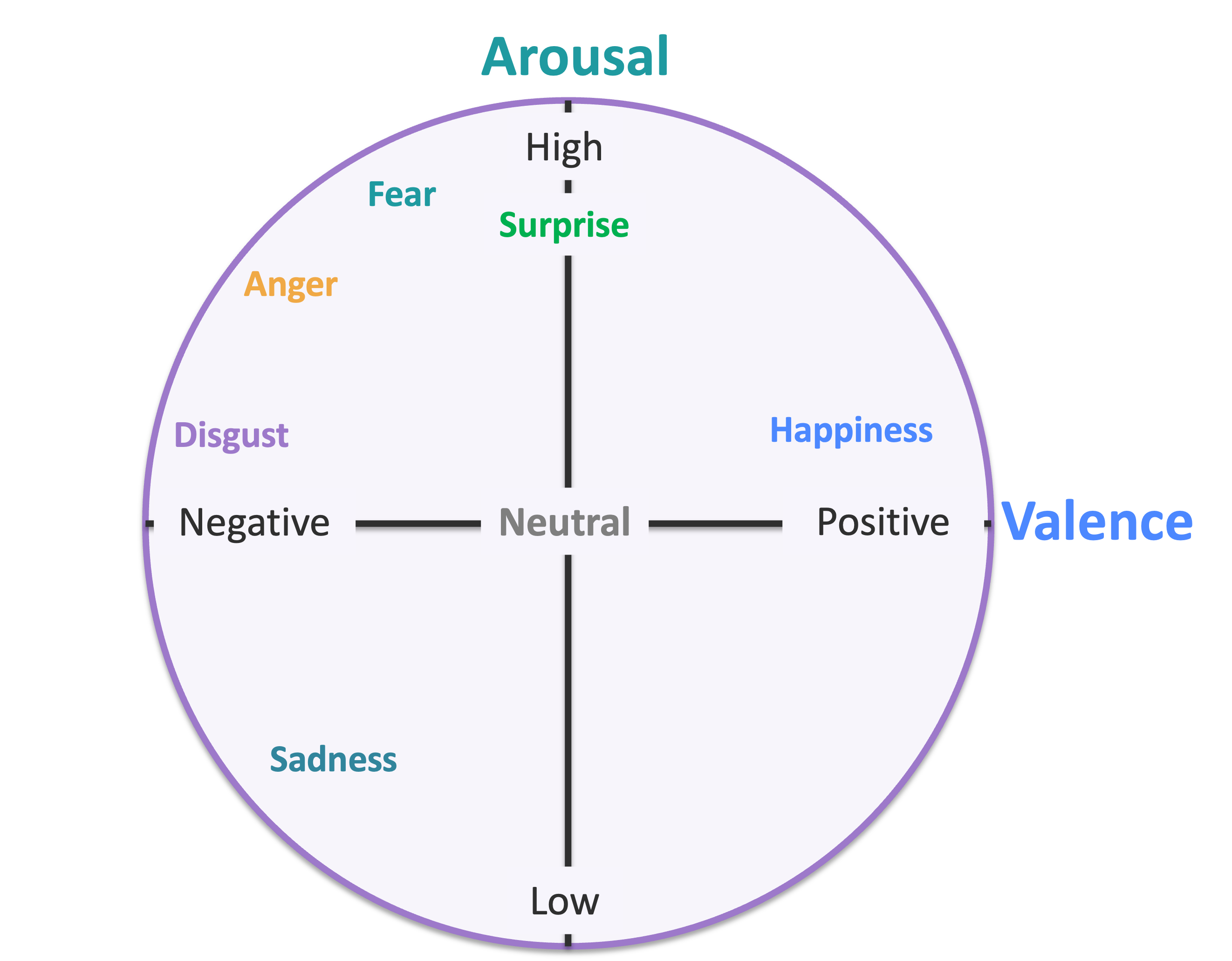}
  \caption{Valence--arousal circumplex. Discrete emotions are overlaid on a valence-arousal model of affect~\cite{valencearousal, feldman1998independence, universals}. Adapted from~\cite{vaemotion}.}
  \label{fig:VA}
\end{figure}

As shown in \cref{fig:VA}, valence denotes the hedonic tone of affect (negative to positive), whereas arousal indexes the level of activation or energy (low to high). Discrete emotion labels map to prototypical regions in the valence–arousal plane, whereas the circumplex’s continuous labels explicitly encode affect intensity and gradation -- information absent from categorical classes. Both methods are widely used in benchmarks and downstream applications, and they motivate representations that are sensitive to subtle, muscle-driven facial deformations~\cite{valencearousal, feldman1998independence}.

Early FER approaches rely on handcrafted 2D geometry cues -- sparse landmarks~\cite{landmarksFER}, contour shapes~\cite{contourshape}, and optical flow~\cite{opticalFER} -- to approximate muscle motion. Denser 3D observations, such as 2.5D depth maps~\cite{2.5DFER}, laser-scan meshes~\cite{scanFER}, and point clouds~\cite{pointcloudsFER}, provide finer deformation detail but are costly to acquire and often dataset-specific. These signals improve fidelity but, without explicit structure, can conflate person-specific geometry with expression.

3DMMs~\cite{first3DMM, BaselFace, FLAME, 3DMMsurvey} provide an interpretable factorization of facial variation into identity-related shape and expression-related deformation, offering geometry-aware representations beyond purely 2D appearance features. Recent works leverage 3DMM parameters or intermediate 3D features for emotion recognition and valence-arousal regression from images~\cite{expnet, koujan2020real, 3DMMcontinuous, bejaoui2017fully, 3dexpau,ig3d,emoca,smirk}, demonstrating that 3D facial representations are predictive for affect while being more robust to pose and illumination than 2D appearance alone. At the same time, prior studies~\cite{3Dlip, emoca, spectre} note that aggressively weighting emotion losses can inflate expressions and introduce artifacts, raising questions about \emph{geometric fidelity} despite higher emotion scores. We follow this 3DMM-based line and focus on learning affect-informative expression representations for downstream emotion analysis.

\subsection{Monocular 3DMM-based Face Reconstruction}

Here we focus on monocular 3DMM-based reconstruction for FER-ready facial codes. For a survey of model-free pipelines, \eg using direct mesh, voxels or implicit regression, see~\cite{3dfacereconreview,3dfacereconsurvey}. Within 3DMM  reconstruction methods, optimization-based fitting via analysis-by-synthesis is interpretable yet slow and sensitive to occlusions and initialization~\cite{op3d02,op3d18,op3d20,op3d21}, whereas learning-based regressors and neural rendering improve speed and realism but can trade geometric semantics for pixel fidelity and blur identity–expression boundaries.

We organize learning-based work by the supervision they employ: sparse~\cite{deca, deng19accurate, guo2020towards} or dense~\cite{alp17densereg, wood20223d} keypoints and contours~\cite{liu2017dense} calibrate coarse geometry but suffer from sparsity and detector noise; photometric terms~\cite{deng19accurate, deca, Genova2018CVPR, shang2020self, yang2020facescape} impact all parameters but are fragile to misalignment and rendering quality; perceptual losses~\cite{zhang2018unreasonable} align with human judgment yet are hard to balance and may exaggerate expressions~\cite{emoca, smirk}; identity features or multi-image constraints~\cite{deca,deng19accurate,Genova2018CVPR,Wu_2019_CVPR} strengthen subject consistency but risk biasing identity shape; explicit 3D supervision~\cite{expnet,guo2020towards,tuan2017regressing, Wu_2019_CVPR} (meshes or 3DMM parameters) offers stronger targets but inherits the biases of the pseudo ground truth. This motivates a recent line of work that injects affect- or expression-aware constraints into monocular 3DMM reconstruction, aiming to better capture the spectrum of facial expressions for FER-related tasks.

In this line, EMOCA~\cite{emoca} was the first to introduce an external emotion‐consistency loss via a pretrained affect classifier, explicitly encouraging affect preservation. SMIRK~\cite{smirk} complements this line with an expression-cycle consistency path and template/expression injection to encourage expression diversity and reduce over-compensation during training. TEASER~\cite{teaser} further improves expression detail via multi-scale appearance tokenization, providing stronger localized guidance for expression reconstruction without explicit affect supervision. Motivated by these complementary directions, we adopt a hybrid supervision scheme that combines image-consistency training with additional parameter-space constraints.

\section{Method}
\begin{figure*}[tb]  
  \centering
  \includegraphics[width=0.92\textwidth]{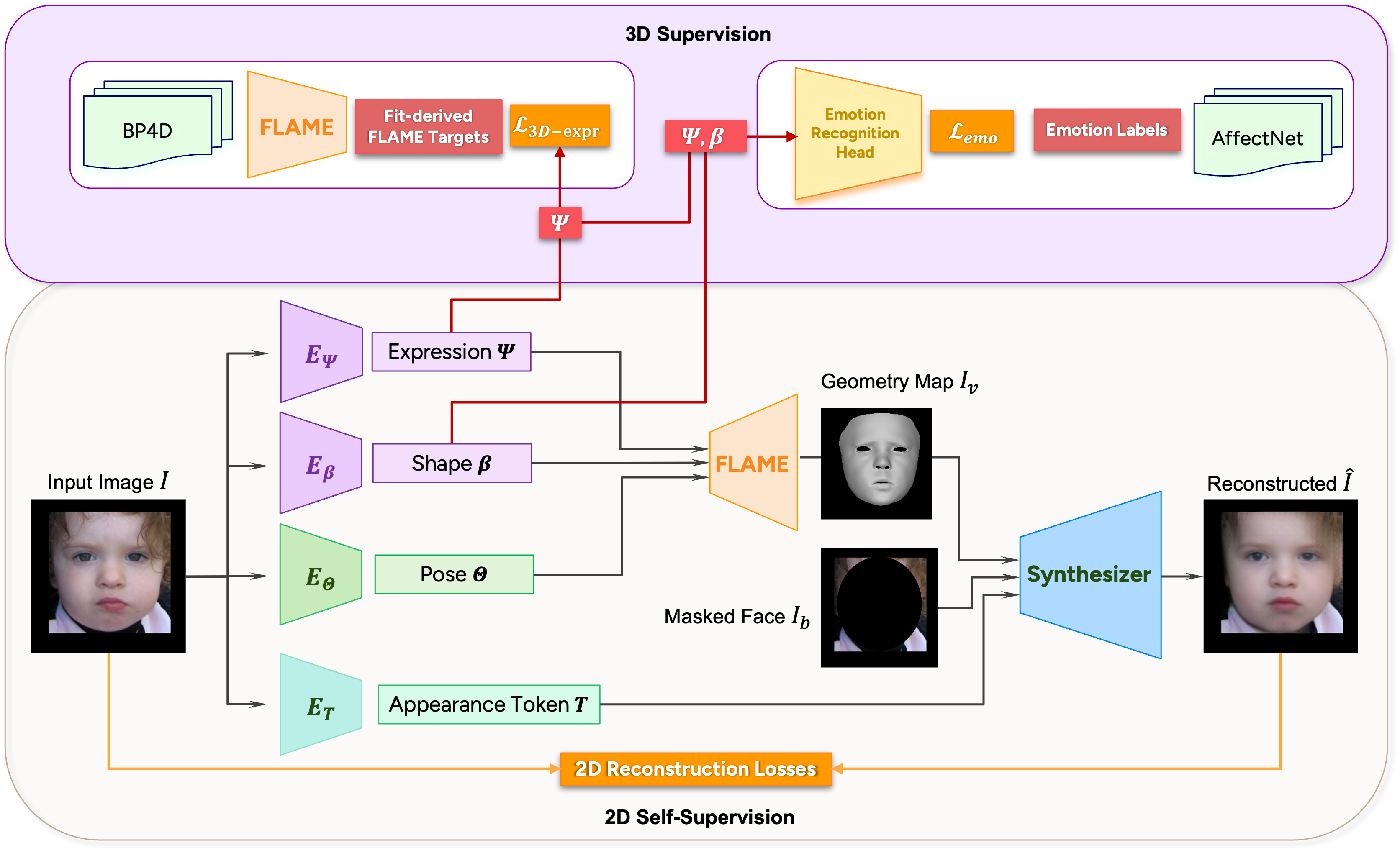}
  \caption{Illustration of the FIELDS pipeline.
  Firstly, given an input image $I$, encoders predict FLAME parameters (shape $\beta$, expression $\Psi$, pose $\Theta$) and an appearance token $T$. Then FLAME produces a face mesh model $V$ and Geometry Map $I_v$.
  Lastly, the Synthesizer fuses $I_v$, $T$, and Masked Face $I_b$ to generate the reconstructed image $\hat I$. Besides the inherited 2D consistency losses (at the bottom gray box),
  we introduce (i) fit-derived expression targets from BP4D used to supervise FLAME expression parameters via $\mathcal{L}_{\text{3D-expr}}$ (purple box, left),
  and (ii) an auxiliary emotion supervision to minimize the loss $\mathcal{L}_{\text{emo}}$ produced by the emotion head based on $(\Psi,\beta)$ with AffectNet labels (purple box, right).}
  \label{fig:pipeline}
\end{figure*}

We present FIELDS, a hybrid 2D/3D supervision framework that extends the TEASER~\cite{teaser} self-supervised monocular 3D face reconstruction pipeline with two complementary signals: (i) scan-derived (fit-derived) supervision on FLAME~\cite{FLAME} expression parameters from BP4D~\cite{bp4d} to anchor expression learning, and (ii) an auxiliary affect recognition head trained jointly to inject affect supervision directly in the 3DMM parameter space. As shown in \cref{fig:pipeline}, 
FIELDS consists of: (1) a FLAME-based encoder and neural synthesizer for estimating shape, pose, and expression; (2) a scan-derived expression anchoring loss on FLAME expression--jaw parameters; and (3) an auxiliary affect objective combining discrete emotion and valence--arousal supervision.

\subsection{Architecture}
\label{sec:pre}
\noindent\textbf{FLAME Model.}$\;$  We leverage the FLAME model \cite{FLAME} to parameterize 3D facial geometry as identity shape \(\beta\in\mathbb{R}^{300}\), expression \(\psi\in\mathbb{R}^{100}\) and pose \(\theta\in\mathbb{R}^{3k+3}\), where \(k=4\) is the number of joints (neck, jaw, left eyeball, and right eyeball) and the last 3 dimensions are the global head rotation. Given $\{\beta,\psi, \theta\}$, FLAME generates the 3D face mesh $V \in \mathbb{R}^{5023\times3}$ by $V = \mathrm{FLAME} (\beta, \psi, \theta).$

Following previous works \cite{teaser,smirk,emoca},
we use $\Psi = \{\psi_{exp},\psi_{eye},\theta_{jaw}\}$ to aggregate a group of parameters governing expression-related geometry,
where $\psi_{eye}\in\mathbb{R}^{2}$ denotes the eye closure blendshapes~\cite{mica}, $\theta_{jaw}\in\mathbb{R}^{3}$ represents the jaw rotation,
and $\psi_{exp}\in\mathbb{R}^{50}$ are the first 50 dimensions of the expression parameters.
As a result, with the subset $\theta_{head}\subset\theta$, we can formally define our FLAME model as:
\begin{align}
  V = \mathrm{FLAME} (\beta, \Psi, \theta_{head}).
\end{align}

\noindent\textbf{Encoder.}$\;$ Following the notation of SMIRK~\cite{smirk} and TEASER~\cite{teaser}, the image encoder $E(I)$ consists of four different branches, namely three geometry encoders $E_{\beta}(I)$, $E_{\Psi}(I)$, $E_{\Theta}(I)$ and one appearance token encoder $E_T(I)$.

Given the input image $I$, geometry encoders output the FLAME parameters, as well as the camera parameters $c\in\mathbb{R}^{3}$, here we group the global transformation parameters as $\Theta = \{c,\theta_{head}\}$. Thus, the geometry encoders are formally defined as:
\begin{align}
E_{\beta}(I) = \beta,\;E_{\Psi}(I) = \Psi,\; E_{\Theta}(I) = \Theta.
\end{align}

For appearance features, the multi-level CNN features $\{x_i\}_{i=1}^K$ are extracted from different stages of the token encoder $E_T$; we pool and project each via $
z_i = F_i\bigl(P(x_i)\bigr)\,,\quad i=1,\dots,K, $ then form the appearance token by channel-wise concatenation:
\begin{align}
E_T(I)\;=\; \bigl[z_1\;\odot\;z_2\;\odot\;\cdots\;\odot\;z_K\bigr]
\;=\; T,
\end{align}
here $P(\cdot)$ is the average pooling, $F_i(\cdot)$ is a fully connected layer, and $\odot$ denotes concatenation.
This yields a single appearance token $T$
that fuses both high-level semantics and
low-level detailed textures.

\noindent\textbf{Neural Synthesizer.}$\;$ We adopt TEASER’s Token-guided Face Synthesizer~\cite{teaser} as our neural synthesizer $\mathrm{G^{*}}$. It ingests three inputs: (1) the rasterized geometry map $I_v$, which is the monochrome rendering of the reconstructed face mesh $V$ and generated by a neural rasterization renderer; \;(2) the background $I_b$, which applies a mask function $M(I)$ to crop the face out and convey the background information; \;(3) the multi‐scale appearance token $T$. It produces the final image $\hat I$ via a lightweight U‐Net \cite{ronneberger2015u} structure model:
\begin{equation}
\begin{aligned}
  \hat I = \mathrm{G^{*}}\bigl(I_v \,\odot\, I_b \,\odot\, T\bigr).
\end{aligned}
\end{equation}
We use $I_v$ for geometric alignment, $I_b$ for background
consistency, and $T$ for semantic and appearance details.

\noindent\textbf{Emotion Recognition Head.}$\;$ Given the FLAME parameters predicted by the encoders \(\hat\beta, \hat\Psi\), we employ a lightweight 4-layer Multi-Layer Perceptron (MLP) with batch normalization~\cite{ioffe2015batch} and LeakyReLU \cite{xu2015empirical} activations to jointly predict continuous valence–arousal and discrete emotion categories:
\begin{equation}
\begin{aligned}
    \bigl[\hat v,\,\hat a,\,\mathbf{\hat p}\bigr]
  = \mathrm{MLP}_{\mathrm{emo}}(\hat\beta, \hat\Psi)\,,
\end{aligned}
\end{equation}
where \(\hat v,\hat a\in\mathbb{R}\) are the regressed valence-arousal scores, and \(\mathbf{\hat p}\) is the softmax probability over \(C=7\) emotion classes.

\subsection{Training with Hybrid 2D/3D Supervision}
\label{sec:opt}
\noindent\textbf{Scan-derived Expression Anchors.}$\;$
We leverage the BP4D~\cite{bp4d} 4D scan dataset by fitting each scan to the FLAME model by minimizing the scan-to-mesh distance~\cite{tfflame}. This fitting process yields scan-derived (fit-derived) expression parameters $\psi_{exp}$ for a wide range of spontaneous expression intensities.

\noindent\textbf{3D Facial Expression Loss.}$\;$ To directly supervise the expression encoder, we apply a mean squared error (MSE) loss between the predicted expression parameters \(\hat\psi_{exp}\) and scan-derived (fit-derived) targets \(\psi_{exp}\):
\begin{equation}
    \begin{aligned}
        \mathcal{L}_{\text{3D-expr}}
  = \bigl\lVert \hat\psi_{exp} - \psi_{exp}\bigr\rVert_{2}^{2}\,.
    \end{aligned}
\end{equation}
This loss encourages the encoder to align its FLAME expression parameters with the scan-derived reference, providing a geometric and intensity anchor that complements image-level self-supervision.

\noindent\textbf{Auxiliary Emotion Supervision.}$\;$
\label{sec:emoloss}
We train this emotion recognition head using a combination of regression and classification losses:
\begin{equation}
\begin{aligned}
  \mathcal{L}_{\mathrm{emo}}
  = \lambda_{r}\,\bigl(v - \hat v\bigr)^{2}
  + \lambda_{r}\,\bigl(a - \hat a\bigr)^{2}
  + \lambda_{c}\,\mathrm{CE}\bigl(\mathbf{p},\,\mathbf{\hat p}\bigr)\,,
\end{aligned}
\label{eq:Lemo}
\end{equation}
where \((v,a)\) are the ground-truth valence and arousal labels, \(\mathbf{p}\) is the discrete class label, \(\mathrm{CE}\) denotes cross-entropy, and \(\lambda_{r},\lambda_{c}\) balance the regression and classification terms.

The joint objective combines discrete emotion classification
with continuous valence--arousal regression.
The classification term encourages affect-discriminative
representations, while the VA regression terms encourage sensitivity
to continuous affect variation.
Together, they provide parameter--space affect supervision that
complements the geometric constraints imposed by scan-derived anchors.


\noindent\textbf{2D Consistency Supervision.}$\;$
\label{sec:2dcloss}
Following SMIRK~\cite{smirk} and TEASER~\cite{teaser}, both the encoder and synthesizer leverage a unified 2D consistency loss, denoted by \(\mathcal{L}_{\text{2D-cons}}\), which combines several pixel‐, perceptual- and landmark‐based terms. Concretely, it includes a \textbf{photometric loss} \(\mathcal{L}_{\text{photo}} = \lVert I - \hat I\rVert_{1}\), a \textbf{perceptual (VGG) loss} \(\mathcal{L}_{\text{perc}} = \lVert \Gamma(I) - \Gamma(\hat I)\rVert_{1}\) with the VGG feature extractor \cite{johnson2016perceptual} \(\Gamma(\cdot)\), and a \textbf{standard landmark loss}
\(\mathcal{L}_{\text{lmk}} = \sum_{i=1}^{K}\lVert \mathbf{k}_i - \mathbf{k}_i'\rVert_{2}^{2}\),
where \(\mathbf{k}_i\) are the detected 2D landmarks and \(\mathbf{k}_i'\) their projections from the reconstructed mesh. To handle yaw variations \(\theta_y\), a \textbf{pose‐dependent landmark loss}
\(\mathcal{L}_{\text{pdl}} = \lVert M_L(\theta_y)\,L - M_L(\theta_y)\,L_v\rVert_{2}^{2}\) is also applied, where \(L\) denotes detected 2D landmarks, \(L_v\) denotes the corresponding landmarks projected from the reconstructed mesh, and \(M_L(\theta_y)\) is the visibility mask~\cite{teaser}.
Finally, it focuses on expression‐sensitive regions by parsing mouth and eye masks \(M_m\) and \(M_e\) and computing a \textbf{region loss}
\(\mathcal{L}_{\text{rg}} = \lVert (M_m + M_e)\odot I - (M_m + M_e)\odot \hat I\rVert_{2}^{2}\). For compactness, fixed per-term weights are assembled in a coefficient vector $\mathbf{w}_{\text{2D}}$, and the 2D consistency loss $\mathcal{L}_{\text{2D-cons}}$ is obtained by its dot product ($\langle \cdot, \cdot \rangle$) with the loss vector:
\begin{equation}
\begin{aligned}
\mathcal{L}_{\text{2D-cons}}=\langle \mathbf{w}_{\text{2D}},\,[\mathcal{L}_{\text{photo}},\mathcal{L}_{\text{perc}},\mathcal{L}_{\text{lmk}},\mathcal{L}_{\text{pdl}},\mathcal{L}_{\text{rg}}]\rangle.
\end{aligned}
\label{eq:2dloss}
\end{equation}
This unified loss is then used in Eqs.~\eqref{eq:encoder-loss} and \eqref{eq:renderer-loss}.

\noindent\textbf{Regularization.}$\;$
\label{sec:reg}
Before joint training, we pretrain the geometry encoders $B$ as a base model by aligning their predictions to MICA~\cite{mica} shape estimates, which provides a stable initialization of the parameter manifold~\cite{smirk}. Then we adopt the frozen base model $B$ for parameter-space regularization: the encoder’s predicted parameters are constrained to be close to the corresponding outputs of $B$. Let $[B(I)]_{\beta}$ and $[B(I)]_{\psi_{\text{expr}}}$ denote the base model’s shape and expression parameters of image $I$.
We define the \textbf{shape regularizer} as \(\mathrm{Reg}_{\text{shape}}=\| \hat{\beta} -\operatorname{sg}\!\bigl([B(I)]_{\beta}\bigr)\bigr\|_2^2\) and the \textbf{expression regularizer} as \(\mathrm{Reg}_{\text{expr}}=\| \hat\psi_{exp}  - \operatorname{sg}\!\bigl([B(I)]_{\psi_{exp}}\bigr)\bigr\|_2^2\), where $\operatorname{sg}(\cdot)$ indicates stop-gradient on the base model outputs (i.e., gradients do not flow into $B$).
With the hyperparameters $\lambda_{\text{shape-reg}}$ and $\lambda_{\text{expr-reg}}$ that weight the shape- and expression-regularization terms, the total regularizer is:
\begin{equation}
\mathcal{L}_{\text{reg}}
=\lambda_{\text{shape-reg}}\,\mathrm{Reg}_{\text{shape}}
+\lambda_{\text{expr-reg}}\,\mathrm{Reg}_{\text{expr}}.
\label{eq:rg}
\end{equation}

\subsection{Training Procedure}
\label{sec:train}
We alternate two training steps per iteration -- an Encoder Pass and a Synthesizer Pass -- to prevent one component from compensating the other’s errors.

\textbf{Encoder Pass.}$\;$
We freeze the pretrained pose encoder and train the remaining learnable modules (shape $E_{\beta}$, expression $E_{\Psi}$, and the auxiliary affect head) with
\begin{equation}
\mathcal{L}_{E}
= \lambda_{\text{2D}}\,\mathcal{L}_{\text{2D-cons}}
+ \lambda_{\text{3D}}\,\mathcal{L}_{\text{3D-expr}}
+ \lambda_{\text{emo}}\,\mathcal{L}_{\text{emo}}
+ \mathcal{L}_{\text{reg}}.
\label{eq:encoder-loss}
\end{equation}

\textbf{Synthesizer Pass.}$\;$
During the synthesizer pass, we keep all encoders fixed and optimize only the synthesizer with:
\begin{equation}
\mathcal{L}_{G^{*}}
= \lambda_{\text{2D}}\,\mathcal{L}_{\text{2D-cons}}
+ \lambda_{\text{cyc}}\,\mathcal{L}_{\text{cyc}}
+ \lambda_{\text{tok}}\,\mathcal{L}_{\text{tok-cyc}}.
\label{eq:renderer-loss}
\end{equation}
Here $\mathcal{L}_{\text{cyc}}$ enforces cycle consistency on expression parameters recovered from the rendered image, and $\mathcal{L}_{\text{tok-cyc}}$ preserves appearance tokens across render–reencode.

\begin{table*}[t]
  \centering
  \scriptsize
  \setlength{\tabcolsep}{3.2pt}
  \renewcommand{\arraystretch}{1.03}
  \caption{Valence--arousal regression (5-fold) on AffectNet.
  Pearson correlation coefficient (PCC$\uparrow$), concordance
  correlation coefficient (CCC$\uparrow$), root mean squared error
  (RMSE$\downarrow$), and sign agreement (SAGR$\uparrow$) are reported
  as means with superscripted standard deviations
  (\textsuperscript{\scriptsize$\pm$}).
  The best and second-best mean results per column are shown in
  \textbf{bold} and \underline{underlined}.}
  \label{tab:v_a_merged}

  \resizebox{\textwidth}{!}{%
  \begin{tabular}{lcccccccc}
    \toprule
    Model
      & V-PCC$\uparrow$ & V-CCC$\uparrow$
      & V-RMSE$\downarrow$ & V-SAGR$\uparrow$
      & A-PCC$\uparrow$ & A-CCC$\uparrow$
      & A-RMSE$\downarrow$ & A-SAGR$\uparrow$ \\
    \midrule

    EMOCA~\cite{emoca}
      & \msup{\underline{0.733}}{0.008}
      & \msup{\underline{0.721}}{0.007}
      & \msup{\underline{0.343}}{0.008}
      & \msup{\underline{0.790}}{0.005}
      & \msup{\underline{0.591}}{0.014}
      & \msup{\underline{0.580}}{0.014}
      & \msup{\underline{0.361}}{0.008}
      & \msup{\underline{0.763}}{0.005} \\

    SMIRK~\cite{smirk}
      & \msup{0.702}{0.004}
      & \msup{0.690}{0.001}
      & \msup{0.363}{0.005}
      & \msup{0.781}{0.005}
      & \msup{0.551}{0.004}
      & \msup{0.538}{0.004}
      & \msup{0.376}{0.003}
      & \msup{0.745}{0.004} \\

    TEASER~\cite{teaser}
      & \msup{0.724}{0.005}
      & \msup{0.715}{0.005}
      & \msup{0.348}{0.004}
      & \msup{0.787}{0.003}
      & \msup{0.583}{0.006}
      & \msup{0.573}{0.009}
      & \msup{0.366}{0.006}
      & \msup{0.757}{0.009} \\

    \midrule

    FIELDS
      & \bmsup{0.747}{0.003}
      & \bmsup{0.737}{0.006}
      & \bmsup{0.338}{0.006}
      & \bmsup{0.801}{0.002}
      & \bmsup{0.616}{0.008}
      & \bmsup{0.604}{0.008}
      & \bmsup{0.352}{0.011}
      & \bmsup{0.764}{0.008} \\

    \bottomrule
  \end{tabular}%
  }
\end{table*}

Alternating Eqs.~\eqref{eq:encoder-loss} and~\eqref{eq:renderer-loss} stabilizes training and reduces appearance compensation in the synthesizer while keeping identity-consistent 3DMM parameters.
Additional training details are provided in the supplementary.

\section{Experiments}

\subsection{Dataset}

We train on six public datasets: LRW~\cite{lrw}, CelebA~\cite{celeba}, FFHQ~\cite{ffhq}, MEAD~\cite{mead}, AffectNet~\cite{affectnet}, and BP4D~\cite{bp4d}.
AffectNet provides discrete emotion and continuous valence--arousal labels; we follow the AffectNet-7 protocol by removing \textit{Contempt} and use its public class-balanced validation split only for final evaluation. RAF-DB~\cite{li2017reliablerafdb,li2019reliablerafdb} is used only for external cross-dataset affective utility evaluation.
BP4D provides spontaneous-expression 3D scans, which is fitted to the FLAME model to obtain the FLAME parameters as the ground-truth.
We use a subject-independent evaluation protocol for it.
Additional details including the train/test split for BP4D are provided in the supplementary.

\subsection{Evaluation}
We evaluate the affective utility and geometric fidelity of
the learned representations through three complementary analyses:
(1) \textbf{Affective utility}, assessed by categorical emotion classification
and continuous valence--arousal regression on AffectNet, together with within- and
cross-dataset RAF-DB KNN-based classification using expression--jaw parameters;
(2) \textbf{Geometric fidelity}, assessed by rigid- and
similarity-aligned vertex errors and FLAME parameter MSE against scan-fitted references on held-out BP4D subjects; and
(3) \textbf{Qualitative comparisons}, examining expression
consistency and facial geometry on AffectNet and BP4D.

\begin{table}[t]
  \centering
  \scriptsize
  \setlength{\tabcolsep}{3.5pt}
  \renewcommand{\arraystretch}{1.25}

  \caption{Emotion classification (5-fold) on AffectNet.
  Accuracy$\uparrow$, macro-averaged precision$\uparrow$, and
  macro-F1$\uparrow$ are reported as means with superscripted standard
  deviations (\textsuperscript{\scriptsize$\pm$}), in percentages.
  The best and second-best mean results per column are shown in
  \textbf{bold} and \underline{underlined}.}
  \label{tab:emotion_classification_pct}

  \resizebox{\columnwidth}{!}{%
  \begin{tabular}{lccc}
    \toprule
    Model
      & Accuracy(\%)$\uparrow$
      & Precision(\%)$\uparrow$
      & F1(\%)$\uparrow$ \\
    \midrule

    EMOCA~\cite{emoca}
      & \meanstd{52.96}{0.62}
      & \meanstd{56.97}{0.31}
      & \meanstd{52.49}{0.65} \\

    SMIRK~\cite{smirk}
      & \meanstd{48.69}{0.62}
      & \meanstd{53.53}{0.46}
      & \meanstd{48.03}{0.91} \\

    TEASER~\cite{teaser}
      & \umeanstd{55.27}{0.51}
      & \umeanstd{57.74}{0.67}
      & \umeanstd{55.05}{0.71} \\

    \midrule

    FIELDS
      & \bmeanstd{55.99}{1.45}
      & \bmeanstd{59.25}{0.59}
      & \bmeanstd{55.56}{1.58} \\

    \bottomrule
  \end{tabular}%
  }
\end{table}

\noindent\textbf{Affective utility.}
Following EMOCA~\cite{emoca}, we evaluate the affective information
preserved by each reconstruction through a downstream predictor. For
AffectNet, each reconstruction model is frozen, and its predicted FLAME
identity and expression parameters are used to train a lightweight
4-layer MLP for emotion classification and valence--arousal (VA)
regression. Since the official AffectNet test set is unavailable, we use
the official training split for evaluator training and 5-fold model
selection, and reserve the public class-balanced validation split only
for final testing. 


For this MLP evaluation, all methods use the same feature definition, folds, evaluator architecture, and final test split. As shown in \cref{tab:emotion_classification_pct,tab:v_a_merged}, FIELDS achieves the best result on all eleven metrics for AffectNet emotion classification and valence--arousal regression. Arousal is the more challenging dimension, as indicated by the consistently lower concordance correlation coefficient (CCC) scores across all methods. FIELDS nevertheless achieves its largest CCC gain on arousal, suggesting that its predicted FLAME parameters better preserve facial information relevant to continuous arousal estimation.

To assess the affective information carried by expression-related parameters without relying on identity-shape parameters or a learned MLP evaluator, we perform KNN classification on RAF-DB using only expression--jaw parameters. We evaluate both within- and cross-dataset settings on
RAF-DB. The former measures emotion discrimination within the target
dataset, whereas the latter tests whether this structure remains useful
when the reference features come from another dataset.

As shown in \cref{tab:cross_dataset_combined}, FIELDS ranks the first in all four metrics. In the within-dataset setting, it outperforms EMOCA, a leading affect-aware 3D face reconstruction method, by $5.2$ percentage points (pp) in accuracy and $5.4$ pp in macro-F1. Under cross-dataset evaluation, FIELDS remains the best-performing method, leading EMOCA by $4.4$ and $6.7$ pp in accuracy and macro-F1, respectively. This suggests that the affective information retained in its predicted parameters generalizes more effectively across datasets. The larger macro-F1 margin further indicates that the gain extends beyond the dominant classes and yields more balanced recognition across emotion categories. Together with the AffectNet MLP results, these findings show that the affective advantage of FIELDS persists across learned and non-parametric evaluators and across datasets. Detailed KNN protocols and additional analyses, including per-class accuracy, confusion matrices, and embedding visualizations, are provided in the supplementary material.
\begin{table}[t]
\centering
\scriptsize
\setlength{\tabcolsep}{3.2pt}
\renewcommand{\arraystretch}{0.95}

\caption{Within- and cross-dataset KNN emotion classification on RAF-DB.
The reference database is constructed from the training split named
before the arrow, and queries are taken from the official RAF-DB test
split. Accuracy$\uparrow$ and macro-F1$\uparrow$ are reported as
percentages. The best and second-best mean results per column are shown
in \textbf{bold} and \underline{underlined}.}
\label{tab:cross_dataset_combined}

\resizebox{\columnwidth}{!}{%
\begin{tabular}{lcc|cc}
\toprule
& \multicolumn{2}{c|}{\textbf{RAF-DB}$\rightarrow$\textbf{RAF-DB}}
& \multicolumn{2}{c}{\textbf{AffectNet}$\rightarrow$\textbf{RAF-DB}} \\
\cmidrule(lr){2-3}
\cmidrule(lr){4-5}

Model
& Acc(\%)$\uparrow$ & F1(\%)$\uparrow$
& Acc(\%)$\uparrow$ & F1(\%)$\uparrow$ \\
\midrule

EMOCA~\cite{emoca}
& \underline{67.8} & \underline{54.1}
& \underline{60.5} & \underline{38.2} \\

SMIRK~\cite{smirk}
& 59.6 & 41.1
& 53.0 & 29.5 \\

TEASER~\cite{teaser}
& 62.4 & 45.6
& 49.1 & 26.2 \\

\midrule

FIELDS
& \textbf{73.0} & \textbf{59.5}
& \textbf{64.9} & \textbf{44.9} \\

\bottomrule
\end{tabular}%
}
\end{table}

\begin{table*}[t]
  \centering
  \footnotesize
  \setlength{\tabcolsep}{3pt}
  \renewcommand{\arraystretch}{1.12}

  \caption{3D fidelity evaluation on BP4D against scan-fitted FLAME
  references. We report FLAME parameter MSE and frame-weighted mean
  corresponding-vertex errors (mm) under rigid and similarity alignment.
  Full and Eye denote the full mesh and eye region, respectively.
  Lower is better; best and second-best results per column are
  \textbf{bold} and \underline{underlined}.}
  \label{tab:bp4d_geom_merged}
  \resizebox{\textwidth}{!}{%
  \begin{tabular}{@{}lcc@{\hspace{9pt}}*{6}{c}
                         @{\hspace{9pt}}*{6}{c}@{}}
    \toprule
    & \multicolumn{2}{c}{\textbf{FLAME param MSE} $\downarrow$}
    & \multicolumn{6}{c}{\textbf{Rigid error (mm)} $\downarrow$}
    & \multicolumn{6}{c}{\textbf{Similarity error (mm)} $\downarrow$} \\
    \cmidrule(lr){2-3}
    \cmidrule(lr){4-9}
    \cmidrule(lr){10-15}

    Model
    & $\{\psi_{\mathrm{exp}},\theta_{\mathrm{jaw}}\}$ & $\beta$
    & Full & Face & Lips & Eye & Forehead & Nose
    & Full & Face & Lips & Eye & Forehead & Nose \\
    \midrule

    EMOCA~\cite{emoca}
    & \underline{1.273} & --
    & 6.64 & 5.71 & 9.33
    & \underline{3.99} & \underline{5.05} & \underline{4.61}
    & 4.30 & 4.14 & 7.18
    & \underline{2.53} & 4.26 & \underline{3.81} \\

    SMIRK~\cite{smirk}
    & 1.277 & \underline{0.426}
    & \underline{6.15} & \underline{5.03} & \textbf{5.87}
    & 4.20 & 5.10 & 5.36
    & 4.34 & 4.01 & 6.45
    & 2.70 & 3.82 & 4.03 \\

    TEASER~\cite{teaser}
    & 3.345 & 0.431
    & 6.35 & 5.25 & 6.55
    & 4.22 & 5.16 & 5.83
    & \underline{4.24} & \underline{3.81} & \underline{6.01}
    & 2.59 & \underline{3.60} & 4.00 \\

    \midrule
    FIELDS
    & \textbf{0.173} & \textbf{0.349}
    & \textbf{5.72} & \textbf{4.55} & \underline{6.53}
    & \textbf{3.45} & \textbf{4.79} & \textbf{4.46}
    & \textbf{3.88} & \textbf{3.33} & \textbf{5.91}
    & \textbf{2.13} & \textbf{3.09} & \textbf{3.28} \\

    \bottomrule
  \end{tabular}
  }
\end{table*}
\begin{figure*}[t]
  \centering
  \includegraphics[width=\textwidth]{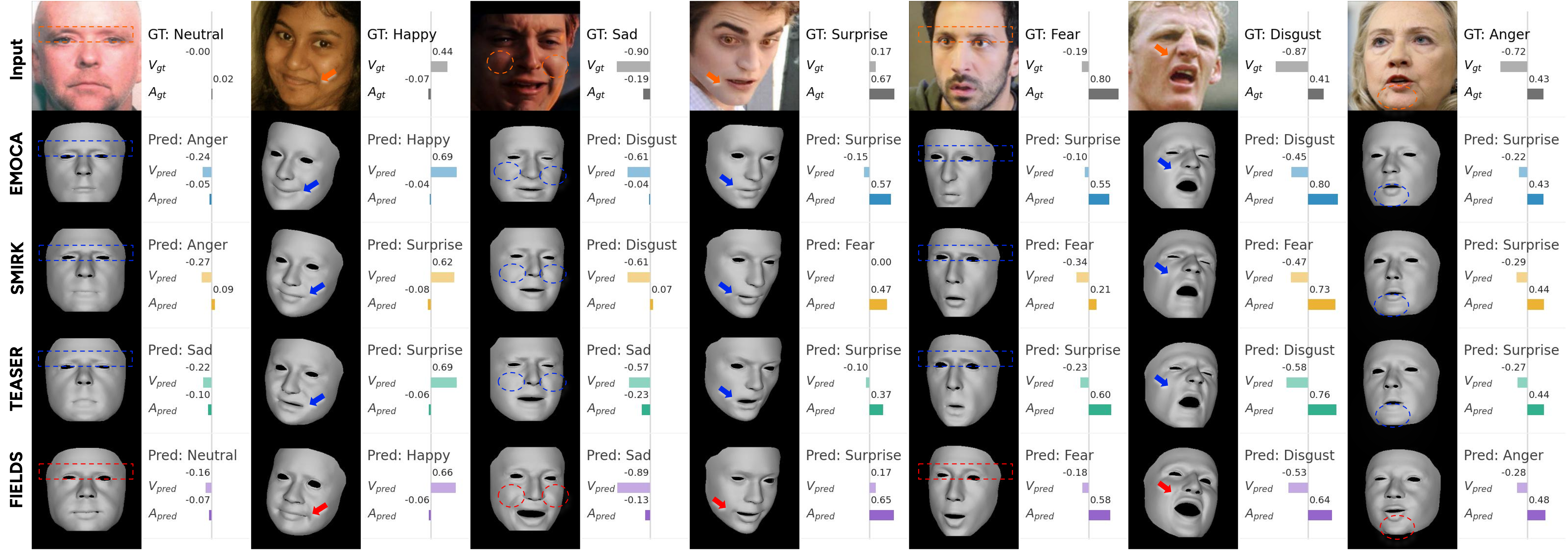}
\caption{Qualitative comparison on AffectNet with ground-truth annotations and per-method predictions. Predictions are produced by the same predictor trained under the EMOCA protocol. Top row: input image with discrete label and valence/arousal (\(V_{gt}, A_{gt}\)).
Rows below: rendered reconstructions from EMOCA~\cite{emoca}, SMIRK~\cite{smirk}, TEASER~\cite{teaser}, and FIELDS, each annotated with its predicted label and valence/arousal (\(V_{pred}, A_{pred}\)).      Highlighted regions indicate localized facial cues discussed in the text.
}
  \label{fig:3Drenderedface}
\end{figure*}

\noindent\textbf{Geometric fidelity.}
We evaluate reconstruction accuracy on six held-out BP4D subjects, with one subject from each race--gender group defined in our split. Training anchors are derived exclusively from training subjects,
enabling evaluation on unseen identities. All methods are evaluated on the same test samples against the common scan-fitted FLAME references. Tab.~\ref{tab:bp4d_geom_merged} reports FLAME parameter MSE together with mean corresponding-vertex errors for the full mesh and five facial regions, namely face, lips, eyes, forehead, and nose, defined by FLAME \cite{FLAME}.

We report the per-vertex-error (pve) between the predicted and ground-truth meshes after rigid and similarity alignments.
Rigid alignment removes global rotation and translation while preserving scale, whereas similarity alignment additionally removes global scale differences through a positive uniform scale per frame. FIELDS achieves the lowest error on
full-mesh and all regions under both alignment protocols except rigid-aligned lips. Compared with EMOCA, a leading affect-aware reconstruction method, FIELDS reduces lip-region error by $30.0\%$ under rigid alignment and $17.7\%$ under similarity alignment. This highlights improved geometric accuracy in a region central to facial expression, with corresponding visual examples discussed in the qualitative analysis later.

FIELDS also yields the lowest expression--jaw and identity--shape parameter MSE among the selected comparison methods.
These metrics provide complementary evidence of agreement with the fitted parameter targets. Detailed alignment formulations, region definitions, and complete results under raw, centered, rigid, and similarity evaluation are provided in the supplementary material.

\noindent\textbf{Qualitative comparisons.}
\label{sec:quali}
Fig.~\ref{fig:3Drenderedface} illustrates affect prediction
and reconstructed expressions on AffectNet. FIELDS correctly classifies all seven displayed examples, including the neutral and anger examples
misclassified by all three comparison methods. Its valence–arousal predictions are also closer to the ground truth of the sad and surprise examples. In the happy example, both EMOCA and FIELDS predict the correct category, but EMOCA produces a more pronounced mouth-corner lift than is visible in the input, whereas FIELDS better preserves the subtle smile, which is more consistent with the original image. This illustrates that correct emotion classification can coexist with exaggerated facial deformation, highlighting the importance of preserving expression geometry alongside affective information.

\begin{figure*}[t]
  \centering
  \includegraphics[width=0.935\linewidth]{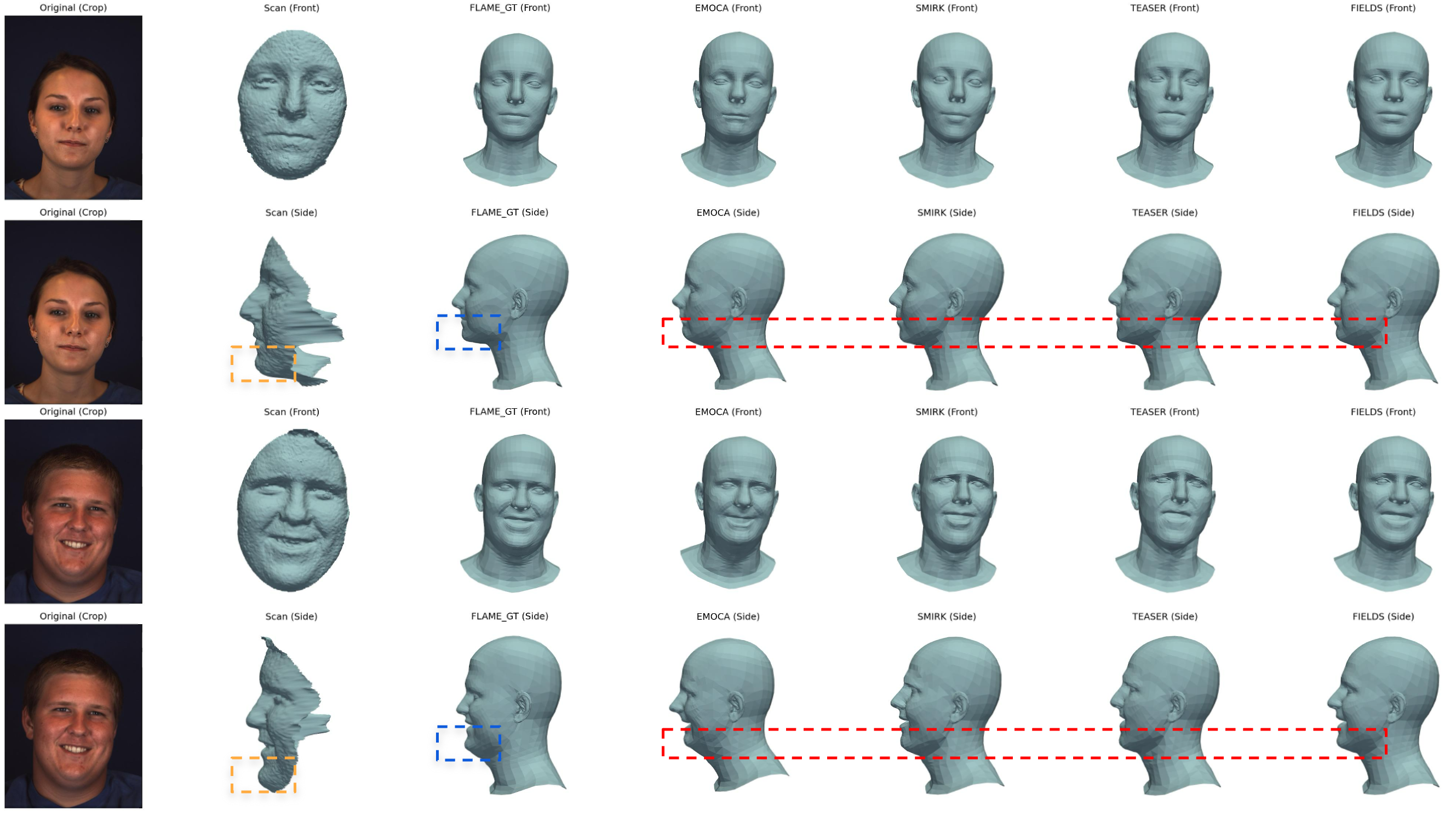}
  \caption{Qualitative comparison on BP4D. For each subject, we show the input RGB frame, the corresponding 3D scan, the scan-derived FLAME reference (fitted), and reconstructions from EMOCA~\cite{emoca}, SMIRK~\cite{smirk}, TEASER~\cite{teaser}, and FIELDS, rendered from both front and side views. Highlighted regions indicate areas discussed in the text.}
  \label{fig:bp4d}
\end{figure*}

Fig.~\ref{fig:bp4d} examines geometric fidelity against BP4D scans. The fitted FLAME references closely follow the visible facial surface and profile of the scans, supporting their use as geometric supervision targets. The highlighted side views further show that FIELDS reconstructs the chin and jaw contour more faithfully than the selected comparison methods, with closer agreement to both the fitted references and the scans. These observations are consistent with the
quantitative geometric improvements reported in
Tab.~\ref{tab:bp4d_geom_merged}.

In summary, these results show that FIELDS learns FLAME representations that support stronger affect prediction across evaluators and datasets while achieving better geometric accuracy. The qualitative examples further illustrate how these improvements are reflected in the faithfulness of the generated facial expression and geometry to the original image. This supports the joint pursuit of affective utility and geometric fidelity in face reconstruction.

\begin{table*}[!htbp]
  \centering
  \scriptsize
  \setlength{\tabcolsep}{2.5pt}
  \renewcommand{\arraystretch}{1.0}

  \caption{Ablation of the proposed supervision signals.
  Affective performance is reported as
  mean\textsuperscript{\scriptsize$\pm$}std over five folds;
  geometric evaluation reports FLAME parameter MSE and
  rigid-/similarity-aligned full-mesh errors (mm) on BP4D.
  The best and second-best mean results per column are shown in
  \textbf{bold} and \underline{underlined}.}
  \label{tab:unified_ablation_eccv}

  \resizebox{\textwidth}{!}{%
  \begin{tabular}{l c cccc cc cc}
    \toprule
    & \multicolumn{1}{c}{\textbf{Emo Cls.}}
    & \multicolumn{4}{c}{\textbf{V/A Regression}}
    & \multicolumn{2}{c}{\textbf{FLAME MSE} $\downarrow$}
    & \multicolumn{2}{c}{\textbf{3D errors} (mm) $\downarrow$} \\
    \cmidrule(lr){2-2}
    \cmidrule(lr){3-6}
    \cmidrule(lr){7-8}
    \cmidrule(lr){9-10}

    Model
      & Acc (\%) $\uparrow$
      & V-CCC $\uparrow$
      & V-RMSE $\downarrow$
      & A-CCC $\uparrow$
      & A-RMSE $\downarrow$
      & $\{\psi_{exp},\theta_{jaw}\}$
      & $\beta$
      & Rigid
      & Similarity \\
    \midrule

    F0: Pre-train
      & \meanstd{47.68}{0.75}
      & \msup{0.647}{0.008}
      & \msup{0.384}{0.005}
      & \msup{0.528}{0.012}
      & \msup{0.378}{0.009}
      & 0.678
      & 0.357
      & 5.739
      & 4.045 \\

    F1: Baseline
      & \meanstd{52.25}{0.67}
      & \msup{0.686}{0.011}
      & \msup{0.367}{0.006}
      & \msup{0.560}{0.009}
      & \msup{0.366}{0.004}
      & 5.954
      & \underline{0.354}
      & 6.250
      & 4.532 \\

    F2: F1+$\mathcal{L}_{\text{3D-expr}}$
      & \meanstd{51.81}{0.67}
      & \msup{0.705}{0.010}
      & \msup{0.353}{0.006}
      & \msup{0.568}{0.006}
      & \msup{0.367}{0.005}
      & \textbf{0.159}
      & \underline{0.354}
      & \textbf{5.474}
      & \textbf{3.872} \\

    F3: F1+$\mathcal{L}_{\text{emo}}$
      & \umeanstd{54.28}{1.56}
      & \bmsup{0.740}{0.009}
      & \bmsup{0.336}{0.005}
      & \bmsup{0.609}{0.009}
      & \bmsup{0.348}{0.006}
      & 3.161
      & 0.354
      & 6.128
      & 4.418 \\

    Alt.: F1+$\mathcal{L}_{\text{img-emo}}$
      & \meanstd{52.7}{0.82}
      & \msup{0.714}{0.003}
      & \msup{0.351}{0.005}
      & \msup{0.579}{0.010}
      & \msup{0.363}{0.004}
      & 1.632
      & 0.360
      & 5.857
      & 4.143 \\

    \midrule

    F4: F2+F3
      & \bmeanstd{55.99}{1.45}
      & \msup{\underline{0.737}}{0.006}
      & \msup{\underline{0.338}}{0.006}
      & \msup{\underline{0.604}}{0.008}
      & \msup{\underline{0.352}}{0.011}
      & \underline{0.173}
      & \textbf{0.349}
      & \underline{5.720}
      & \underline{3.884} \\

    \bottomrule
  \end{tabular}%
  }
\end{table*}

\subsection{Ablation Study}

Tab.~\ref{tab:unified_ablation_eccv} evaluates scan-derived anchoring
(F2), parameter-space affect supervision (F3), and their combination
(F4) against baseline F1. Alt.\ combines EMOCA-style image-level affect
supervision~\cite{emoca} with F1.
Anchoring yields the lowest expression--jaw MSE ($0.159$) and
geometric errors ($5.474$/$3.872$\,mm) for rigid/similarity alignment.
It also reduces both geometric errors when supplemented with affect
supervision (F3$\rightarrow$F4), supporting its role in constraining
reconstructed geometry.

Firstly, F3 achieves the best VA regression results and the second-best classification, demonstrating the effectiveness of the auxiliary emotion supervision. Compared with Alt., it provides stronger affective performance but higher geometric errors. Moreover, F2 achieves the lowest geometric errors with both rigid and similarity alignments. It also achieves the best accuracy on the FLAME expression--jaw parameters decoding. Besides, F4 achieves the highest classification accuracy with the second-best VA regression performance. Additionally, F4 also outperforms Alt.\ on all reported affective metrics and both geometric errors. In summary, these results support the complementary
roles of geometric anchoring and affect supervision:
F2 favors geometric accuracy, while F4 offers stronger joint
performance.

\section{Conclusions}
FIELDS brings geometric anchoring and affect supervision together
to learn expression representations within the FLAME parameter space.
Our results show that stronger affective utility can accompany
improved geometric fidelity, with the two supervision signals making
complementary contributions.
The resulting expression representation retain affect-discriminative information
across datasets while remaining directly connected to facial geometry.
This connection gives the learned representation value for both
affect analysis and further investigation of controllable facial
expression synthesis. Future work will explore temporal expression modeling and
affect-guided expression editing.

\clearpage
\ifFGfinal
\paragraph*{Acknowledgements.}
This work was supported by the Wallenberg AI, Autonomous Systems and Software Program (WASP) funded by the Knut and Alice Wallenberg Foundation, and the Swedish e-Science Research Centre (SeRC). Some of the computations were enabled by
the supercomputing resources Berzelius and LUMI provided by the National Supercomputer Centre at Linköping University and the Knut and Alice Wallenberg foundation.

\fi
{
    \small
    \bibliographystyle{ieee}
    \bibliography{main}
}

\clearpage
\setcounter{page}{1}
\maketitlesupplementary
\ifarxiv
\thispagestyle{plain}
\fi

This supplementary material provides additional details, experiments, and visualizations that support the results presented in the main paper.
\begin{itemize}
\item Section~\ref{suppA} describes additional implementation and dataset details, including data preprocessing, training settings, and the BP4D fitting procedure used to obtain pseudo ground-truth supervision. It also defines the BP4D geometric evaluation protocols and facial-region masks.

\item Section~\ref{suppB} provides additional analyses of the model behavior on the AffectNet validation set, including per-class accuracy and confusion matrix analysis to better understand the category-wise performance and error patterns of different methods. It also reports a non-parametric KNN evaluation on AffectNet and the complete BP4D full-mesh results under raw, centered, rigid, and similarity alignment.

\item Section~\ref{suppC} presents an ablation study analyzing the sensitivity of the model to the emotion loss weight.

\item Section~\ref{suppD} provides additional qualitative comparisons and visualizations of reconstructed meshes on different datasets, including representative examples and failure cases.

\item Section~\ref{suppE} presents additional embedding visualizations to further illustrate the structure of the learned expression representations.

\end{itemize}

\section{Implementation and Dataset Details}
\label{suppA}

\subsection{Datasets}
\label{suppA_datasets}
The proposed method is trained on six publicly available datasets:
LRW~\cite{lrw}, CelebA~\cite{celeba}, FFHQ~\cite{ffhq},
MEAD~\cite{mead}, AffectNet~\cite{affectnet}, and BP4D~\cite{bp4d}.
RAF-DB~\cite{li2017reliablerafdb,li2019reliablerafdb} is used only for
external evaluation.
\noindent\textbf{LRW} contains over 500,000 short video clips (each video has 29 frames, totaling 1.6 seconds) of 500 target words spoken by hundreds of different speakers. 
\noindent\textbf{CelebA} comprises 202,599 face images of 10,177 identities under diverse poses, expressions, and backgrounds.
\noindent\textbf{FFHQ} offers 70,000 high-quality, 1024×1024 resolution face images sampled from Flickr albums. It covers a wide range of ages, ethnicities, and backgrounds.
\noindent\textbf{MEAD} provides over 42,000 video clips of 60 actors uttering scripted sentences under six basic emotions in the lab environment.
\noindent\textbf{BP4D} is a 4D facial expression dataset with synchronized high-resolution RGB video and 3D dynamic scans from 41 participants performing eight emotion-elicitation tasks.
\noindent\textbf{AffectNet} is drawn from ``in the wild'' photos sourced from the Web. This subset of 291,651 facial images was manually annotated with one of eight discrete emotion categories (labels 0–7) as well as Valence/Arousal values.
\noindent\textbf{RAF-DB} is an in-the-wild facial-expression dataset.
We use its Basic Emotion subset, which contains $12{,}271$ official
training images and $3{,}068$ official test images across seven emotion
categories. RAF-DB is not used to train or select the reconstruction
models. In the within-dataset KNN evaluation, its training split provides
the labeled reference features and its test split provides the queries.

\begin{figure}[!htbp]
  \centering
  \capstart 
  \includegraphics[width=0.43\textwidth]{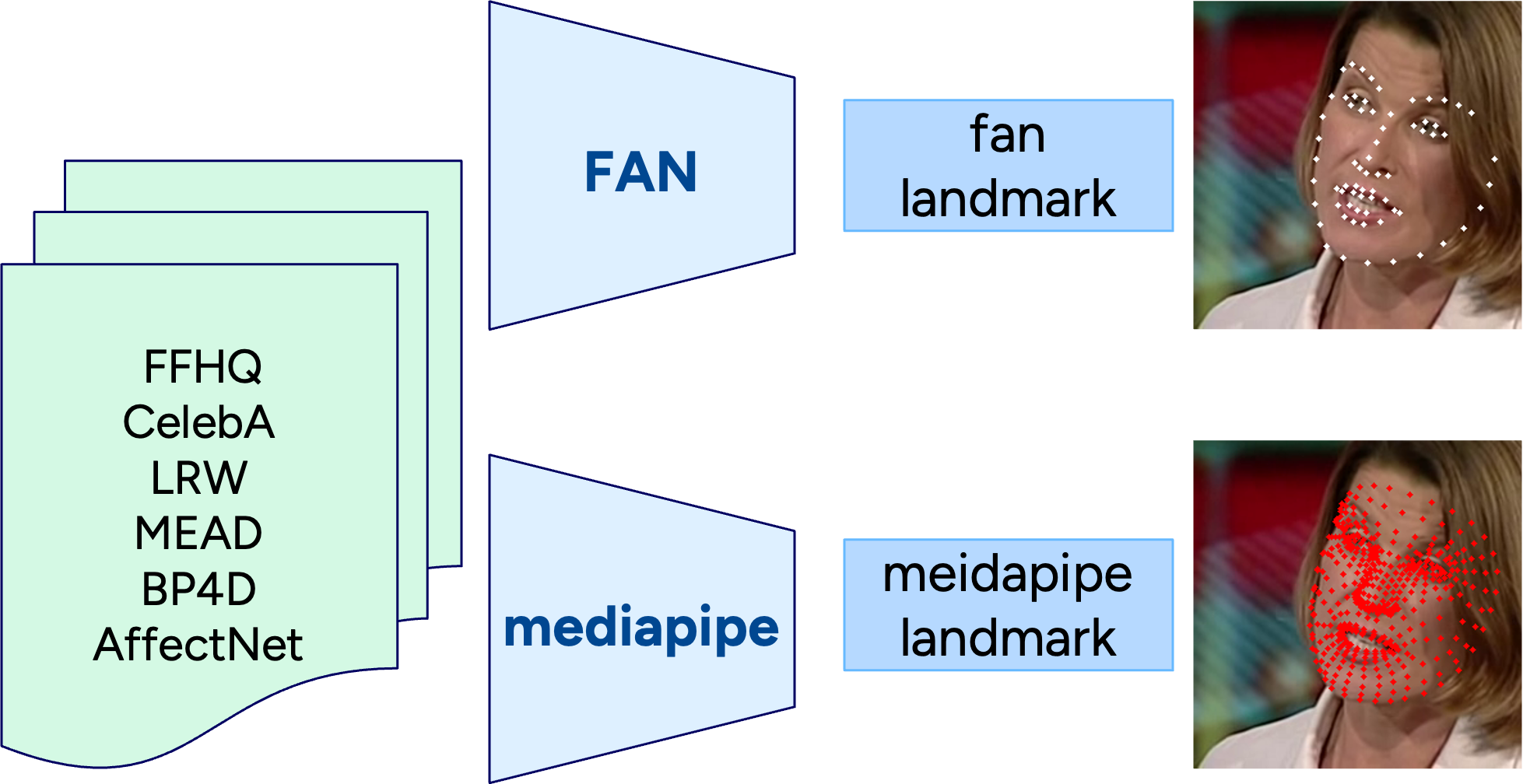}
  \caption{Data Pre-Processing Pipeline.}
  \label{fig:preprocess}
\end{figure}

\subsection{Preprocessing}
\label{suppA_preprocessing}
All datasets undergo facial landmark detection using MediaPipe~\cite{mediapipe} and FAN~\cite{fan}. The 203 InsightFace landmarks~\cite{InsightFace2022}, which are used in TEASER~\cite{teaser}, are detected during training. Based on the MediaPipe landmarks, we crop each face to a $224\times224$ resolution (\cref{fig:preprocess}). For video datasets, we randomly sample frames from each clip. Following SMIRK~\cite{smirk}, we use mixed-dataset sampling: frames from different datasets are drawn according to a predefined sampling ratio during training and validation.

\begin{figure*}[!htbp]  
  \centering
  \capstart 
  \includegraphics[width=0.93\textwidth]{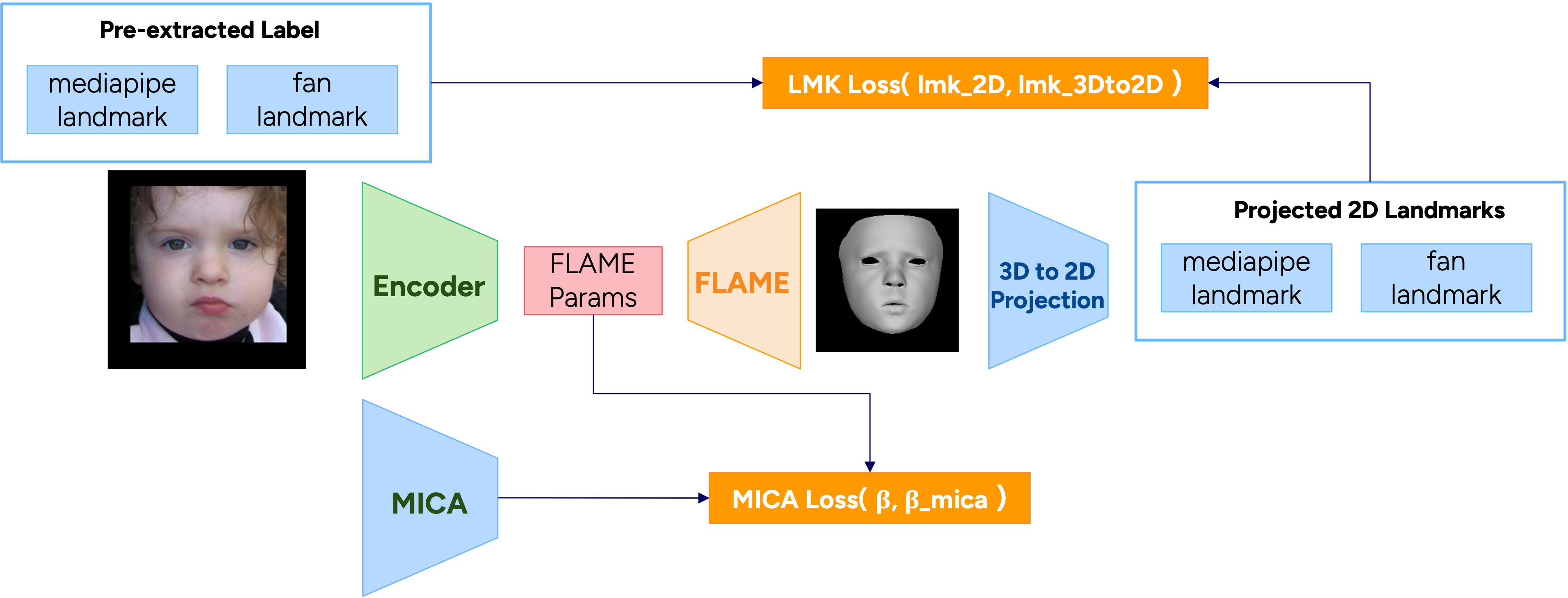}
  \caption{Illustration of the pre-training pipeline. Given an input image, the reconstruction encoder predicts FLAME parameters $(\hat\beta,\hat\Psi,\hat\Theta)$.
FLAME maps these parameters to a mesh and 3D landmarks, which are differentiably projected to image space.
In parallel, two sets of 2D landmarks are pre-extracted by MediaPipe~\cite{mediapipe} and FAN~\cite{fan} as detailed in \cref{fig:preprocess}.
MICA~\cite{mica} is run on $I$ to obtain a reference identity shape $\beta_{\text{mica}}$.}
  \label{fig:pretrain}
\end{figure*}

\subsection{Training Details}
\label{suppA_training}
As illustrated in \cref{fig:pretrain}, the goal of the pre-training stage is to obtain a stable encoder for FLAME parameters (shape $\beta$, expression $\Psi$, pose $\Theta$) \emph{without} using the synthesizer, by anchoring identity to MICA and aligning geometry with 2D landmarks.

We optimize only the parameter encoder with a landmark reprojection term and a MICA shape term:
\begin{equation}
\begin{aligned}
\mathcal{L}_{\text{pre}}
&= \lambda_{\text{lmk}}\Big( \|\hat K - K^{\text{mp}}\|_2^2
   + \|\hat K - K^{\text{fan}}\|_2^2 \Big) \\
&\quad + \lambda_{\text{mica}}\,\|\hat\beta - \beta_{\text{mica}}\|_2^2 ,
\end{aligned}
\end{equation}

where $\hat K$ are projected 3D landmarks from the FLAME given $(\hat\beta,\hat\Psi,\hat\Theta)$; we obtain 2D landmarks $\{K^{\text{mp}},K^{\text{fan}}\}$ from data pre-processing; the loss weights $\lambda_{\text{lmk}}=100$ and $\lambda_{\text{mica}}=10$.
No renderer/token branch or pixel/perceptual losses are used in this stage.

Only the FLAME-parameter encoder is updated; the synthesizer and token encoder are not used.
Model selection/early stopping is based on the MICA shape loss on the validation set.
The resulting encoder is taken as the \emph{base model} $B$ for the main training and for later regularization. 

For the encoder objective in Eq.\,\eqref{eq:encoder-loss}, we set overall loss weights $\lambda_{\text{2D}}=1$ and the corresponding loss weights vector in Eq.\,\eqref{eq:2dloss} as $\mathbf{w}_{\text{2D}}=(10,\,10,\,10,\,100,\,500)$. The emotion term uses $\lambda_{\text{emo}}=5$ with $\lambda_r=1$ and $\lambda_c=1$ in Eq.\,\eqref{eq:Lemo}; the parameter regularizer in Eq.\,\eqref{eq:rg} uses $\lambda_{\text{shape-reg}}=100$ and $\lambda_{\text{expr-reg}}=10^{-2}$. For the synthesizer objective in Eq.\,\eqref{eq:renderer-loss}, we reuse the same $\mathcal{L}_{\text{2D-cons}}$ weights, and set $\lambda_{\text{cyc}}=\lambda_{\text{tok}}=1$. We train for at most 90 epochs with Adam (lr $=2\!\times\!10^{-3}$) and batch size 64. A two-stage schedule is used: epochs 1–6 sample BP4D at 30\% with the weight $\lambda_{\text{3D}}=5$; epochs 7–90 use BP4D at 10\% with $\lambda_{\text{3D}}=1$. The cross-dataset sampling ratios are LRW 20\%, CelebA 26\%, AffectNet 26\%, FFHQ 10\%, and MEAD 8\%. 

\subsection{BP4D Data Split}
\label{suppA_bp4d_split}
We use a subject-exclusive, stratified split over the six strata
$\{\text{Asian},\ \text{White},\ \text{Black}\}\!\times\!\{\text{Male},\ \text{Female}\}$.
From the $41$ subjects, we allocate $6$ to \textit{val} and $6$ to \textit{test} -- one subject per stratum in each split (uniform $1/6$ per stratum) -- and use the remaining $29$ for \textit{train}.
This yields the following per-split composition (proportions computed over subjects in the split):

\begin{itemize}
\item \textbf{Train} ($29$ subjects; $269{,}161$ frames): 
White\_M $0.414$, White\_F $0.241$, Asian\_F $0.207$, Black\_F $0.138$
(\emph{no} Asian\_M / Black\_M in train due to subject count).
\item \textbf{Val} ($6$ subjects; $50{,}455$ frames): 
\emph{F014} (Asian, F), \emph{F019} (White, F), \emph{F023} (Black, F), 
\emph{M003} (Black, M), \emph{M008} (Asian, M), \emph{M009} (White, M).
\item \textbf{Test} ($6$ subjects; $47{,}734$ frames): 
\emph{F010} (White, F), \emph{F011} (Black, F), \emph{F021} (Asian, F), 
\emph{M004} (White, M), \emph{M005} (Asian, M), \emph{M017} (Black, M).
\end{itemize}

Dataset sizes (images/frames): \textit{train} $269{,}161$, \textit{val} $50{,}455$, \textit{test} $47{,}734$.
This subject-level stratification prevents identity leakage while ensuring that both \textit{val} and \textit{test} cover all race–gender strata uniformly.

\subsection{BP4D Fitting Procedure}
\label{suppA_bp4d_fitting}

We fit each frame of the BP4D dataset to the FLAME head model to obtain compatible pose, shape, and expression parameters, as well as mesh models as pseudo ground truth that will be used for model training and evaluation.

\noindent\textbf{Scan Pre-processing.}
BP4D scans contain hair regions and various noisy surface points, while the FLAME model does not include vertices corresponding to hair. To ensure accurate registration, we first remove these undesired regions from the raw scans.
We use the ground-truth 83 3D facial landmarks as a reference to determine which regions should be retained. Specifically, we construct an ellipse that tightly encloses all 83 landmarks after projecting them onto a common plane.
We first estimate the best-fitting plane for the 83 landmarks using principal component analysis (PCA). The plane is defined by the first two principal components, which correspond to the directions of maximum variance among the landmarks. We then project all 83 landmarks into this 2D plane and compute the radii of the ellipse:
$$
r_a=\max(|u|), ~~~ r_b=\max(|v|),
$$
where $(u,v)$ are the 2D coordinates of the projected landmarks along the two principal axes. The elliptical region is therefore defined as:
\begin{equation}
    \left(\frac{u}{r_a}\right)^2 + \left(\frac{v}{r_b}\right)^2 \le 1
\end{equation}

\begin{figure}[t]
    \centering
    \includegraphics[width=0.7\linewidth]{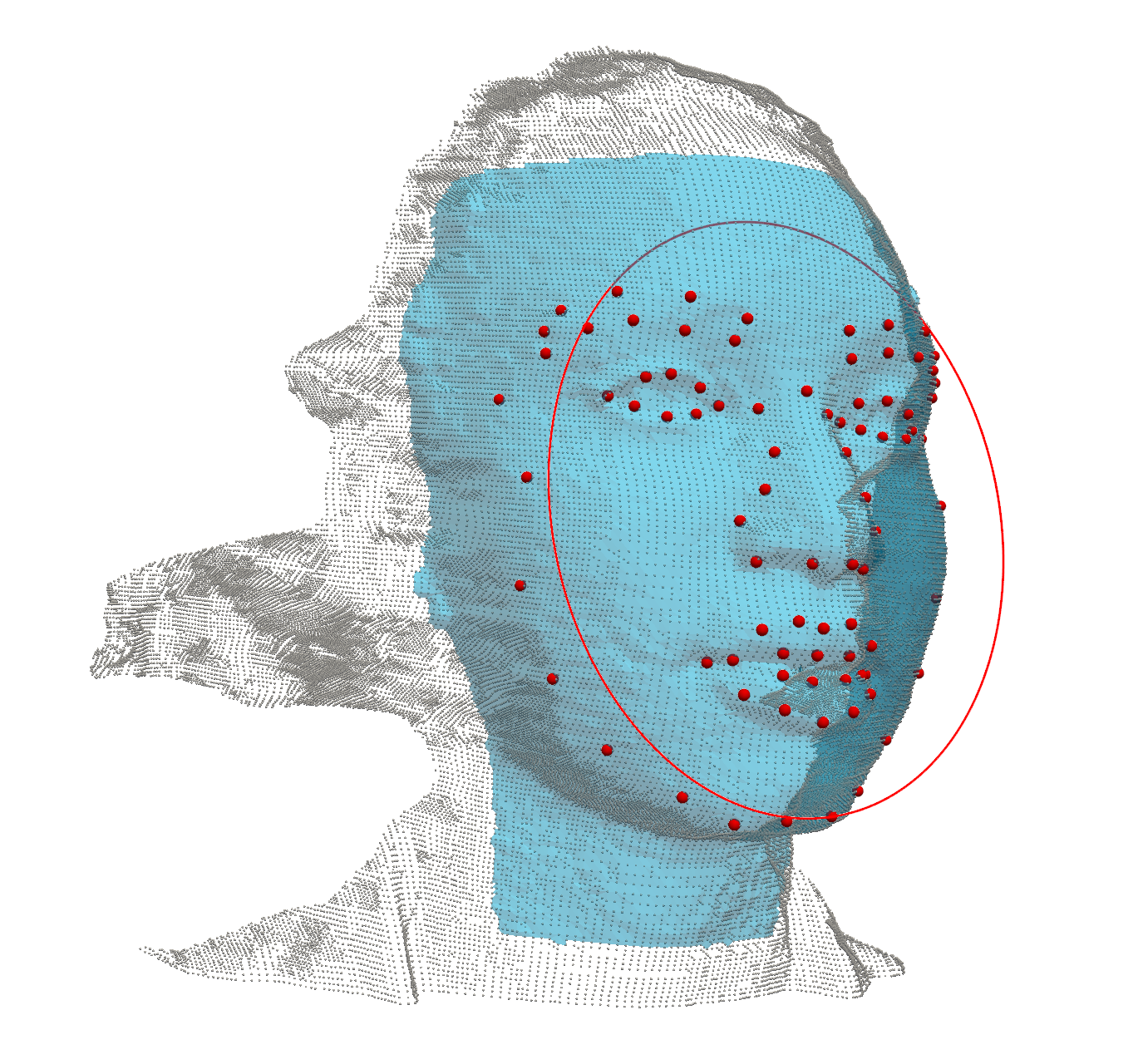}
    \caption{An example of the raw scan (gray), the landmarks with the fitted ellipse (red), and the resulting preprocessed scan (blue).}
    \label{fig:bp4d2flame_preprocessing1}
\end{figure}

Finally, we project all scan points onto the same PCA plane and retain those whose projected coordinates satisfy the elliptical constraint. This effectively removes hair and outlier regions while preserving the facial surface.
An example of the raw scan (gray), the landmarks with the fitted ellipse (red), and the resulting preprocessed scan (blue) is shown in \cref{fig:bp4d2flame_preprocessing1}.

\noindent\textbf{Fit FLAME to a 3D Scan.}
We employ the official FLAME fitting procedure \cite{FLAME}, which optimizes the model parameters by jointly minimizing the scan-to-mesh distance, the 3D landmark alignment error, and several regularization terms on shape, pose, and expression. 
The weights for the scan distance, landmark, shape, pose, and expression term are set to 2.0, 0.1, 1e-4, 1e-3, and 1e-4, respectively.

\subsection{BP4D Geometric Evaluation Protocol}
\label{suppA_bp4d_evaluation}

\noindent\textbf{Alignment Protocols.}
For each valid BP4D test frame $f$, the predicted FLAME mesh is
\begin{equation}
\widehat V_f
=\mathrm{FLAME}\!\left(
\widehat\beta_f,
\widehat\Psi_f,
\widehat\theta_{\mathrm{head},f}
\right)
\in\mathbb R^{N\times3},
\qquad N=5023.
\label{eq:bp4d_predicted_mesh}
\end{equation}
We denote its $i$-th vertex by
$\widehat v_{f,i}\in\mathbb R^3$. The scan-fitted FLAME reference is
$V_f^{\mathrm{ref}}
=\mathrm{FLAME}(\beta_f^{\mathrm{ref}},\Psi_f^{\mathrm{ref}},
\theta_{\mathrm{head},f}^{\mathrm{ref}})
\in\mathbb R^{N\times3}$, and $v_{f,i}^{\mathrm{ref}}$ denotes its
$i$-th vertex.
Because both meshes share the FLAME topology, vertices with the same index
are in correspondence. Given an alignment transformation
$\mathcal A_f$, the full-mesh error for frame $f$ is
\begin{equation}
e_f(\mathcal A_f)
=\frac{1}{N}\sum_{i=1}^{N}
\left\|
\mathcal A_f(\widehat v_{f,i})
-v_{f,i}^{\mathrm{ref}}
\right\|_2 .
\label{eq:bp4d_frame_error}
\end{equation}
The reported result is the mean of \cref{eq:bp4d_frame_error} over
the common set $\mathcal F$ of $47{,}734$ valid frames from the six
held-out test subjects:
\begin{equation}
E(\mathcal A)
=\frac{1}{|\mathcal F|}\sum_{f\in\mathcal F}e_f(\mathcal A_f).
\label{eq:bp4d_dataset_error}
\end{equation}
All distances are reported in millimeters. The alignment parameters
below are estimated independently for every frame using all corresponding
vertices. Although the transformations are fitted by minimizing squared
distances, the reported metric is the mean Euclidean vertex distance in
\cref{eq:bp4d_frame_error}, rather than an MSE or RMSE.

Under the \emph{raw} protocol, no additional alignment is applied:
\begin{equation}
\mathcal A_f^{\mathrm{raw}}(x)=x,
\qquad x\in\mathbb R^3.
\label{eq:bp4d_raw_alignment}
\end{equation}
This diagnostic retains differences in global translation, rotation,
scale, and non-rigid facial shape.

The \emph{centered} protocol aligns the mesh centroids while leaving
rotation and scale unchanged. Defining
\begin{equation}
\widehat\mu_f
=\frac{1}{N}\sum_{i=1}^{N}\widehat v_{f,i},
\qquad
\mu_f^{\mathrm{ref}}
=\frac{1}{N}\sum_{i=1}^{N}v_{f,i}^{\mathrm{ref}},
\end{equation}
the centered transformation is
\begin{equation}
\mathcal A_f^{\mathrm{cen}}(x)
=x-\widehat\mu_f+\mu_f^{\mathrm{ref}}.
\label{eq:bp4d_centered_alignment}
\end{equation}
It removes global translation while retaining rotation, scale, and
non-rigid shape differences.

The \emph{rigid} protocol estimates a rotation and translation by
solving
\begin{equation}
(R_f^{\mathrm{rig}},t_f^{\mathrm{rig}})
=\underset{\substack{R\in\mathrm{SO}(3),\\
                     t\in\mathbb R^3}}{\arg\min}
\sum_{i=1}^{N}
\left\|
R\widehat v_{f,i}+t-v_{f,i}^{\mathrm{ref}}
\right\|_2^2,
\label{eq:bp4d_rigid_fit}
\end{equation}
with
$\mathcal A_f^{\mathrm{rig}}(x)
=R_f^{\mathrm{rig}}x+t_f^{\mathrm{rig}}$.
The constraint $R\in\mathrm{SO}(3)$ excludes reflections.
Rigid alignment removes global pose and translation while preserving the
predicted metric scale; its error therefore reflects both scale mismatch
and non-rigid shape differences.

The \emph{similarity} protocol additionally estimates one positive
uniform scale per frame:
\begin{equation}
(s_f^{\mathrm{sim}},R_f^{\mathrm{sim}},t_f^{\mathrm{sim}})
=\underset{\substack{s>0,\,R\in\mathrm{SO}(3),\\
                     t\in\mathbb R^3}}{\arg\min}
\sum_{i=1}^{N}
\left\|
sR\widehat v_{f,i}+t-v_{f,i}^{\mathrm{ref}}
\right\|_2^2.
\label{eq:bp4d_similarity_fit}
\end{equation}
The corresponding transformation is
$\mathcal A_f^{\mathrm{sim}}(x)
=s_f^{\mathrm{sim}}R_f^{\mathrm{sim}}x+t_f^{\mathrm{sim}}$.
Similarity alignment evaluates non-rigid shape agreement after removing
global translation, rotation, and uniform scale. Reporting all four
protocols separates errors caused by the output coordinate frame and
global scale from errors in the reconstructed facial shape.

\begin{figure*}[t]
  \centering
  \begin{subfigure}{0.485\textwidth}
    \centering
    \includegraphics[width=\linewidth]{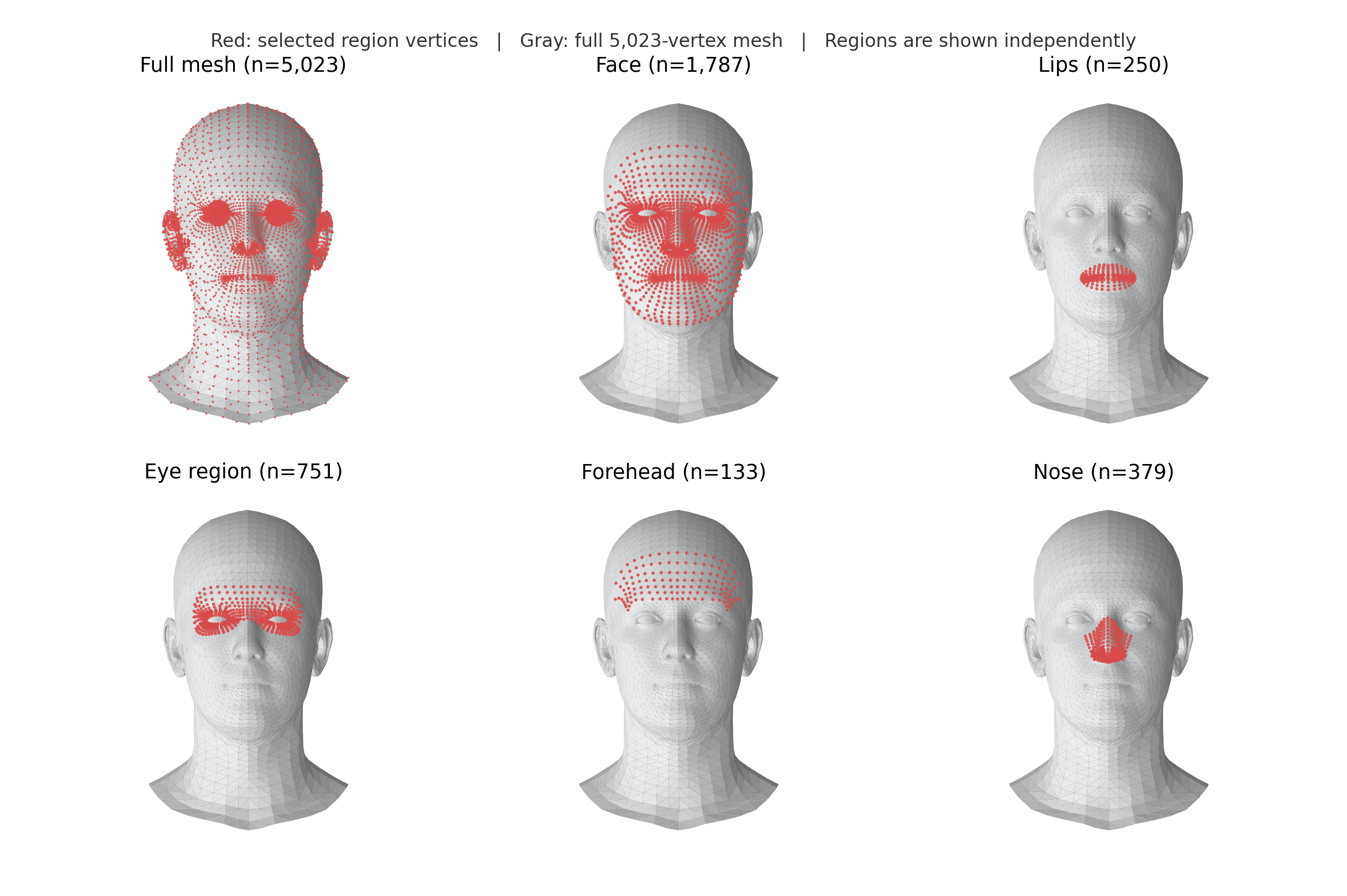}
    \caption{Frontal view.}
  \end{subfigure}\hfill
  \begin{subfigure}{0.485\textwidth}
    \centering
    \includegraphics[width=\linewidth]{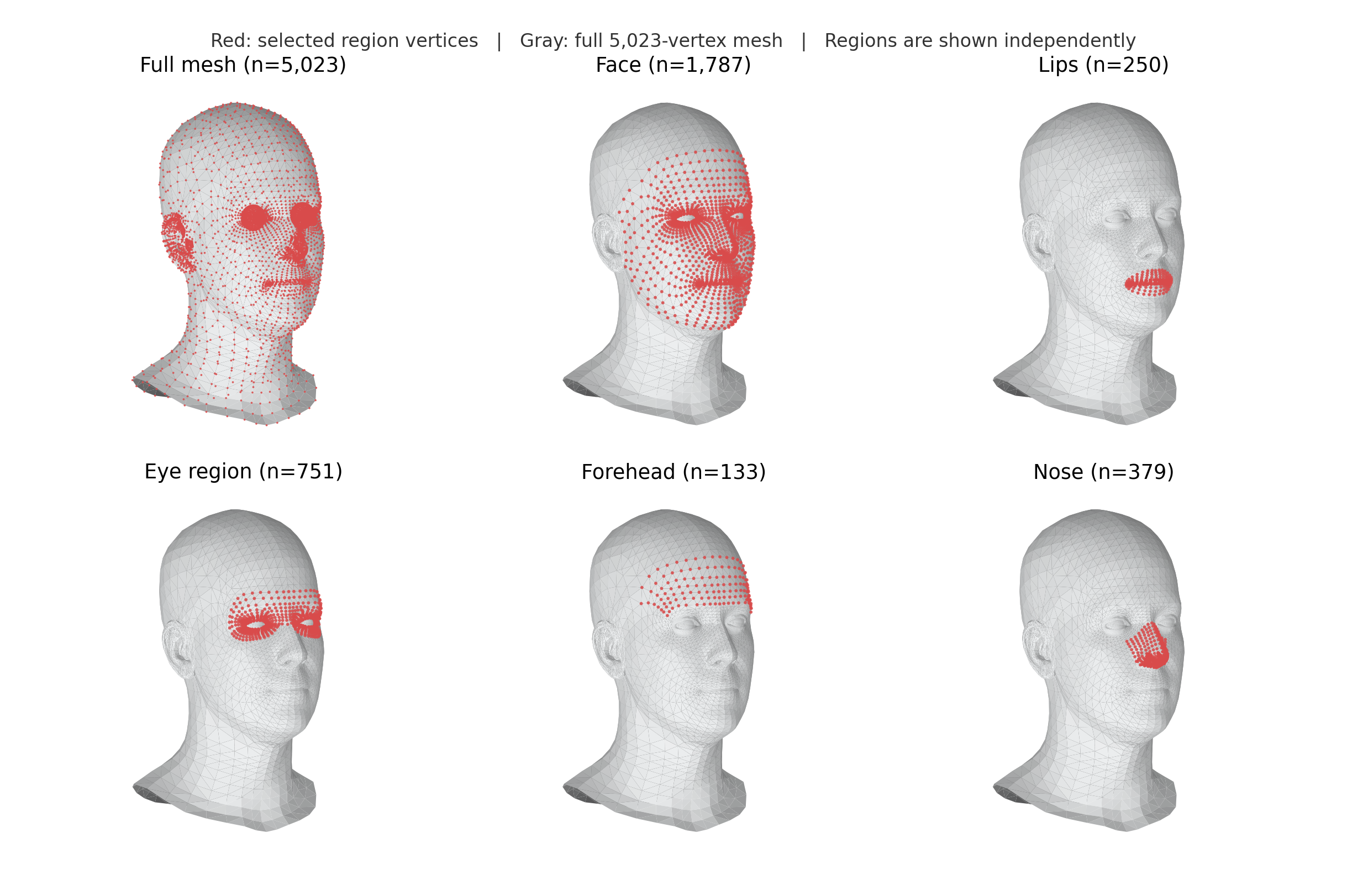}
    \caption{Oblique view.}
  \end{subfigure}
  \caption{FLAME-based facial-region masks used for BP4D geometric
  evaluation. Red vertices belong to the selected region, while the gray
  surface shows the complete FLAME mesh. The panels
  visualize the full mesh($5023$), face ($1{,}787$ vertices), lips ($250$), eye region
  ($751$), and forehead ($133$) and nose ($379$) independently from frontal and oblique
  views.}
  \label{fig:bp4d_evaluation_regions}
\end{figure*}

\noindent\textbf{FLAME-based Facial Regions.}
In addition to the full mesh, we report errors for the face, lips, eye,
forehead, and nose regions. These regions are defined using the semantic
vertex masks distributed with FLAME. Let $\Omega_r\subseteq\{1,\ldots,N\}$
denote the fixed vertex-index set for region $r$.
\Cref{fig:bp4d_evaluation_regions} visualizes the principal masks from
frontal and oblique views.

After estimating the frame-level transformation $\mathcal A_f$ from
the full mesh, we apply the same transformation to every region and
compute
\begin{equation}
e_{f,r}(\mathcal A_f)
=\frac{1}{|\Omega_r|}\sum_{i\in\Omega_r}
\left\|
\mathcal A_f(\widehat v_{f,i})
-v_{f,i}^{\mathrm{ref}}
\right\|_2 .
\label{eq:bp4d_region_error}
\end{equation}
The dataset-level regional error is obtained by averaging
$e_{f,r}$ over the same valid-frame set $\mathcal F$ as in
\cref{eq:bp4d_dataset_error}. All methods use identical FLAME vertex
correspondences, region masks, frames, and full-mesh alignment
transformations. Consequently, the regional results measure local
reconstruction error under a common global alignment rather than after
independently realigning each facial region.

\section{Additional Analyses}
\label{suppB}
To further analyze the behavior of the reconstructed expression representations, we provide additional analyses on the AffectNet validation set. While the main paper focuses on overall performance metrics, the following analyses aim to better understand how different emotion categories are captured by the learned representations.

Specifically, we first report the per-class emotion classification accuracy to examine category-wise performance differences across methods. We then analyze the confusion matrices to investigate the structure of misclassifications and to assess whether the learned representations preserve meaningful affective relationships between emotions.

\subsection{Per-Class Accuracy}
\label{suppB_perclass}

\begin{figure}[t]
  \centering
  \includegraphics[width=\linewidth]{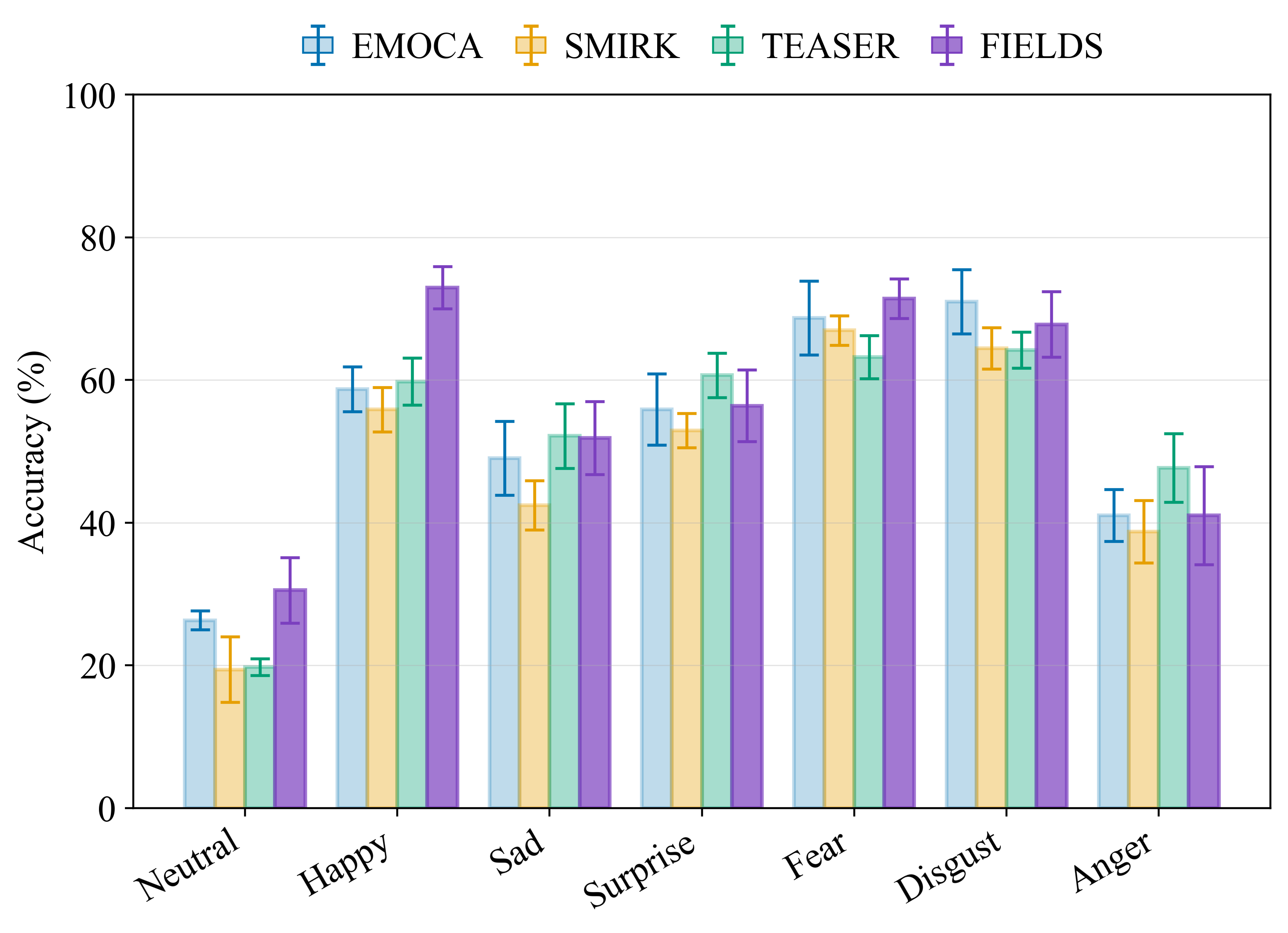}
  \caption{Emotion classification, per-class accuracy (mean$\pm$std) -- grouped bars for each emotion.}
  \label{fig:pre-class}
\end{figure}

Fig.~\ref{fig:pre-class} reports the per-class emotion classification accuracy on the AffectNet validation set. Rather than uniformly improving all emotion categories, the results reveal where the proposed supervision is most beneficial.

In particular, our method achieves the clearest gains on \emph{Neutral} and \emph{Happy}, and also performs strongly on \emph{Fear}. For several other categories, such as \emph{Sad} and \emph{Surprise}, the performance remains competitive with existing methods. In contrast, emotions such as \emph{Disgust} and \emph{Anger} remain challenging across all approaches.

These class-wise trends suggest that the proposed supervision is particularly helpful for emotions that are well captured by the global facial configuration represented in the FLAME expression space, while categories relying on subtle local cues, such as fine wrinkles or intra-oral details, remain more difficult to model with compact 3DMM-based representations.

\subsection{Confusion Matrix Analysis}
\label{suppB_confusion}

Fig.~\ref{fig:confusion-matrix} shows the confusion matrices of emotion predictions on the
AffectNet validation set. Beyond overall accuracy, the confusion
patterns reveal several consistent trends across methods.

First, most misclassifications occur between perceptually related
emotions, such as \emph{Fear} and \emph{Surprise}, or
\emph{Anger} and \emph{Disgust}. Such structured confusions
are consistent with known ambiguities in facial expression
recognition and indicate that the learned representations
capture a meaningful affective structure.

Second, positive-valence emotions such as \emph{Happy} are rarely
confused with negative categories, suggesting that the reconstructed
expression parameters preserve the coarse valence polarity.

Finally, our method achieves the strongest diagonal responses for
\emph{Neutral}, \emph{Happy}, and \emph{Fear}, while remaining
competitive on other categories. These results are consistent
with the per-class accuracy analysis and further support that
the proposed supervision improves the affective utility of the
reconstructed expression parameters.
\begin{table}[t]
  \centering
  \scriptsize
  \setlength{\tabcolsep}{3.5pt}
  \renewcommand{\arraystretch}{1.25}
  \caption{Full-mesh BP4D errors (mm) under four alignment protocols.
  Lower is better; the best and second-best results per column are shown
  in \textbf{bold} and \underline{underlined}, respectively.}
  \label{tab:bp4d_all_alignments}
  \resizebox{\columnwidth}{!}{%
  \begin{tabular}{lcccc}
    \toprule
    Method
      & Raw $\downarrow$
      & Centered $\downarrow$
      & Rigid $\downarrow$
      & Similarity $\downarrow$ \\
    \midrule
    EMOCA~\cite{emoca}
      & 111.111 & \textbf{7.769}
      & 6.643 & 4.297 \\
    SMIRK~\cite{smirk}
      & \underline{98.938} & 9.320
      & \underline{6.147} & 4.342 \\
    TEASER~\cite{teaser}
      & \textbf{98.908} & 9.658
      & 6.348 & \underline{4.237} \\
    \midrule
    FIELDS
      & 99.820 & \underline{8.962}
      & \textbf{5.720} & \textbf{3.884} \\
    \bottomrule
  \end{tabular}
  }
\end{table}

\begin{table*}[t]
  \centering
  \footnotesize
  \setlength{\tabcolsep}{3.8pt}
  \renewcommand{\arraystretch}{1.05}
  \setlength{\abovecaptionskip}{2pt}
  \setlength{\belowcaptionskip}{2pt}

  \caption{AffectNet KNN evaluation using 53-D expression--jaw
  parameters. RMSE$\downarrow$ is lower-better; all other metrics
  are higher-better. The best and second-best results per column are
  shown in \textbf{bold} and \underline{underlined}.}
  \label{tab:knn_exprjaw_compact}

  \begin{tabular*}{\textwidth}
    {@{\extracolsep{\fill}}lcccc|cccc|c@{}}
    \toprule
    Model
      & V-PCC$\uparrow$
      & V-CCC$\uparrow$
      & V-RMSE$\downarrow$
      & V-SAGR$\uparrow$
      & A-PCC$\uparrow$
      & A-CCC$\uparrow$
      & A-RMSE$\downarrow$
      & A-SAGR$\uparrow$
      & E-Acc$\uparrow$ \\
    \midrule

    EMOCA~\cite{emoca}
      & \underline{0.671}
      & \underline{0.635}
      & \underline{0.384}
      & 0.731
      & \underline{0.505}
      & \underline{0.432}
      & \underline{0.376}
      & \underline{0.728}
      & \underline{0.472} \\

    SMIRK~\cite{smirk}
      & 0.635
      & 0.594
      & 0.401
      & \underline{0.736}
      & 0.450
      & 0.370
      & 0.396
      & 0.713
      & 0.448 \\

    TEASER~\cite{teaser}
      & 0.585
      & 0.534
      & 0.433
      & 0.695
      & 0.432
      & 0.350
      & 0.404
      & 0.693
      & 0.429 \\

    \midrule

    FIELDS
      & \textbf{0.713}
      & \textbf{0.696}
      & \textbf{0.355}
      & \textbf{0.790}
      & \textbf{0.572}
      & \textbf{0.525}
      & \textbf{0.349}
      & \textbf{0.740}
      & \textbf{0.514} \\

    \bottomrule
  \end{tabular*}
\end{table*}

\subsection{KNN Evaluation on AffectNet}
The MLP protocol uses the same evaluator architecture and training
protocol for every reconstruction method, but the evaluator still learns
a nonlinear mapping from FLAME parameters to affect labels. We therefore
add a non-parametric KNN probe.
It tests whether affective neighborhood structure is already present in
the reconstructed parameters, without a learned evaluator or identity-shape
features. The input is the 53-D vector
$x=[\psi_{exp};\theta_{jaw}]$, consisting of 50
expression parameters and three jaw-pose parameters.

For each method, we extract $283{,}901$ valid samples from the
AffectNet training split and $3{,}500$ samples from the public validation
split. A standard scaler is fitted only on the training features and then
applied to both splits. The standardized training features and their labels
form the reference database. The public validation features are used only
as queries. Although AffectNet names this split the public validation set,
we use it as a held-out evaluation set because the official test labels are
unavailable. This KNN evaluation uses neither five-fold cross-validation
nor repeated runs; it produces one deterministic result per method.

We use five nearest neighbors in the standardized feature space with
Euclidean distance and inverse-distance weighting. Emotion classification
uses weighted voting, while valence and arousal are predicted by weighted
averaging of the neighbors' annotations.

As shown in \cref{tab:knn_exprjaw_compact}, FIELDS achieves the best
result for every reported metric. The gain therefore remains visible with
a fixed non-parametric predictor and does not depend solely on a learned
MLP evaluator.

\subsection{Effect of Alignment Protocols on BP4D Evaluation}
\label{suppB_alignment}

We complement the rigid and similarity results in the main paper with
raw and centered evaluation. The four protocols are defined in
\cref{suppA_bp4d_evaluation}. All entries in
\cref{tab:bp4d_all_alignments} are full-mesh corresponding-vertex errors
on the same $47{,}734$ BP4D test frames.

Raw error compares meshes in their original coordinate systems.
It therefore includes global translation, rotation, scale, and facial-shape
differences. These coordinate effects dominate the distance and produce
values near $100$\,mm. After centering removes translation, the errors fall
below $10$\,mm for all methods. Raw and centered errors are thus useful
diagnostics, but they do not isolate reconstruction shape accuracy.

Rigid alignment removes translation and rotation while retaining
metric scale. Similarity alignment also removes one uniform scale per
frame. FIELDS achieves the lowest error under both protocols. This shows
that its advantage remains both with and without scale normalization.

\begin{figure*}[!htbp]
  \centering
    \includegraphics[width=0.985\linewidth]{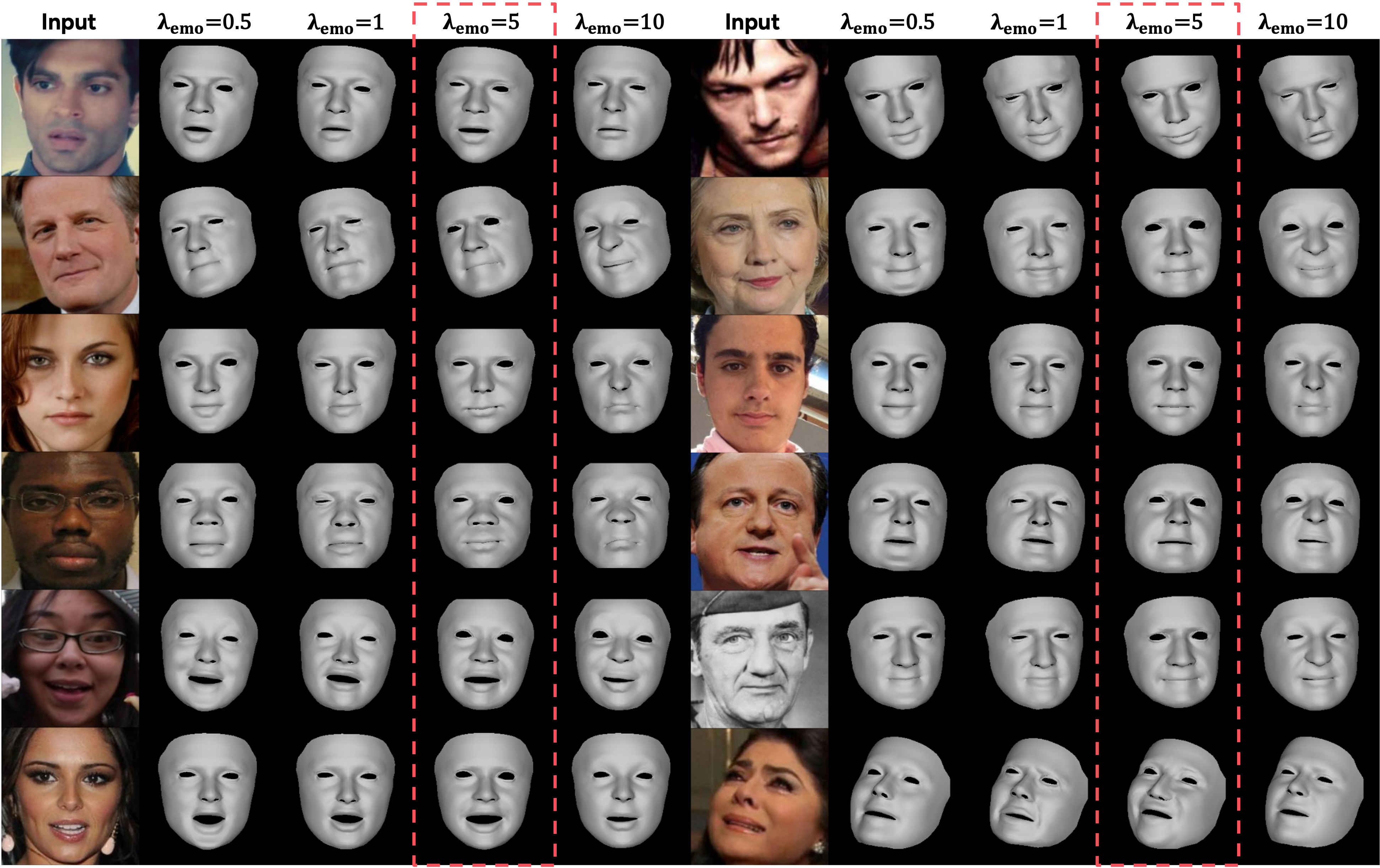}
  \caption{Qualitative effect of the emotion-loss weight $\lambda_{\text{emo}}$ on AffectNet.
For each example, the input image is shown on the left, followed by reconstructions from models trained with $\lambda_{\text{emo}}=0.5$, $1$, $5$, and $10$ (left to right).
Red dashed boxes mark the final setting ($\lambda_{\text{emo}}=5$).
Smaller weights tend to under-express mouth opening and brow motion, whereas $\lambda_{\text{emo}}=10$ can produce slightly exaggerated local deformations.}
  \label{fig:emoW-stack}
\end{figure*}

\begin{table*}[t]
  \centering
  \scriptsize
  \setlength{\tabcolsep}{2.5pt}
  \renewcommand{\arraystretch}{1.0}

  \caption{Ablation on the emotion-loss weight
  $\lambda_{\text{emo}}$.
  Affective performance is reported as
  mean\textsuperscript{\scriptsize$\pm$}std over five folds;
  geometric evaluation reports FLAME parameter MSE and
  rigid-/similarity-aligned full-mesh errors (mm) on BP4D.
  The best and second-best mean results per column are shown in
  \textbf{bold} and \underline{underlined}.}
  \label{tab:lambdaemo_unified_suppl}

  \resizebox{\linewidth}{!}{%
  \begin{tabular}{l c cccc cc cc}
    \toprule
    & \multicolumn{1}{c}{\textbf{Emo Cls.}}
    & \multicolumn{4}{c}{\textbf{V/A Regression}}
    & \multicolumn{2}{c}{\textbf{FLAME MSE} $\downarrow$}
    & \multicolumn{2}{c}{\textbf{3D errors} (mm) $\downarrow$} \\
    \cmidrule(lr){2-2}
    \cmidrule(lr){3-6}
    \cmidrule(lr){7-8}
    \cmidrule(lr){9-10}

    $\lambda_{\text{emo}}$
      & Acc (\%) $\uparrow$
      & V-CCC $\uparrow$
      & V-RMSE $\downarrow$
      & A-CCC $\uparrow$
      & A-RMSE $\downarrow$
      & $\{\psi_{exp},\theta_{jaw}\}$
      & $\beta$
      & Rigid
      & Similarity \\
    \midrule

    0.5
      & \meanstd{54.30}{1.83}
      & \msup{0.719}{0.005}
      & \msup{0.346}{0.006}
      & \msup{0.592}{0.007}
      & \msup{0.354}{0.008}
      & 0.188
      & 0.355
      & \underline{5.582}
      & 3.874 \\

    1
      & \meanstd{54.17}{1.17}
      & \msup{0.726}{0.006}
      & \msup{0.344}{0.005}
      & \msup{0.588}{0.012}
      & \msup{0.355}{0.004}
      & \underline{0.176}
      & 0.357
      & \textbf{5.570}
      & \underline{3.821} \\

    5 (Ours)
      & \bmeanstd{55.99}{1.45}
      & \bmsup{0.737}{0.006}
      & \bmsup{0.338}{0.006}
      & \msup{\underline{0.604}}{0.008}
      & \msup{\underline{0.352}}{0.011}
      & \textbf{0.173}
      & \textbf{0.349}
      & 5.720
      & 3.884 \\

    10
      & \umeanstd{55.66}{1.37}
      & \msup{\underline{0.735}}{0.003}
      & \msup{\underline{0.340}}{0.003}
      & \bmsup{0.613}{0.004}
      & \bmsup{0.345}{0.005}
      & 0.195
      & \underline{0.353}
      & 5.588
      & \textbf{3.812} \\

    \bottomrule
  \end{tabular}%
  }
\end{table*}
\section{Additional Ablation Studies}
\label{suppC}
In this section, we analyze the sensitivity of the proposed framework to the emotion loss weight.

\subsection{Emotion Loss Weight Sensitivity}
\label{suppC_loss_weight}

We vary the emotion-loss weight $\lambda_{\text{emo}}$ in the encoder objective (\cref{eq:encoder-loss}), where $\mathcal{L}_{\mathrm{emo}}$ is defined in \cref{eq:Lemo}, and summarize the results in \cref{tab:lambdaemo_unified_suppl}.

\begin{figure*}[t]
  \centering
  \includegraphics[width=\linewidth]{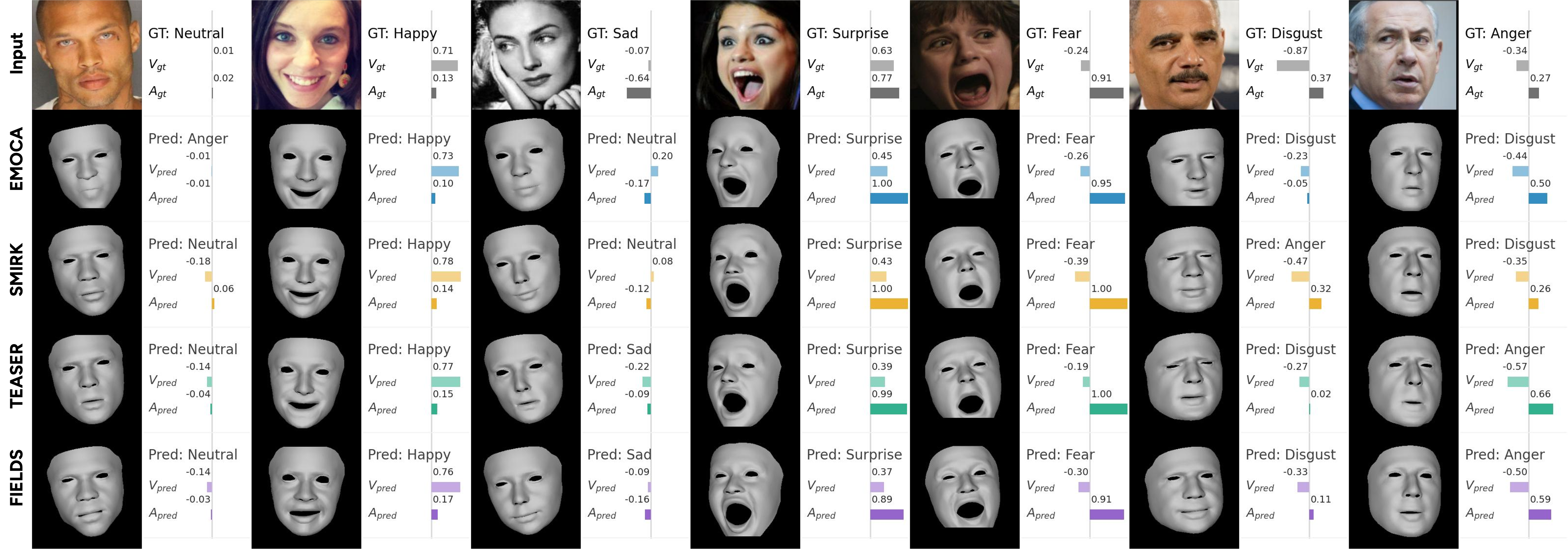}
  \caption{Additional qualitative comparisons on the AffectNet validation set. For each example, we show the input image, the ground-truth affect annotations, and the reconstructed meshes with predicted affect outputs from different methods. The figure provides further visual evidence beyond the representative cases shown in the main paper.}
  \label{fig:affectnet_more}
\end{figure*}

\begin{figure*}[t]
  \centering
  \includegraphics[width=\linewidth]{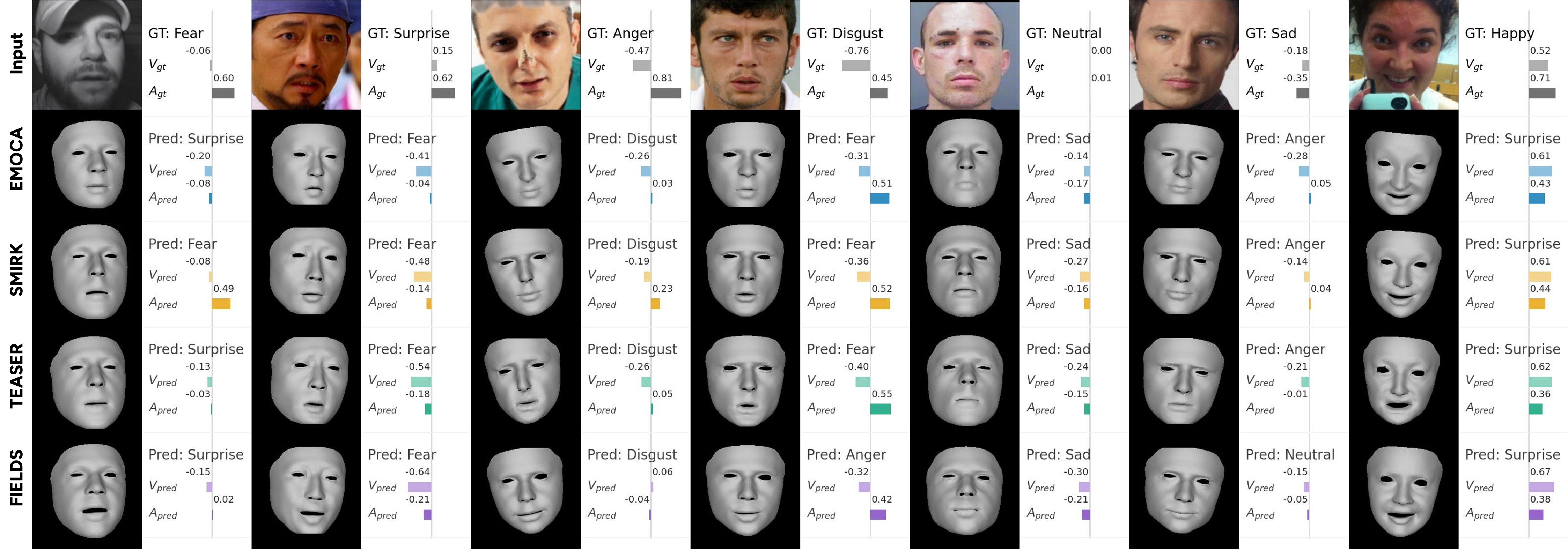}
\caption{
Representative failure cases on AffectNet.
Each column shows the input image with ground-truth affect annotation
and reconstructed faces with predicted affect values for each method.
The selected examples follow the dominant confusion patterns identified
in \cref{fig:confusion-matrix}, where most errors occur between
semantically related emotions such as Fear--Surprise, Anger--Disgust, Neutral--Sad, and Happy--Surprise.
}
  \label{fig:affectnet_failure}
\end{figure*}

The affect-related metrics are more sensitive to $\lambda_{\text{emo}}$ than
the FLAME parameter MSE. Compared with smaller weights ($0.5$ and $1$),
larger values ($5$ and $10$) generally improve emotion classification and
valence--arousal regression. The rigid and similarity errors vary
non-monotonically with the emotion-loss weight: $\lambda_{\text{emo}}=1$
gives the lowest rigid error, whereas $\lambda_{\text{emo}}=10$ gives the
lowest similarity error.

Among the tested settings, $\lambda_{\text{emo}}=5$ gives the highest
classification accuracy, the strongest valence regression performance,
and the lowest FLAME parameter MSE. Although $\lambda_{\text{emo}}=10$
performs slightly better on the arousal metrics and similarity-aligned
geometry, it is less consistent across the affective metrics. We therefore
use $\lambda_{\text{emo}}=5$ in the final model.

Fig.~\ref{fig:emoW-stack} shows qualitative reconstructions on AffectNet as the emotion-loss weight increases from left to right ($\lambda_{\text{emo}}=0.5,1,5,10$). 
With smaller weights, the reconstructed expressions are often under-expressed, with weaker mouth opening and eyebrow motion. 
Increasing the weight leads to more pronounced facial deformations, and $\lambda_{\text{emo}}=5$ generally produces the most balanced results, with expressions that are both visually expressive and geometrically plausible. 
When the weight is further increased to $\lambda_{\text{emo}}=10$, the reconstructions exhibit stronger activations, such as larger mouth opening or more pronounced local deformations, which can appear slightly exaggerated in some cases. 
The quantitative geometric errors in
\cref{tab:lambdaemo_unified_suppl} vary non-monotonically and therefore do
not establish a monotonic relationship between the emotion-loss weight and
geometric accuracy. We select $\lambda_{\text{emo}}=5$ based on its stronger
overall affective performance rather than the lowest aligned mesh error.

\section{Additional Qualitative Results}
\label{suppD}
\subsection{AffectNet Expression Comparison}
\label{suppD_affectnet}

\noindent\textbf{Additional qualitative comparisons.}
\cref{fig:affectnet_more} shows additional examples on the AffectNet validation set.
The predictions and renderings are obtained using the same protocol as described in \cref{sec:quali}.
These examples are intended to provide further visual evidence beyond the representative cases shown in the main text.

Across different identities and affect intensities, FIELDS tends to preserve localized facial deformation patterns that remain visually consistent with the input image.
In particular, differences are most visible around the mouth opening, mouth-corner curvature, cheek deformation, and brow motion.
Compared with the baseline reconstructions, which may match affect targets by amplifying expression intensity, FIELDS more often maintains subtle geometric cues that better reflect the underlying facial expression.

\begin{figure*}[p]
  \centering
  \includegraphics[width=\linewidth]{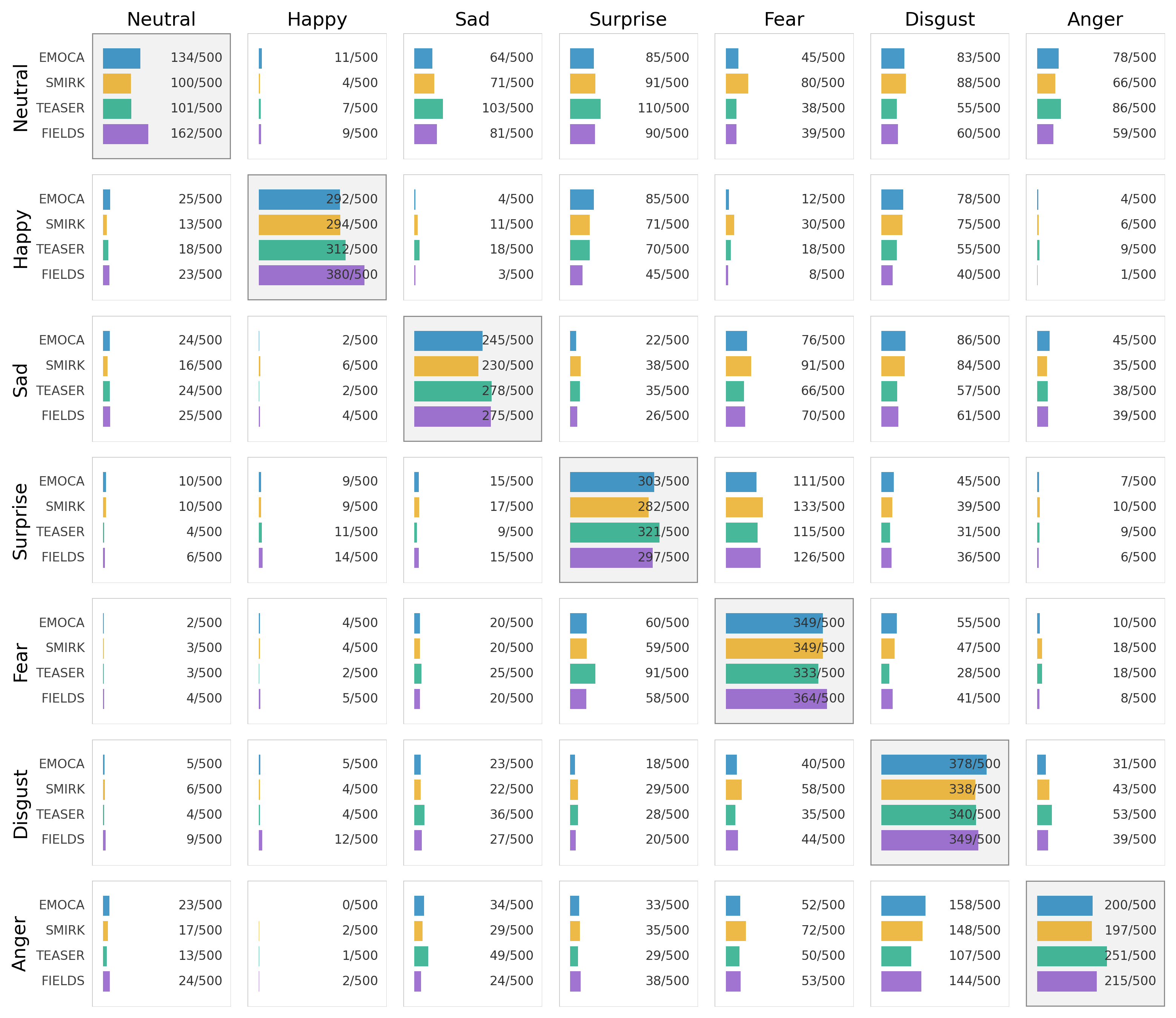}
  \caption{Normalized confusion matrices of emotion predictions on the AffectNet validation set for different reconstruction methods. Rows correspond to ground-truth emotions and columns to predicted emotions. The diagonal entries indicate correct classifications, while off-diagonal entries reveal structured confusions between perceptually related emotions (e.g., fear--surprise and anger--disgust).}
  \label{fig:confusion-matrix}
\end{figure*}

\noindent\textbf{Failure Cases.}
\cref{fig:affectnet_failure} shows representative failure cases on the AffectNet validation set.
The selected examples follow the dominant confusion patterns identified in the confusion matrix (\cref{fig:confusion-matrix}), namely semantically related emotion pairs such as Fear--Surprise, Anger--Disgust, Neutral--Sad, and Happy--Surprise.
These cases are therefore not arbitrary mistakes, but reflect challenging category boundaries in affect recognition.

Across these examples, the remaining errors typically arise when facial cues are visually subtle or shared by multiple nearby emotion categories, such as wide eye opening in Fear/Surprise, lower-face tension in Anger/Disgust, or weak mouth-corner changes in Sad/Neutral.

Even in such ambiguous cases, the reconstruction produced by FIELDS remains visually plausible in many examples and still preserves several informative local cues. This suggests that at least part of the remaining errors may be associated with semantic ambiguity between nearby emotion categories, rather than with obvious reconstruction failure alone.

\subsection{BP4D Reconstruction Examples}
\label{suppD_bp4d}

\begin{figure*}[!htbp]
  \centering

  \begin{subfigure}{\textwidth}
    \centering
    \includegraphics[width=\linewidth]{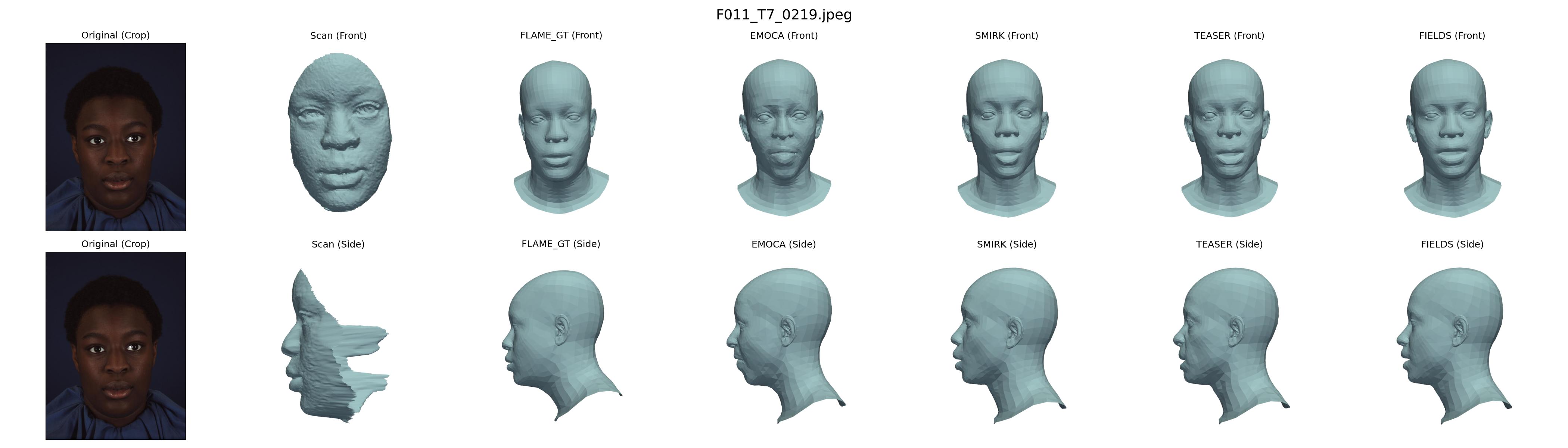}
    \label{fig:stack-b}
  \end{subfigure}
  \vspace{0.1em}
  \begin{subfigure}{\textwidth}
    \centering
    \includegraphics[width=\linewidth]{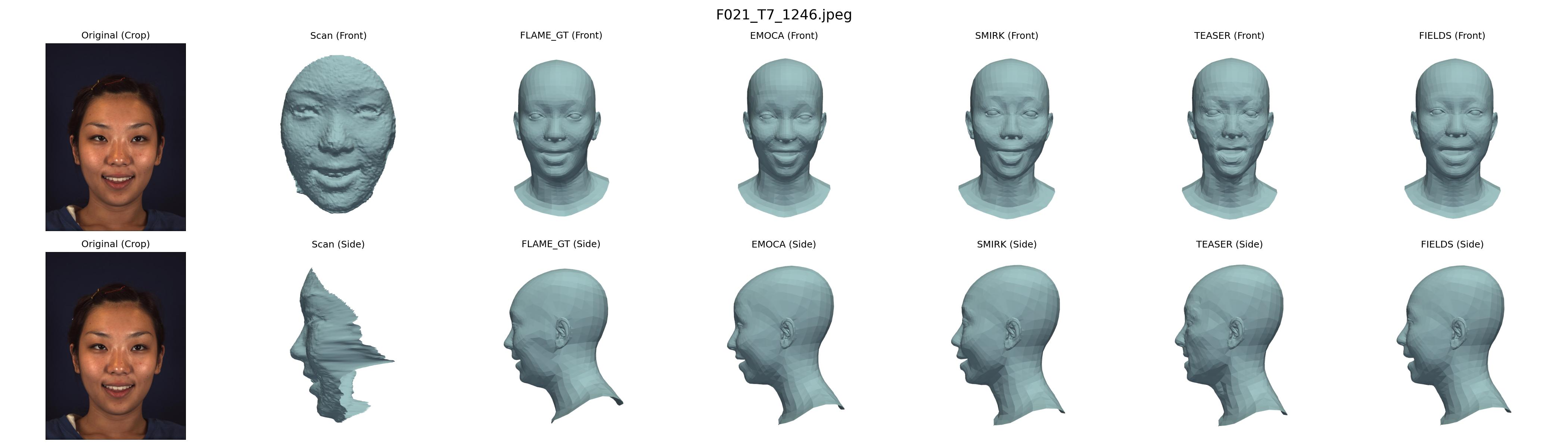}
    \label{fig:stack-c}
  \vspace{0.1em}
  \end{subfigure}
    \begin{subfigure}{\textwidth}
    \centering
    \includegraphics[width=\linewidth]{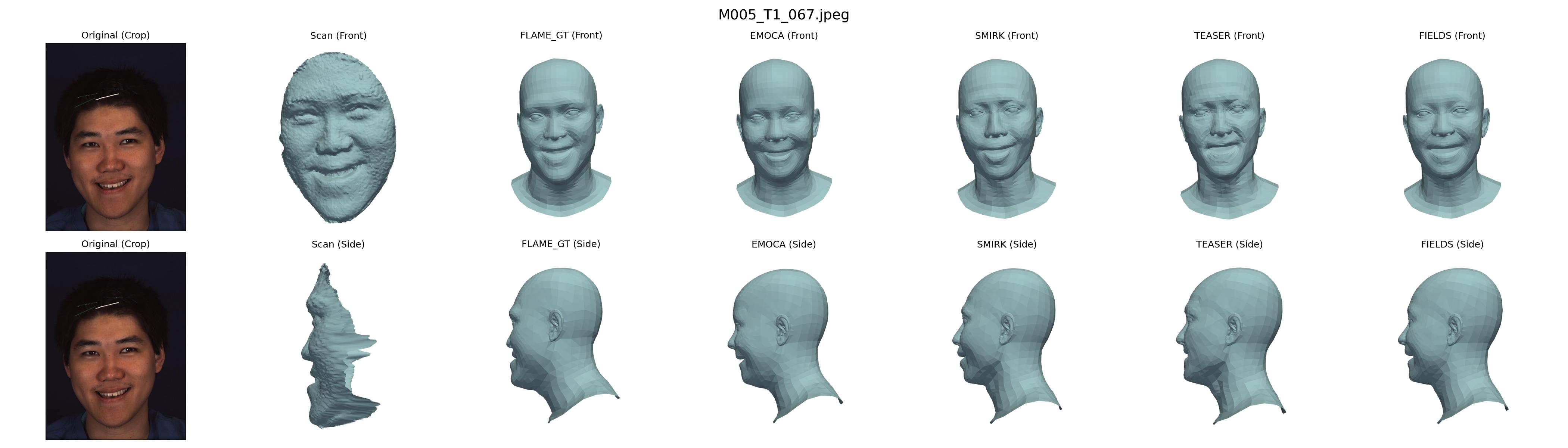}
    \label{fig:stack-b_bp4dm}
  \end{subfigure}
  \vspace{0.1em}
  \begin{subfigure}{\textwidth}
    \centering
    \includegraphics[width=\linewidth]{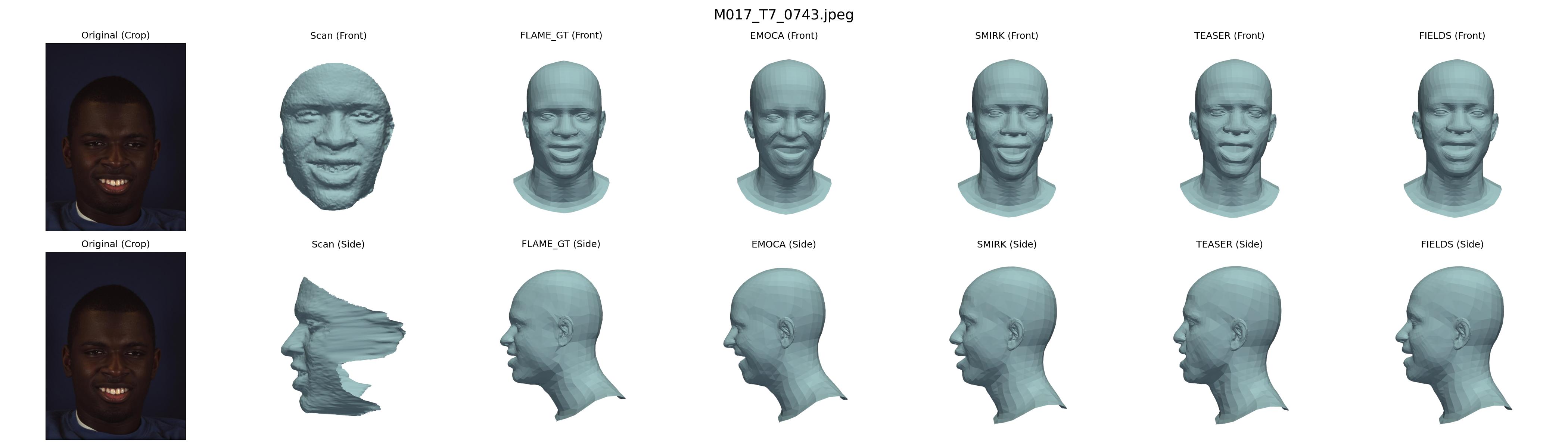}
    \label{fig:stack-c_bp4dm}
  \end{subfigure}

  \caption{BP4D 3D mesh model reconstruction examples. From left to right: BP4D original image, 3D scan, fitted FLAME model, EMOCA\cite{emoca}, SMIRK\cite{smirk}, TEASER\cite{teaser}, FIELDS.}
  \label{fig:bp4df}
\end{figure*}

\cref{fig:bp4df} provides additional qualitative comparisons on the BP4D test split across multiple subjects.
For each subject, we show the input RGB frame, the synchronized 3D scan, the fitted FLAME reference derived from the scan, and the corresponding reconstructions from EMOCA, SMIRK, TEASER, and FIELDS.
To inspect the differences, it is particularly useful to compare the front and side views around the chin, jaw contour, and mouth region, where geometric inaccuracies are more easily revealed.
Across these examples, FIELDS remains visually closer to the fitted FLAME reference, especially in the lower-face profile, and avoids artifacts that are more noticeable in other methods, such as excessive mouth opening, profile compression, or exaggerated facial motion.
This qualitative trend is consistent with the geometric fidelity results reported in the main paper.

\section{Embedding Visualization}
\label{suppE}

To further analyze the structure of the learned expression embeddings, we provide t-SNE and UMAP visualizations of the AffectNet validation set.

\begin{figure*}[!htbp]
  \centering
  \captionsetup[subfigure]{font=scriptsize}
  \newcommand{\subw}{0.41\textwidth}   
  \newcommand{\colgap}{0.03\textwidth} 

  \begin{subfigure}{\subw}
    \centering
    \includegraphics[width=\linewidth]{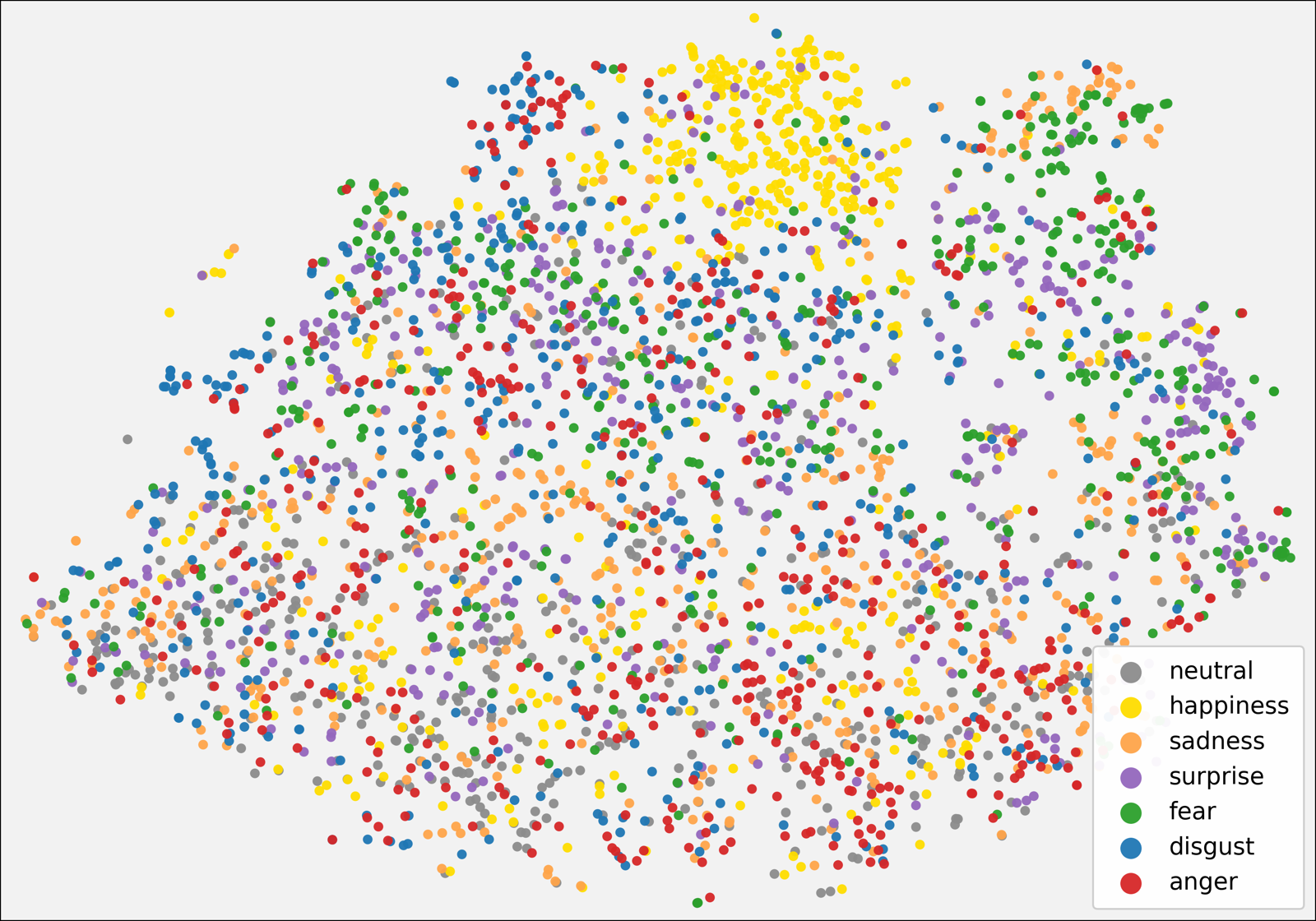}
    \caption{EMOCA Class}
  \end{subfigure}\hspace{\colgap}%
  \begin{subfigure}{\subw}
    \centering
    \includegraphics[width=\linewidth]{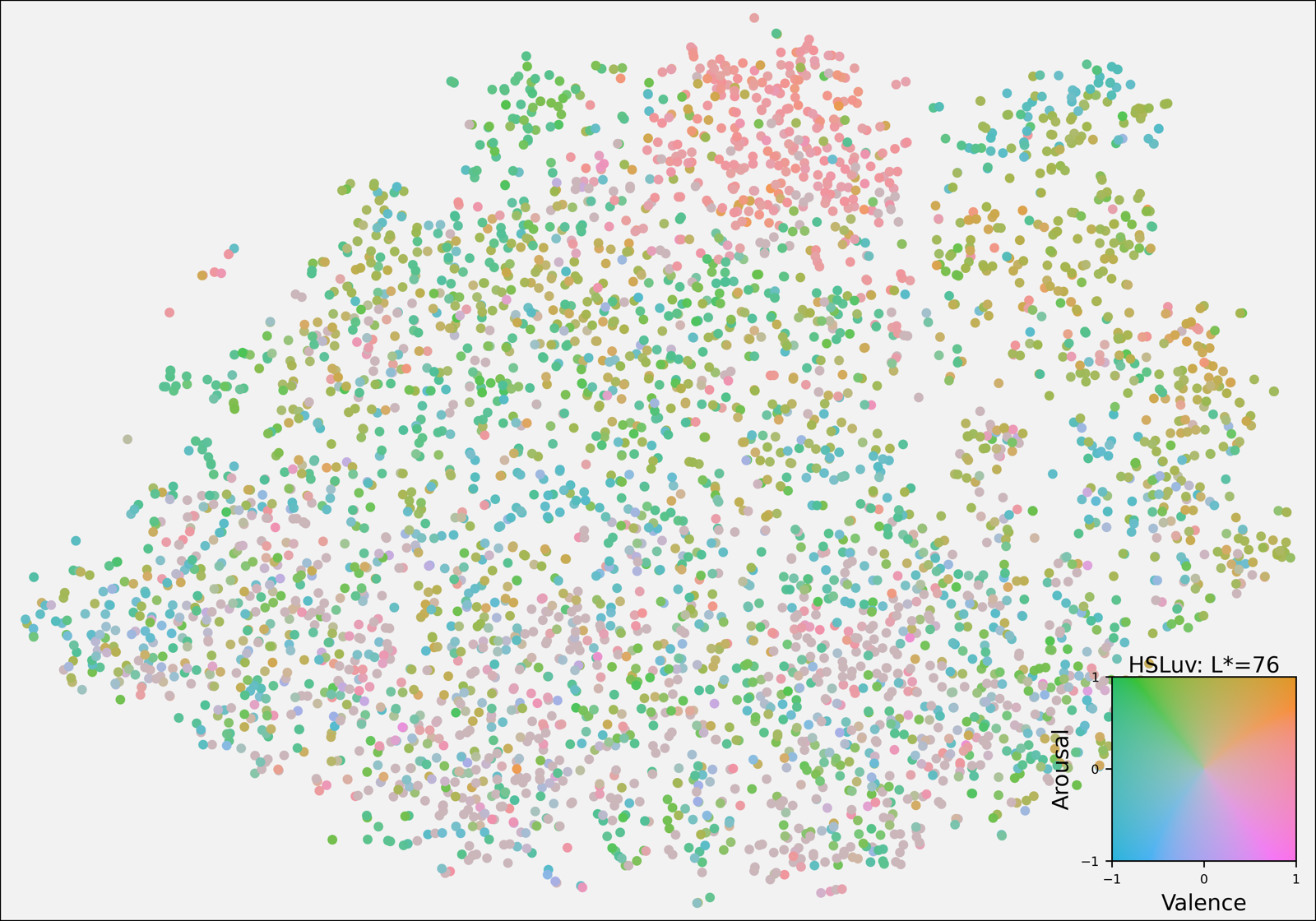}
    \caption{EMOCA V\&A}
  \end{subfigure}


  \begin{subfigure}{\subw}
    \centering
    \includegraphics[width=\linewidth]{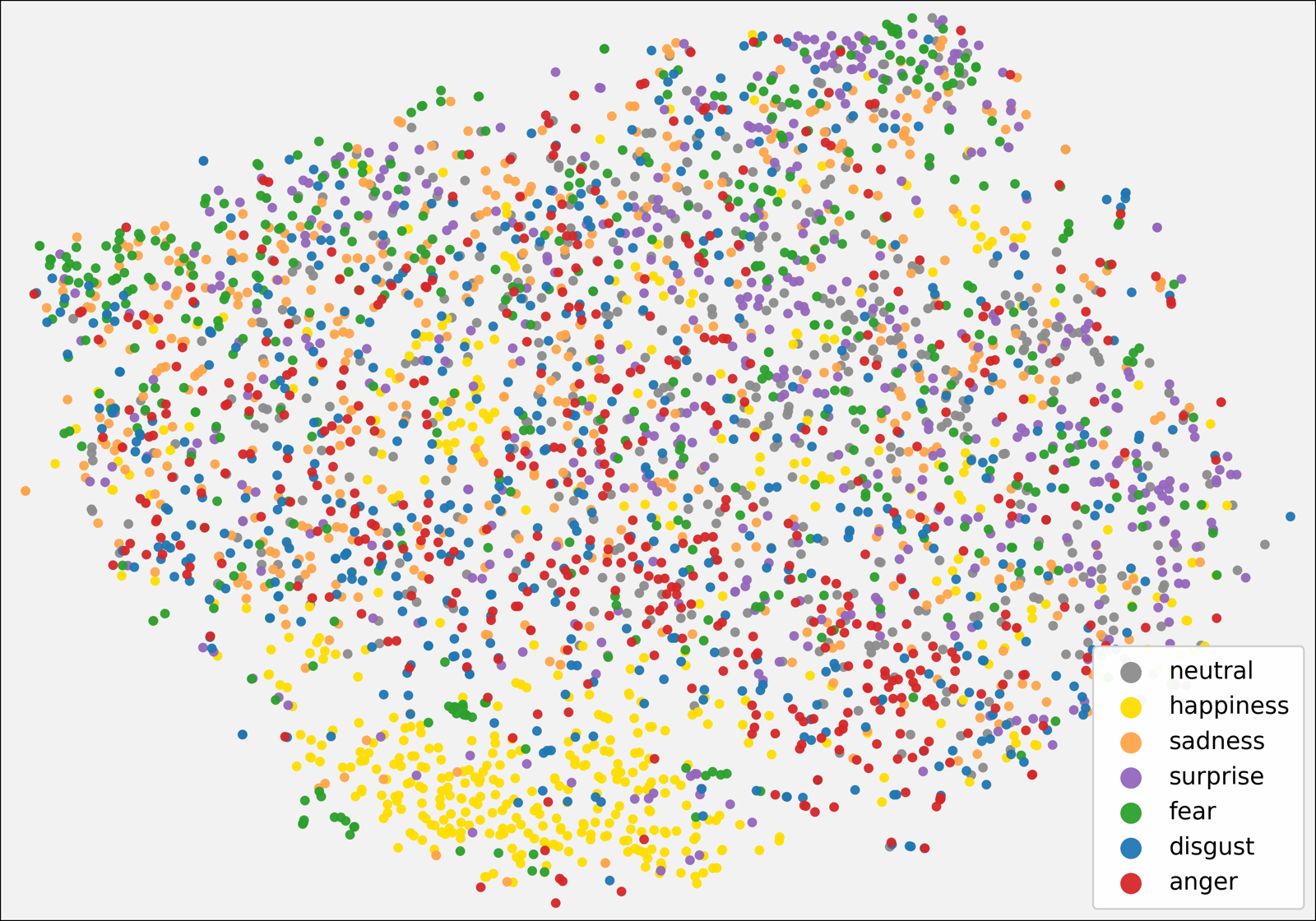}
    \caption{SMIRK Class}
  \end{subfigure}\hspace{\colgap}%
  \begin{subfigure}{\subw}
    \centering
    \includegraphics[width=\linewidth]{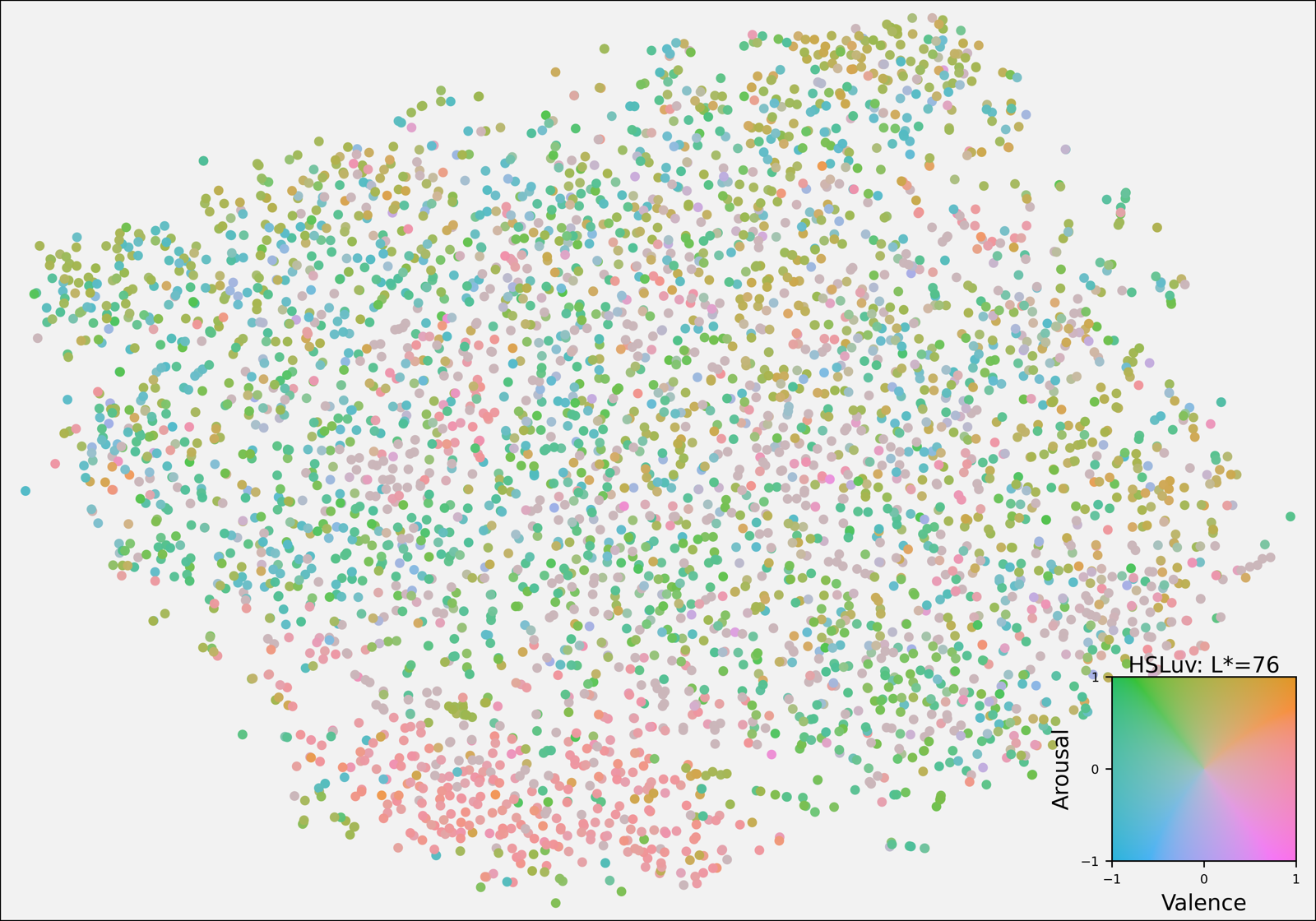}
    \caption{SMIRK V\&A}
  \end{subfigure}


  \begin{subfigure}{\subw}
    \centering
    
    \includegraphics[width=\linewidth]{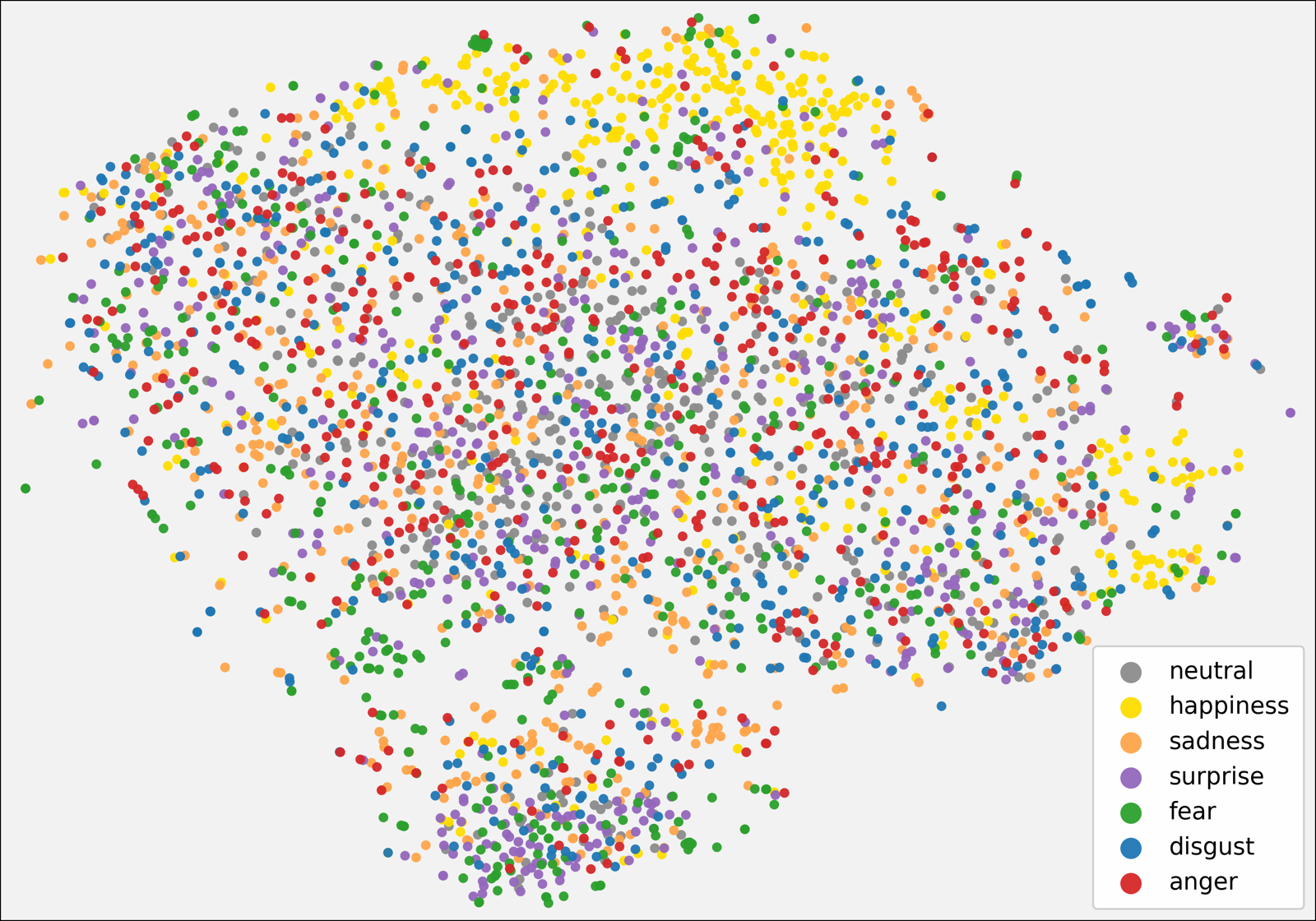}
    \caption{TEASER Class}
  \end{subfigure}\hspace{\colgap}%
  \begin{subfigure}{\subw}
    \centering
    \includegraphics[width=\linewidth]{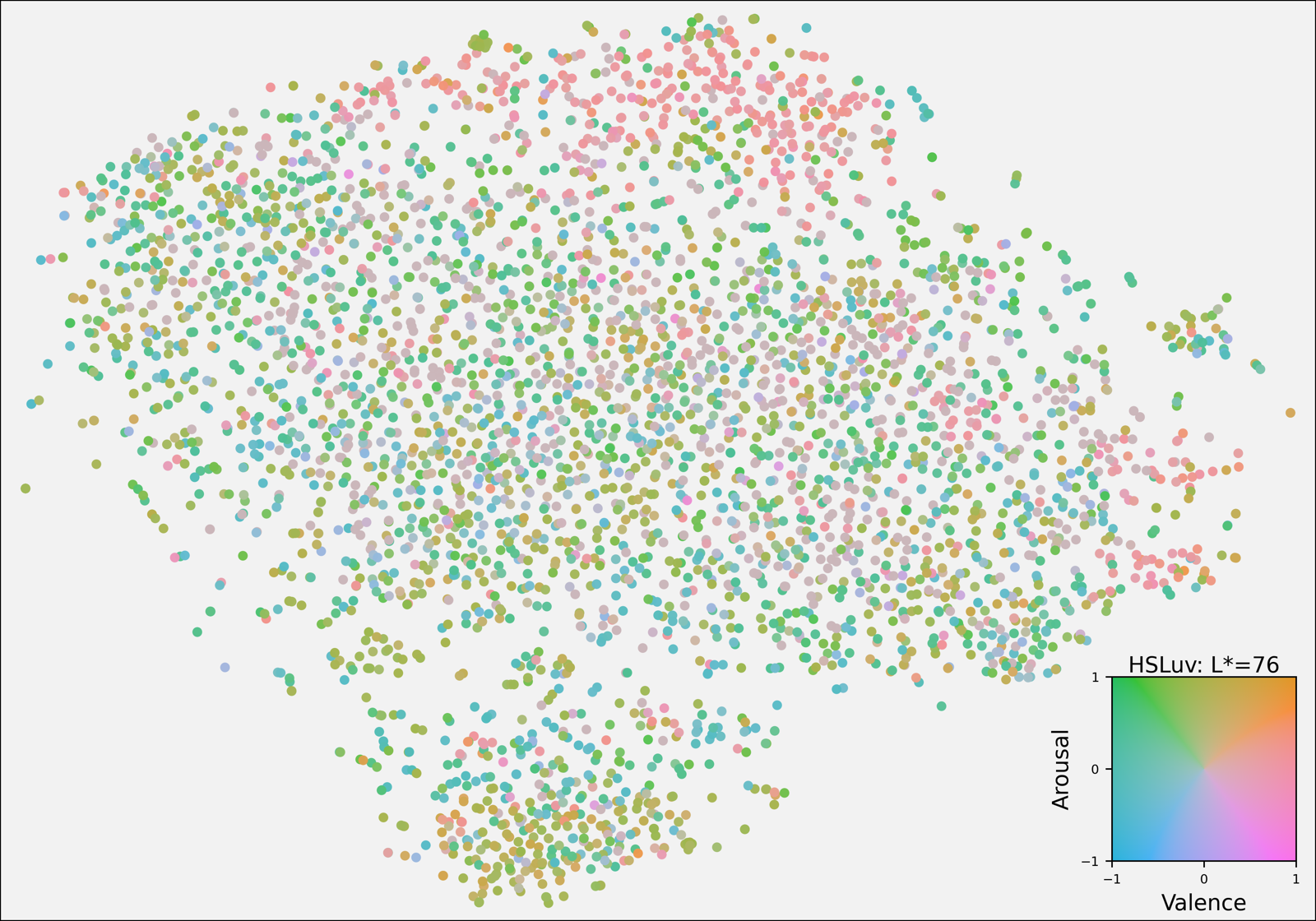}
    \caption{TEASER V\&A}
  \end{subfigure}

  \begin{subfigure}{\subw}
    \centering
    \includegraphics[width=\linewidth]{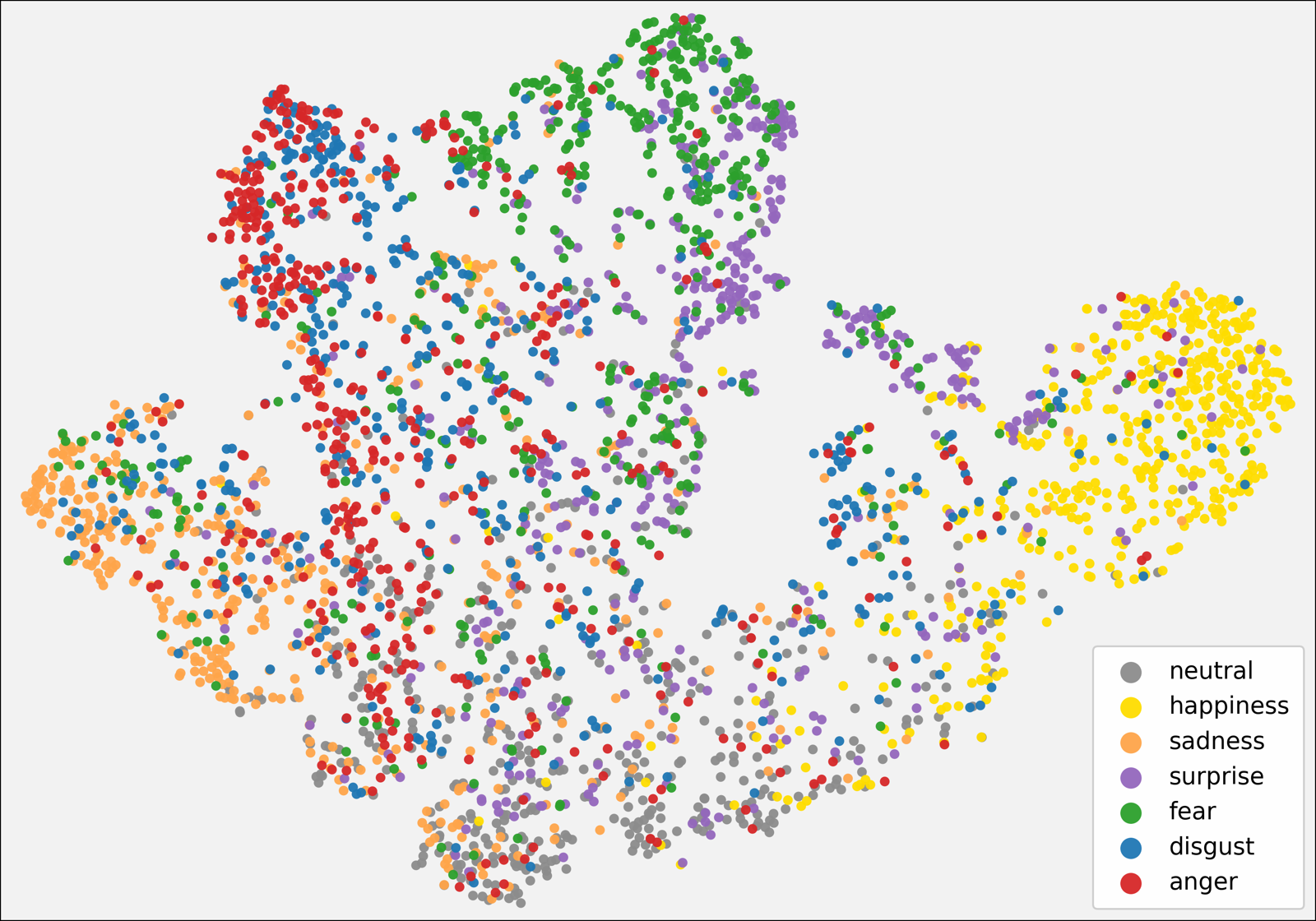}
    \caption{FIELDS Class}
  \end{subfigure}\hspace{\colgap}%
  \begin{subfigure}{\subw}
    \centering
    \includegraphics[width=\linewidth]{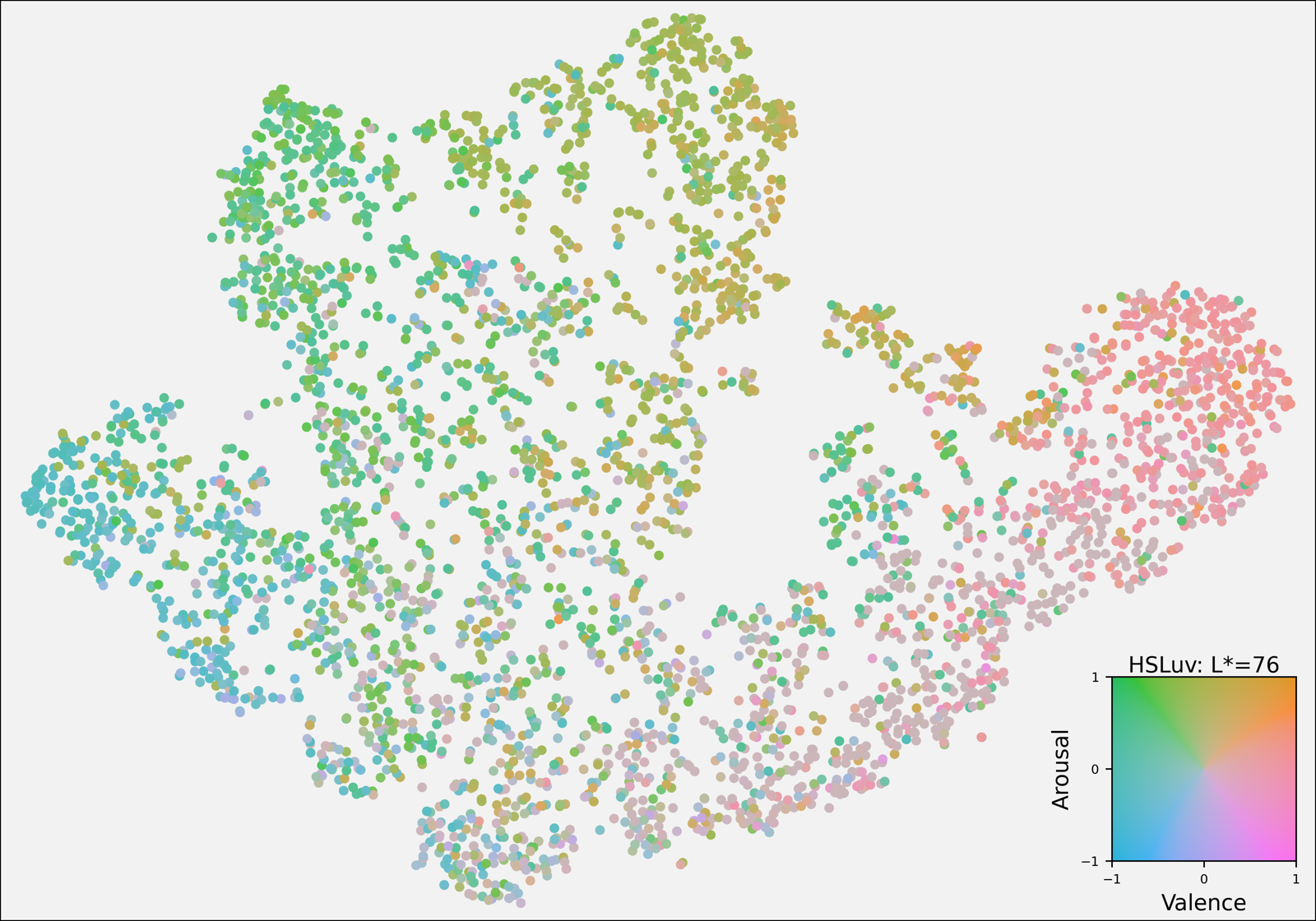}
    \caption{FIELDS V\&A}
  \end{subfigure}

  \caption{t-SNE Cluster Visualization of FLAME Expression Parameters.}
  \label{fig:tsne}
\end{figure*}

Fig.~\ref{fig:tsne} compares the t-SNE clustered embeddings across methods: FIELDS shows more coherent class-wise structure and clearer grouping for several frequent categories, while known hard pairs still exhibit partial overlap. When colored by valence-arousal, the FIELDS embedding exhibits a smoother intensity gradient, consistent with its improved VA regression results. These qualitative trends suggest that the affective utility is carried, at least in part, by the learned expression codes rather than relying solely on identity-related cues.

\begin{figure*}[t]
  \centering
   \captionsetup[subfigure]{font=scriptsize}
  \newcommand{\subw}{0.41\textwidth}   
  \newcommand{\colgap}{0.03\textwidth} 
  
  \begin{subfigure}{\subw}
    \centering
    \includegraphics[width=\linewidth]{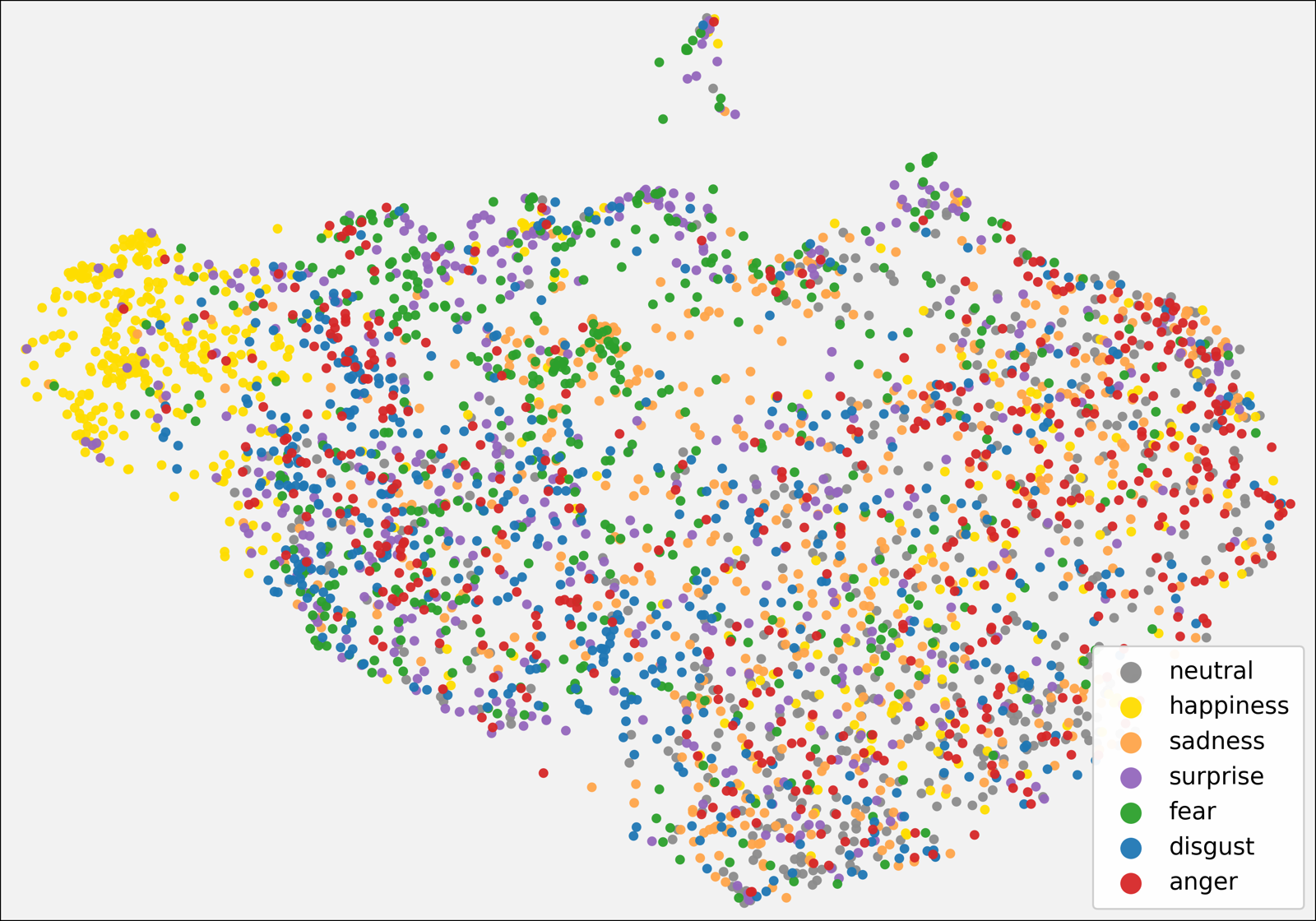}
    \caption{EMOCA Class}
  \end{subfigure}\hspace{\colgap}%
  \begin{subfigure}{\subw}
    \centering
    \includegraphics[width=\linewidth]{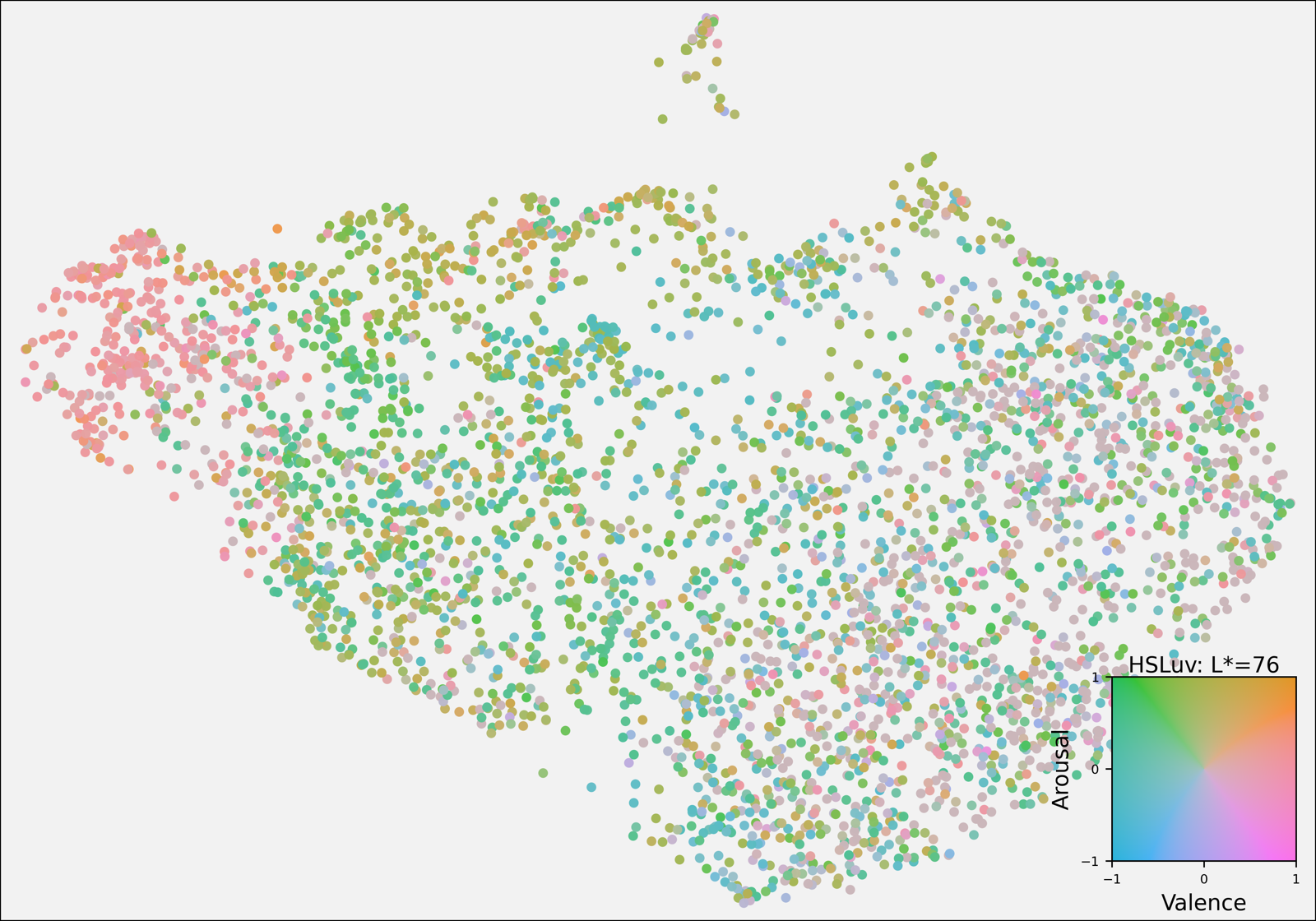}
    \caption{EMOCA V\&A}
  \end{subfigure}


  \begin{subfigure}{\subw}
    \centering
    \includegraphics[width=\linewidth]{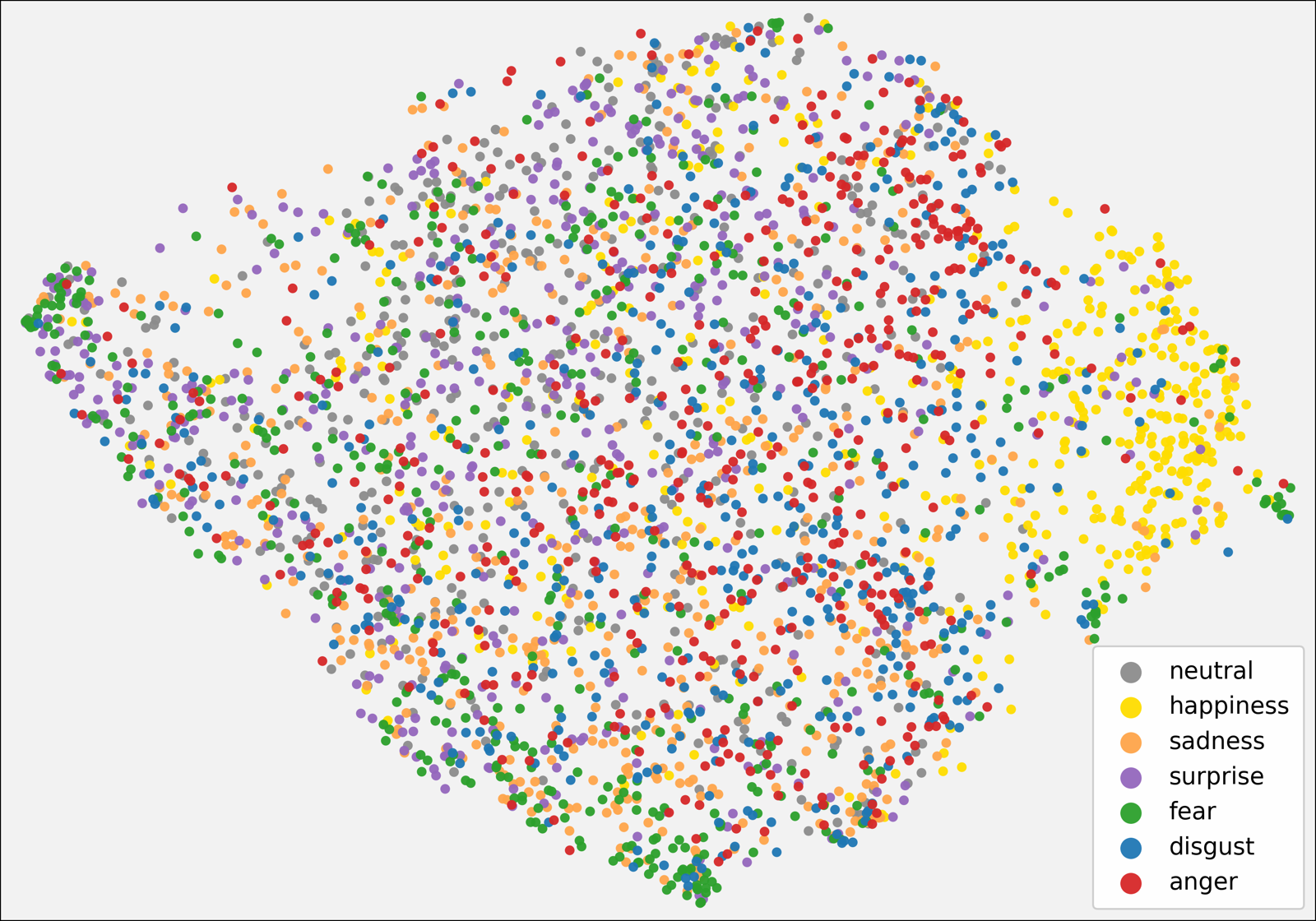}
    \caption{SMIRK Class}
  \end{subfigure}\hspace{\colgap}%
  \begin{subfigure}{\subw}
    \centering
    \includegraphics[width=\linewidth]{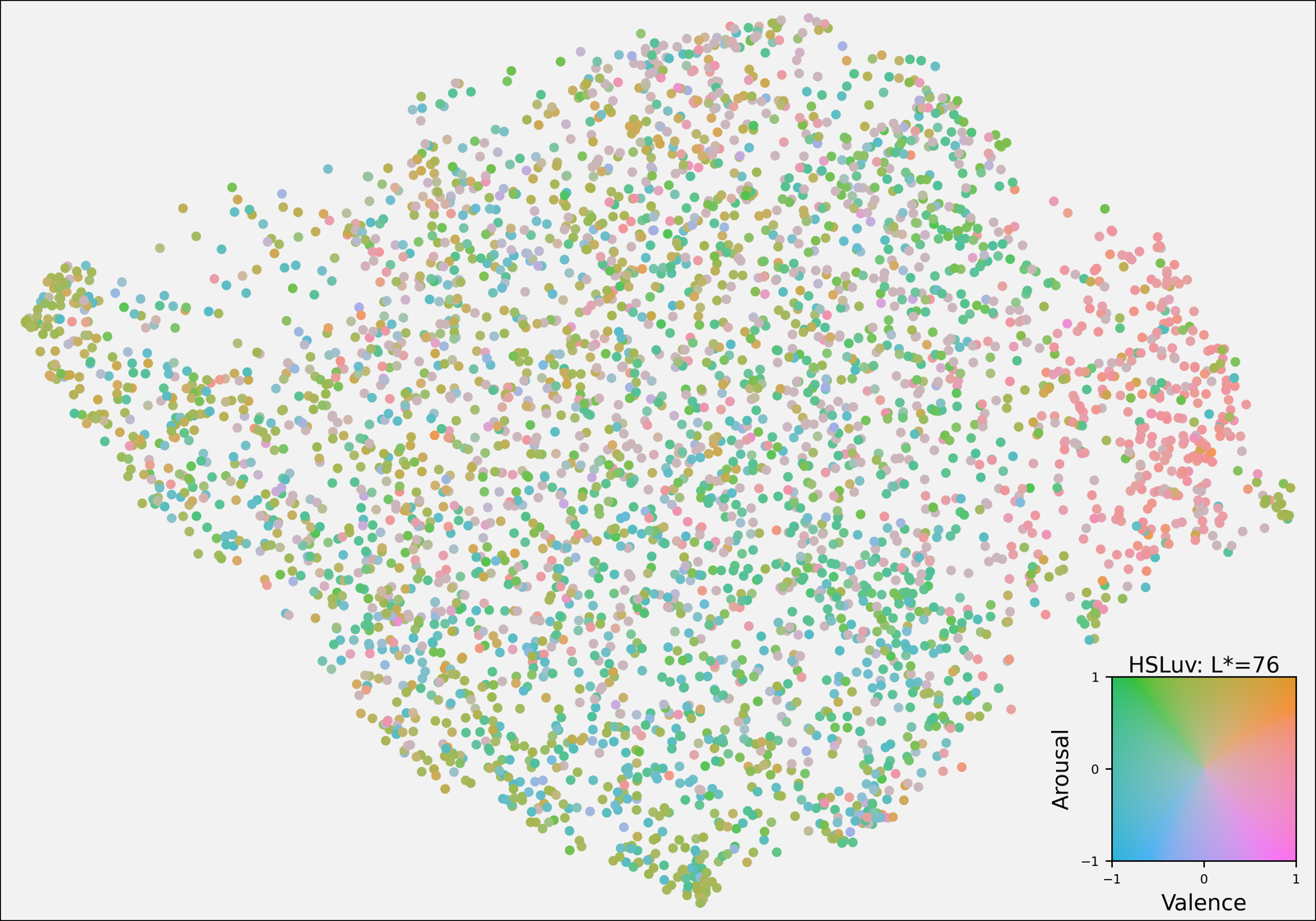}
    \caption{SMIRK V\&A}
  \end{subfigure}


  \begin{subfigure}{\subw}
    \centering
    \includegraphics[width=\linewidth]{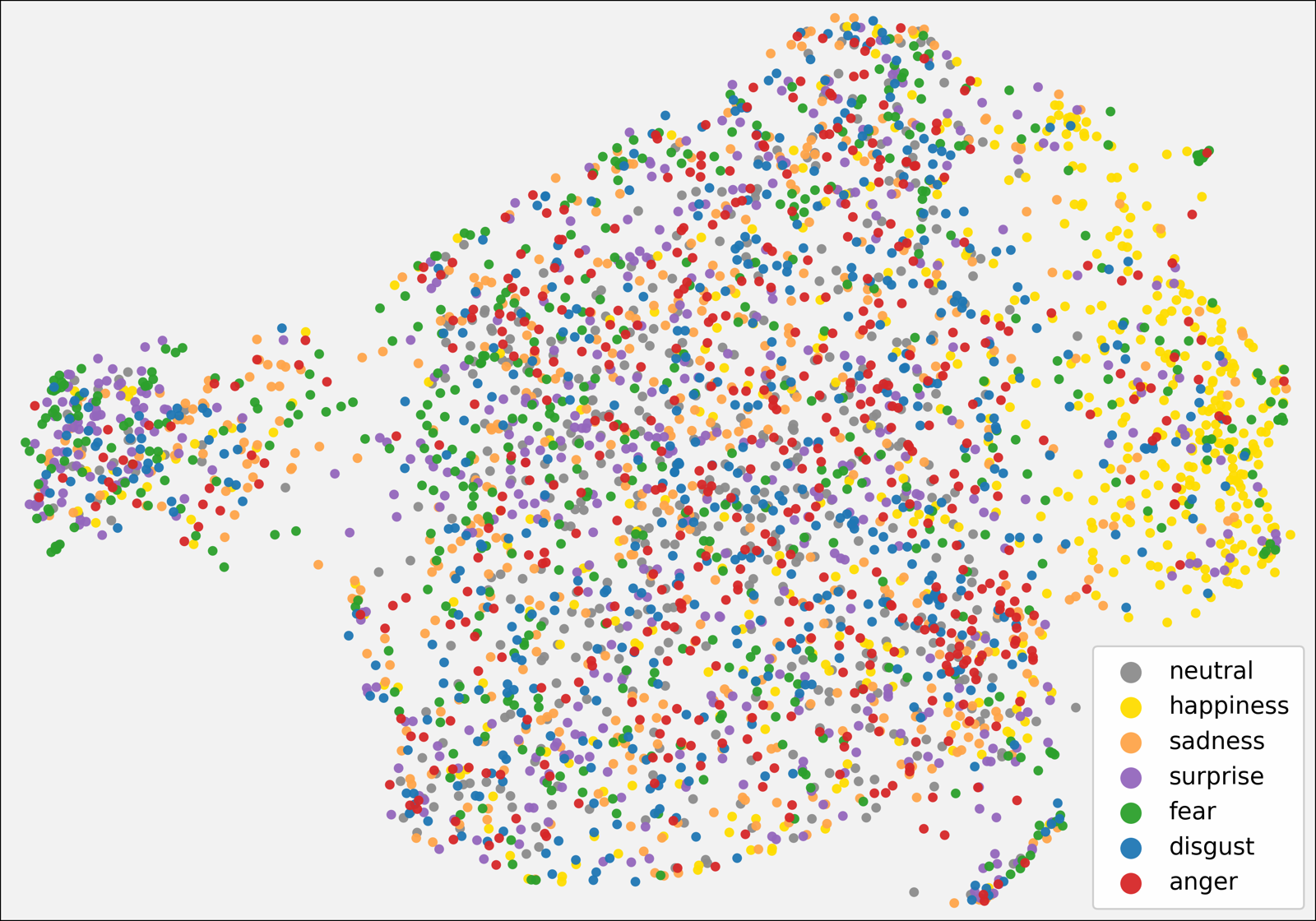}
    \caption{TEASER Class}
  \end{subfigure}\hspace{\colgap}%
  \begin{subfigure}{\subw}
    \centering
    \includegraphics[width=\linewidth]{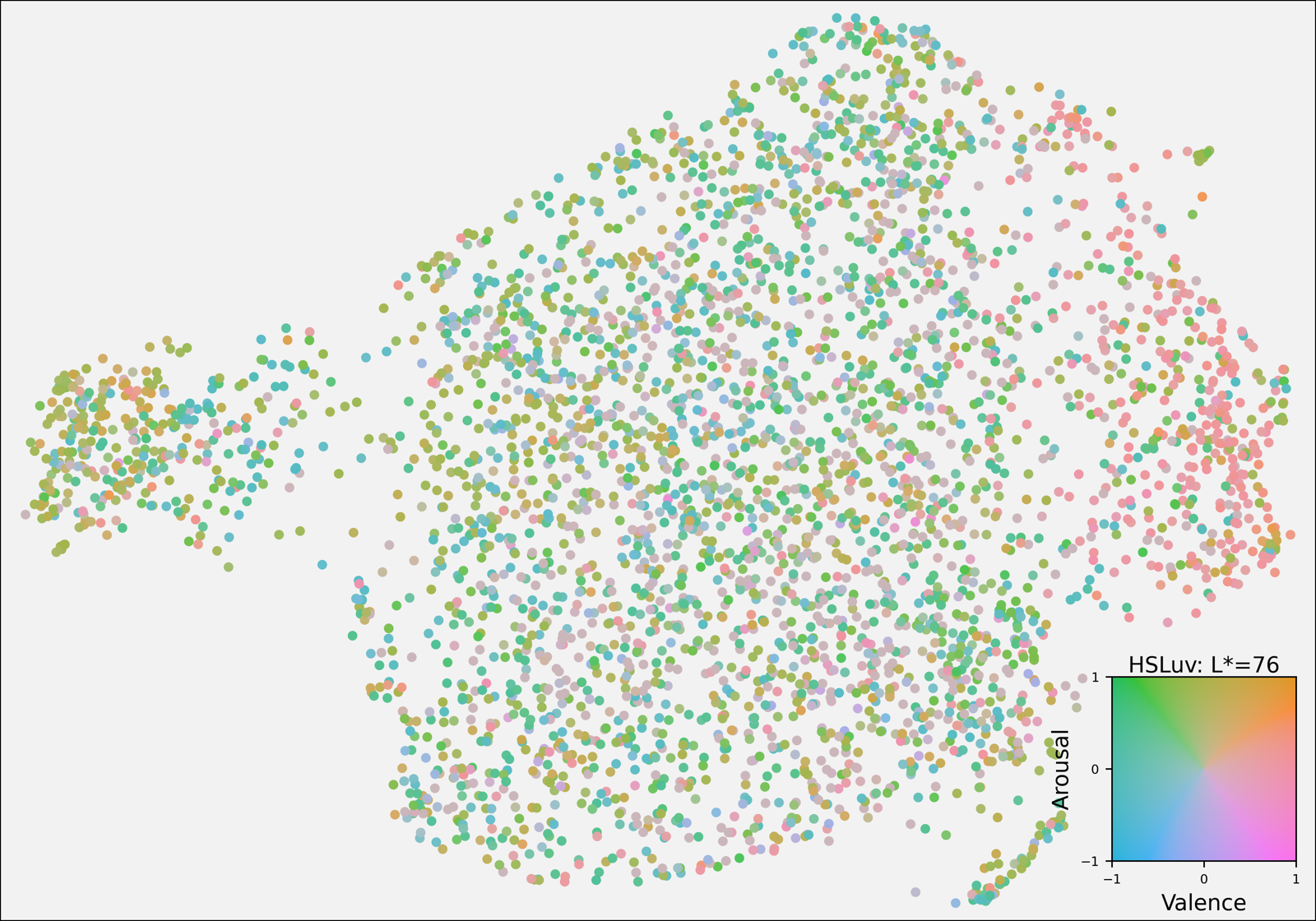}
    \caption{TEASER V\&A}
  \end{subfigure}
  \begin{subfigure}{\subw}
    \centering
    \includegraphics[width=\linewidth]{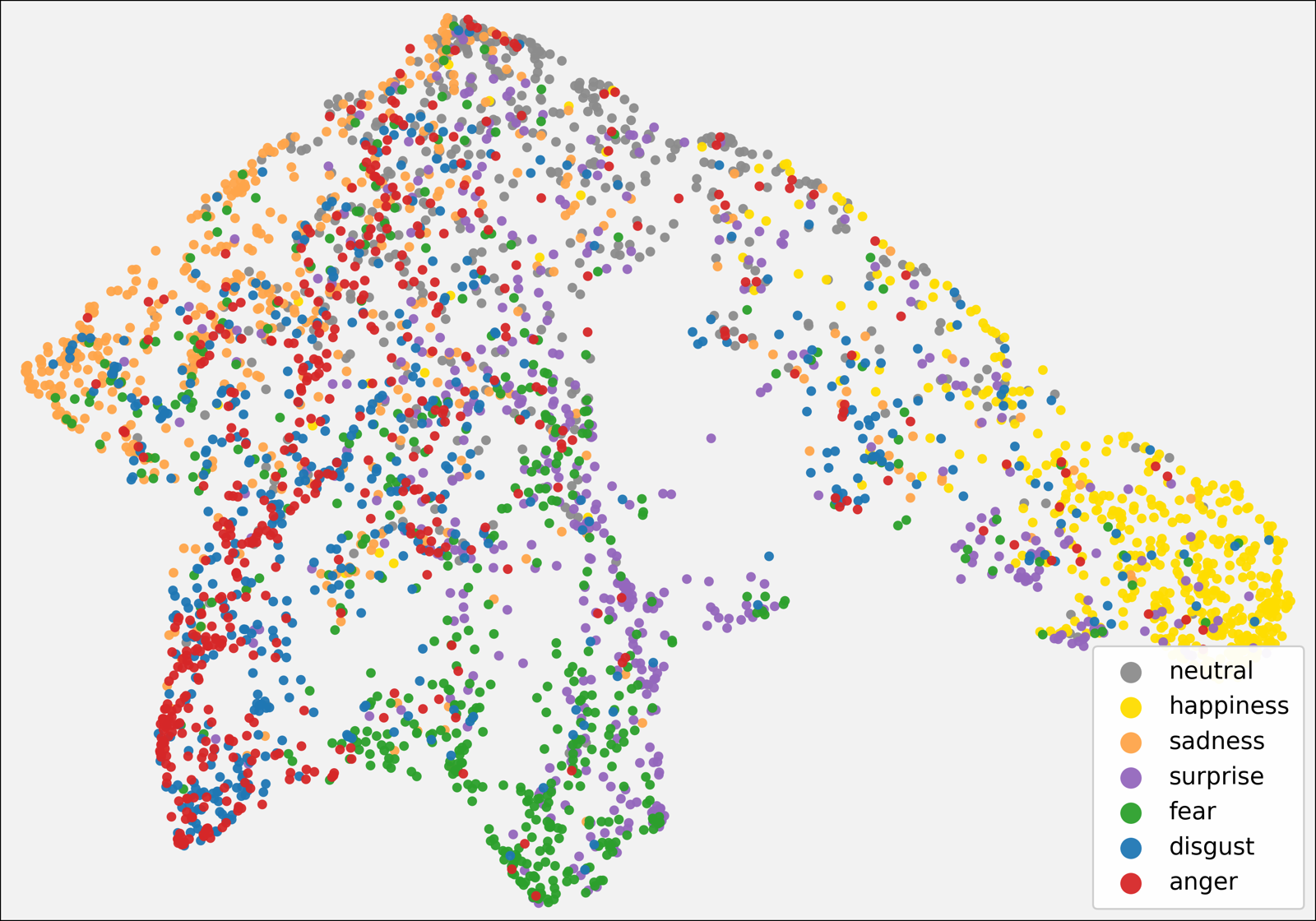}
    \caption{FIELDS Class}
  \end{subfigure}\hspace{\colgap}%
  \begin{subfigure}{\subw}
    \centering
    \includegraphics[width=\linewidth]{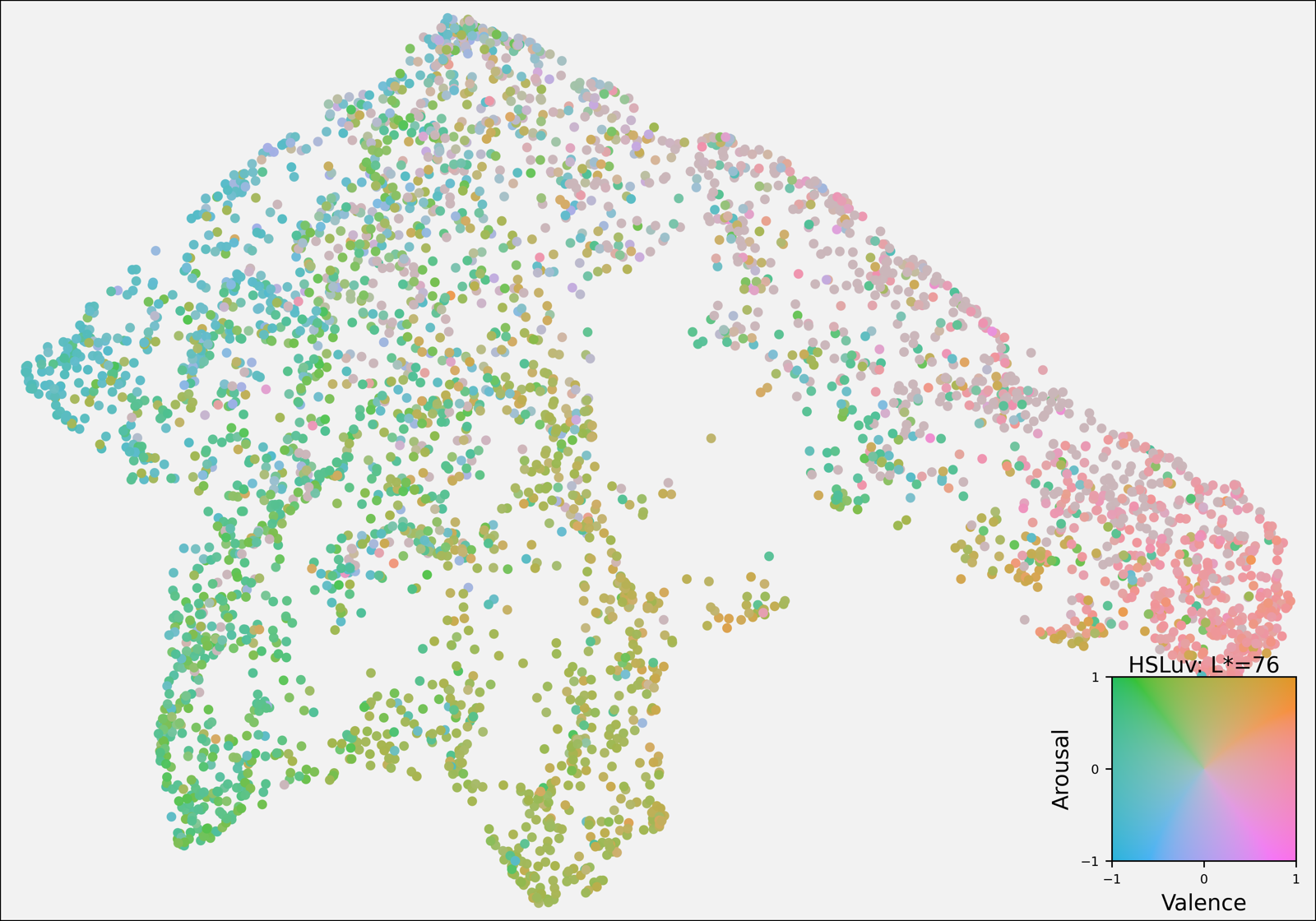}
    \caption{FIELDS V\&A}
  \end{subfigure}

  \caption{UMAP Cluster Visualization of FLAME Expression Parameters.}
  \label{fig:umap}
\end{figure*}

UMAP in Fig.~\ref{fig:umap} offers a complementary view by better revealing the global organization and continuous affective transitions in the embedding space.

\end{document}